\documentclass[]{jair}

\setcopyright{cc}
\copyrightyear{2026}
\acmDOI{10.1613/jair.1.20550}

\JAIRAE{Ofra Amir}
\JAIRTrack{Surveys} % Insert JAIR Track Name only if part of a special track
\acmVolume{87}
\acmArticle{16}
\acmMonth{10}
\acmYear{2026}

\RequirePackage[
  datamodel=acmdatamodel,
  style=acmauthoryear,
  backend=biber,
  giveninits=true,
  uniquename=init
  ]{biblatex}

\usepackage{multirow}

\begin{document}

%\title{JAIR Example Template}
\title[A Survey of XRL]{A Survey of Explainable Reinforcement Learning: \\ Targets, Methods and Needs}

%% used to denote shared contribution to the research and/or corresponding author.
\author{Léo Saulières}
\authornote{Corresponding Author.}
\orcid{0000-0002-4800-9181}
\email{leo.saulieres@inrae.fr}
\affiliation{%
  \institution{Univ Toulouse, INRAE, MIAT}
  \city{Castanet-Tolosan}
  \country{France}
}

%% The short list of authors must be made of the list of all authors' lastnames.
\renewcommand{\shortauthors}{Saulières}
%% If this is too long and overlaps other information printed in the page headers, use
%\renewcommand{\shortauthors}{Xu et al.}

\begin{abstract}
% Powerful AI models
The success of recent Artificial Intelligence (AI) models has been accompanied by the opacity of their internal mechanisms, due notably to the use of deep neural networks.
% XAI
In order to understand these internal mechanisms and explain the output of these AI models, a set of methods have been proposed, grouped under the domain of eXplainable AI (XAI).
% Sub-domain XRL
This paper focuses on a sub-domain of XAI, called eXplainable Reinforcement Learning (XRL), which aims to explain the actions of an agent that has learned by reinforcement learning.
% Short taxonomy description + 250 papers
We propose an intuitive taxonomy based on two questions ‘What’ and ‘How’. The first question focuses on the target that the method explains, while the second relates to the way the explanation is provided. We use this taxonomy to provide a state-of-the-art review of over 250 papers.
% Related domains + needs for XRL
In addition, we present a set of domains close to XRL, which we believe should get attention from the community. Finally, we identify some needs for the field of XRL.
\end{abstract}

%\begin{abstract}
%      A clear and well-documented \LaTeX\ document is presented as an
%  article formatted for publication by ACM in a conference proceedings
%  or journal publication. Based on the ``acmart'' document class, this
%  article presents and explains many of the common variations, as well
%  as many of the formatting elements an author may use in the
%  preparation of the documentation of their work.
%\end{abstract}

%% JAIR Note: 
%% Do not include ACM CCS Concepts or Keywords

%% To be updated by authors.
\received{25 September 2025}
\received[accepted]{02 August 2026}

\maketitle

%%%%%%%%%%%%%%%%%%%%%%%%%%%%%%%%%%%%%%%%%%%%%%%%%%%%%%%%%%%%%%%%%%%%%%%%
\section{Introduction}
\label{sec:intro_SOTA}
%%%%%%%%%%%%%%%%%%%%%%%%%%%%%%%%%%%%%%%%%%%%%%%%%%%%%%%%%%%%%%%%%%%%%%%%

% Introduction
This paper presents a state of the art on explainable and transparent reinforcement learning. It brings together a set of relatively recent works that present new methods specific to Reinforcement Learning (RL) as well as papers using methods derived from the eXplainable Artificial Intelligence (XAI) domain, originally used to explain classifiers, such as LIME (Ribeiro et al. \citeyear{DBLP:conf/kdd/Ribeiro0G16}) or SHAP (Lundeberg and S. Lee \citeyear{DBLP:conf/nips/LundbergL17}).

% RL
RL is a machine learning paradigm where an agent learns to make a sequence of actions within an environment. Given a set of information defined as a state, the agent chooses an action at each time-step, arrives in a new state and receives a reward, determined by the environment's dynamics (transition and reward functions).
The agent's goal is to maximize its cumulative reward (also called return) by learning an optimal policy.

% RL sub-domains
In addition to the eXplainable RL (XRL) methods, we briefly present a set of sub-domains of RL whose main motivation is the performance and generalisation of agent policies. Indeed, they can also be used to explain or make transparent the agent's behavior. 
For example, Relational RL (RRL)  (Dzeroski et al. \citeyear{DBLP:journals/ml/DzeroskiRD01}) consists of the agent reasoning on the basis of \emph{relations} and \emph{objects} rather than reasoning directly with raw data. The relationships on which the agent relies can then be used to explain its behavior.  

This overview is based on a total of $12$ states of the art (Alaarabiou et al. \citeyear{alaarabiouexplicabilite}; Dazeley et al. \citeyear{DBLP:journals/nca/DazeleyVC23}; Gajcin and Dusparic \citeyear{DBLP:journals/csur/GajcinD24}; Glanois et al. \citeyear{DBLP:journals/ml/GlanoisWZLYHL24}; Heuillet et al. \citeyear{DBLP:journals/kbs/HeuilletCR21}; Hickling et al. \citeyear{DBLP:journals/csur/HicklingZAS24}; Krajna et al. \citeyear{DBLP:journals/corr/abs-2203-11547}; Milani, Topin et al. \citeyear{DBLP:journals/csur/MilaniTVF24}; Puiutta and Veith \citeyear{DBLP:conf/cdmake/PuiuttaV20}; Vouros \citeyear{DBLP:journals/csur/Vouros23}; Wells and Bednarz \citeyear{DBLP:journals/frai/WellsB21}; Zelveder et al. \citeyear{DBLP:conf/atal/ZelvelderWF21}) and complementary papers.
    % Selection process
The papers we present are those identified in the 12 surveys mentioned. Furthermore, we have selected papers in the field of XRL cited by the papers presented in the states of the art. In addition, in order to provide as comprehensive an overview as possible, we searched for papers via Google Scholar using the following keywords: \emph{`Reinforcement Learning'}, \emph{`Explainable'}, \emph{`Interpretable'}, \emph{`Understandable'}, \emph{`Transparent'}.
This additional selection was based solely on verifying that papers propose or use an XRL method, without considering the number of citations. 
The cutoff year for paper selection was 2024.

%  Goals of this survey
Our paper has four objectives:
\begin{itemize}
    \item To reflect the methods designed to explain/make transparent agents that have learned by reinforcement. 
    \item To present works that use model-agnostic methods.
    \item To briefly describe domains related to XRL on which researchers should focus.
    \item To draw a list of the needs for the XRL domain.
\end{itemize}

% Describe why interpretability is considered as explainability in this survey
In this state of the art, we present methods aimed at \emph{providing an explanation} as well as those aimed at \emph{making the agent's behavior transparent} by design. 
We believe it is important to cover both approaches in order to give the reader an overview of the methods used to make the agent's behavior understandable to the user. By abuse of language, we refer to both approaches when talking about explanations or XRL in this paper. 

% Examples of Taxonomy
Among researchers in the field of XRL, there is no consensus on the taxonomy to be used to classify the various works. 
To illustrate, Puiutta and Veith %(\citeyear{DBLP:conf/cdmake/PuiuttaV20})
use the taxonomy of classifiers, Milani et al.  %(\citeyear{DBLP:journals/csur/MilaniTVF24}) 
separate the methods using categories that better reflect the aspects of RL and Dazeley et al. 
%(\citeyear{DBLP:journals/nca/DazeleyVC23}) 
use the Causal Explanation Framework (Böhm and Pfister \citeyear{bohm2015people}) to propose a conceptual framework for XRL.

\begin{figure*}[h]
        \centering
        \includegraphics[width=0.9\textwidth]{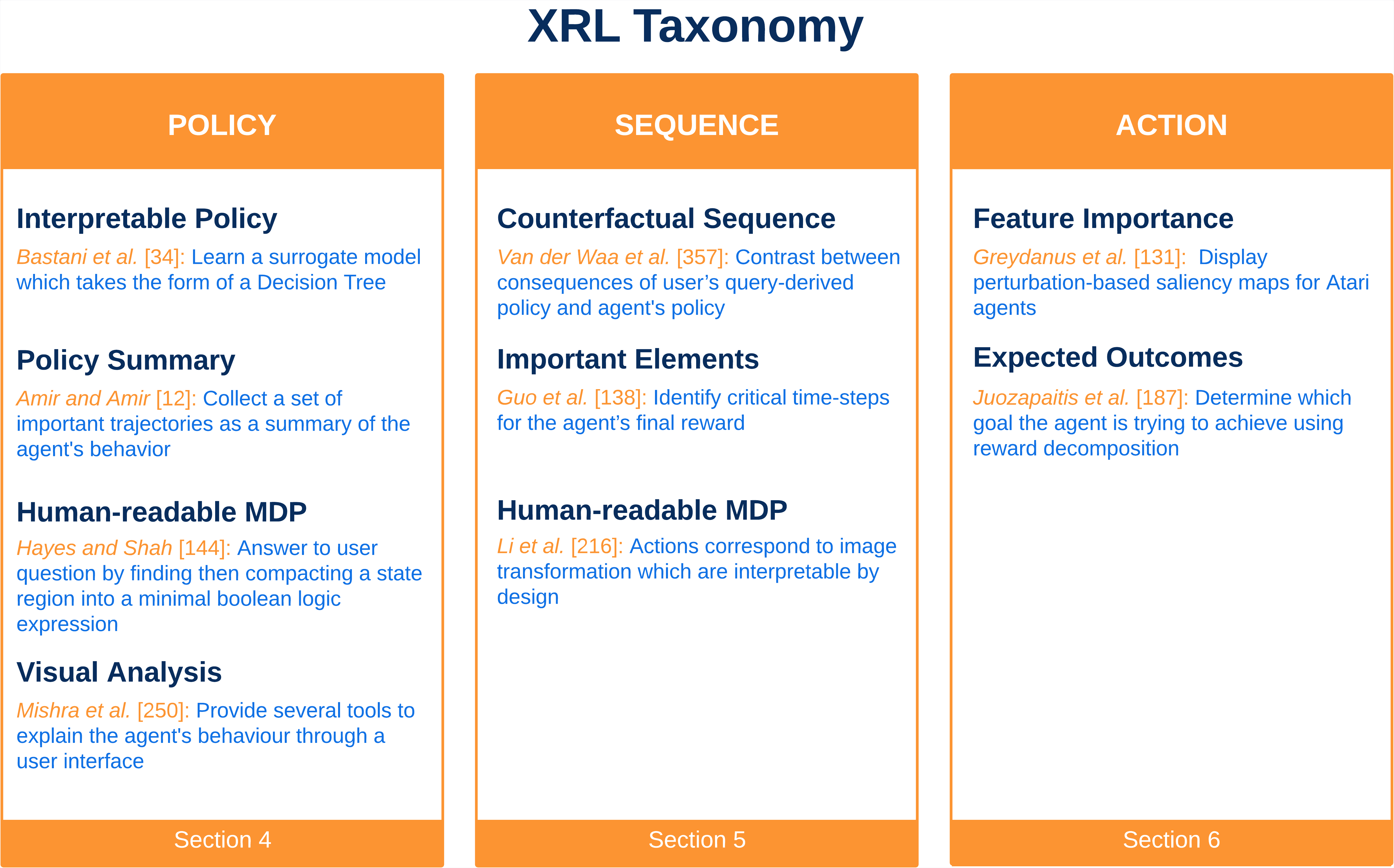}
        \caption[XRL taxonomy overview]{Our taxonomy for Explainable Reinforcement Learning.}
        \label{fig_ch1:Taxonomy}
    \end{figure*}

% Describe our taxonomy (policy / sequence / action)
In the absence of a standard taxonomy, we propose in this paper a new taxonomy based simply on 2 questions: \emph{What} and \emph{How}. More specifically, the first question is used to determine what the method is trying to explain: \emph{``What does the method want to explain? What does it want to make transparent?''}. We have identified three elements: the agent's \emph{policy}, a particular \emph{sequence} of interactions between the agent and the environment, and a particular \emph{action} by the agent in a given situation. The second question allows us to refine the taxonomy, by looking at the way in which the explanation is provided: \emph{``How is it explained?''}. Thus, a set of ways of providing explanations are described in this state of the art, for each of the three above-mentioned elements. An overview of the taxonomy is presented in Figure \ref{fig_ch1:Taxonomy}, where an illustrative work is associated with each sub-category.

% Motivation + Need of a new survey 
We believe that this taxonomy can help readers to quickly identify a body of work that is relevant to the question they want to answer (and also the way in which they want to answer it). This state of the art also makes it possible to determine trends, whether in terms of elements explained or approaches used.

% Plan
The rest of the paper is structured as follows. Section \ref{sec:background_SOTA} provides the background of reinforcement learning. Section \ref{sec:rw_SOTA} describes related work.  Section \ref{sec:policy_SOTA} presents the methods explaining the agent's policy. Section \ref{sec:sequence_SOTA} describes the methods explaining the agent's behavior in a sequence of interactions with the environment. Section \ref{sec:action_SOTA} presents the methods explaining an agent's choice of action from a state. %\leo{TODO: detailed taxonomy cf Figure} 
Section \ref{sec:domains_SOTA}  briefly describes domains related to XRL that we think researchers should pay attention to. Section \ref{sec:needs_SOTA}  highlights the needs for the XRL domain according to the different surveys we studied. 

%%%%%%%%%%%%%%%%%%%%%%%%%%%%%%%%%%%%%%%%%%%%%%%%%%%%%%%%%%%%%%%%%%%%%%%%
\section{Background}
\label{sec:background_SOTA}
%%%%%%%%%%%%%%%%%%%%%%%%%%%%%%%%%%%%%%%%%%%%%%%%%%%%%%%%%%%%%%%%%%%%%%%%
    % MDP
An RL problem is described by a Markov Decision Process (MDP), which is a tuple $\langle \mathcal{S}, \mathcal{A}, R, p \rangle$, where: $\mathcal{S}$ and $\mathcal{A}$ denotes respectively the state space and action space; $R : \mathcal{S}\times \mathcal{A} \rightarrow \mathbb{R}$ is the reward function that provides a scalar feedback signal evaluating the immediate outcome of an action taken in a given state; $p : \mathcal{S} \times \mathcal{A} \rightarrow Pr(\mathcal{S})$ is the transition function that defines the probability distribution over next states given the current state and action.
    % State are composed of features, where each feature f_i can have a value from domain D_i
States are represented by a set of features $\{f_1, f_2, ..., f_n\}$, where each feature $f_i$ takes values in a domain $D_i$.

    % Agent's goal :  Policy (deterministic and stochastic) / V / Q-function - value
The agent's goal is to learn a policy $\pi$ that maximizes the expected cumulative reward, or return. A policy can be either deterministic, mapping each state to an action $\pi(s)=a$, or stochastic, mapping each state to a probability distributions over actions $\pi(a|s)$. To evaluate policies, value functions are defined: the state-value function $V^\pi(s)$ estimates the expected return from state $s$ when following policy $\pi$, and the action-value function $Q^\pi(s,a)$ estimates the expected return when taking action $a$ in state $s$ and subsequently following $\pi$.

    % Q learning
Q-learning is an RL algorithm that learns an approximation of the optimal action-value function $Q^*(s,a)$. It updates estimates iteratively using sampled transitions from interaction within the environment. The update rule is as follows: $Q(s,a) \leftarrow Q(s,a) + \alpha \big(r + \gamma \max_{a'} Q(s',a') - Q(s,a)\big)$ where $\alpha$ is the learning rate, $\gamma$ the discount factor, $r$ the received reward, and $s'$ the next state. The learned $Q$-values allow the agent to derive a policy by selecting actions with maximal expected value.
    
    % Extensions : Multi-agent RL / Partially observable MDP 
Several extensions of the standard MDP formulation have been proposed to address more complex settings, e.g. Partially Observable MDP (POMDP) and Multi-Agent RL (MARL). In MARL, multiple agents interact within the same environment and must learn while accounting for the behavior of other agents. 
In a POMDP, the agent does not have a full visibility of the state it is in. Instead, the agent receives partial information about the state, called an observation.

    % Use of Neural networks for difficult problem: architectures (Multi-Layer Perceptron / Convolutional NN / Recurrent NN) / Deep RL algorithms: DQN, A3C
For complex problems with large or continuous state/action spaces, value functions
or policies can be approximated using neural networks (NN).
Common NN architectures include Multi-Layer Perceptron (MLP), Convolutional Neural Network (CNN) for spatial data such as images, and Recurrent Neural Network (RNN) for sequential data or partially observable settings.
Combining RL with NN has led to the development of Deep Reinforcement Learning (DRL) algorithms such as Deep Q-Network (DQN) (Mnih, Kavukcuoglu, et al. \citeyear{DBLP:journals/nature/MnihKSRVBGRFOPB15}), which approximates the Q-function with a NN, and Asynchronous Advantage Actor-Critic (A3C) (Mnih, Badia, et al. \citeyear{DBLP:conf/icml/MnihBMGLHSK16}), which jointly learns a policy and a value function.

%%%%%%%%%%%%%%%%%%%%%%%%%%%%%%%%%%%%%%%%%%%%%%%%%%%%%%%%%%%%%%%%%%%%%%%%
\section{Related Work}
\label{sec:rw_SOTA}
%%%%%%%%%%%%%%%%%%%%%%%%%%%%%%%%%%%%%%%%%%%%%%%%%%%%%%%%%%%%%%%%%%%%%%%%

% Many surveys that differ in the taxonomy and the focus/scope
Several surveys on XRL have been proposed in recent years, differing both in their scope and in the taxonomies used to organize the literature (Alaarabiou et al. \citeyear{alaarabiouexplicabilite}; Amitai and O. Amir \citeyear{amitai2024survey}; Dazeley et al. \citeyear{DBLP:journals/nca/DazeleyVC23}; Gajcin and Dusparic \citeyear{DBLP:journals/csur/GajcinD24}; Glanois et al. \citeyear{DBLP:journals/ml/GlanoisWZLYHL24}; Heuillet et al. \citeyear{DBLP:journals/kbs/HeuilletCR21}; Hickling et al. \citeyear{DBLP:journals/csur/HicklingZAS24}; Krajna et al. \citeyear{DBLP:journals/corr/abs-2203-11547}; Milani, Topin et al. \citeyear{DBLP:journals/csur/MilaniTVF24}; Puiutta and Veith \citeyear{DBLP:conf/cdmake/PuiuttaV20}; Vouros \citeyear{DBLP:journals/csur/Vouros23}; Wells and Bednarz \citeyear{DBLP:journals/frai/WellsB21}; Zelveder et al. \citeyear{DBLP:conf/atal/ZelvelderWF21}). 

There is no consensus on which taxonomy to adopt for XRL, as the following examples illustrate.
% No consensus of taxonomy
Puiutta and Veith %(\citeyear{DBLP:conf/cdmake/PuiuttaV20})
distinguishing explanations by their scope and the time at which they are extracted. 
In terms of scope, a \emph{local} explanation explains an agent's choice of action, whereas a \emph{global} explanation describes the agent's policy as a whole. 
In terms of time, an \emph{intrinsic} explanation means that the model can be interpreted by itself, whereas a \emph{post-hoc} explanation means that the explanation is provided after the agent's learning phase. 
Milani et al. %(\citeyear{DBLP:journals/csur/MilaniTVF24}) 
separate the methods using categories that better reflect the aspects of RL. 
Works are separated into three categories: \emph{feature importance}, which consists of identifying the features of a state that impact the agent's decision, \emph{learning process and MDP}, which shows the past experiences or the components of the MDP that lead to the agent's behavior and \emph{policy level}, which describes the agent's long-term behavior.
As a final example, Dazeley et al.  %(\citeyear{DBLP:journals/nca/DazeleyVC23}) 
use the Causal Explanation Framework (Böhm and Pfister \citeyear{bohm2015people}) to propose a conceptual framework for XRL, called the Causal XRL Framework. A simplified version is used to categorise the methods into two types: \emph{perception} and \emph{action}. 
The first category includes methods that focus on the impact of the agent's perception on its actions and outcomes. The second category includes methods that focus on the choice of action and its impact on the outcome.

% Specific
Most of the surveys either focus on a specific subset of explanation methods or illustrate the landscape of XRL approaches using a relatively limited number of representative works.
Among the states of the art, Glanois et al. focuses on the interpretability of RL, Amitai and O. Amir focuses on global explanations, and Gajcin and Dusparic focuses on counterfactual explanations whereas the others are not specific to one type of explanation. 
A counterfactual explanation is used to determine which part of the input (e.g. a state in RL) must be modified in order to change the model's output (e.g. the agent's policy). It is interesting to note that this type of method is under-represented in XRL (Gajcin and Dusparic).

% Proposition of an original taxonomy to (try to) categorise each work in the XRL domain.
In contrast to these surveys, this work aims to provide a broad overview of the XRL literature. 
To this end, we review more than $250$ papers and introduce a taxonomy designed to organize the different approaches according to two complementary dimensions: the \emph{target of the explanation} and the \emph{way the explanation is produced}. 
This structure allows groups of related works to be quickly identified by jointly considering what target is explained and how the explanation is generated.
% Post-hoc vs intrinsic and model-specific vs model agnostic in our taxonomy
Within this taxonomy, commonly used distinctions from the XAI literature fall within the \emph{`How'} dimension. 
In particular, \emph{intrinsic} explanations correspond to methods in which the policy is interpretable by design (e.g. taking the form of a set of rules), whereas \emph{post-hoc} explanations are generated after the learning phase without taking into account the interpretability of the policy to be explained. In the RL context, methods that make use of an MDP that is interpretable by design and those that transform an MDP into an interpretable MDP can be regarded as \emph{intrinsic} and \emph{post-hoc} methods respectively, given their aim of explaining the agent’s policy. 
Similarly, methods can be distinguished according to whether they are \emph{model-specific}, based on a specific type of policy (e.g. work focused on explaining DQN and RNN policies presented in this survey), or \emph{model-agnostic}, treating the agent as a black box and deriving explanations from its observed behavior or through surrogate models.

% How does it improve to previous taxonomies
The proposed taxonomy aims to facilitate the exploration of the XRL literature by providing a simple structure that highlights the relationships between explanation targets and explanation ways. 
By explicitly separating the explanation target from the way of explanation, it becomes easier to compare methods that explain similar targets using different techniques, or conversely methods that rely on similar way of explanation but different targets. 
This perspective helps reveal connections between works that are often presented independently in the literature and provides a clearer overview of the XRL domain.

%%%%%%%%%%%%%%%%%%%%%%%%%%%%%%%%%%%%%%%%%%%%%%%%%%%%%%%%%%%%%%%%%%%%%%%%
\section{Policy-level Methods}
\label{sec:policy_SOTA}
%%%%%%%%%%%%%%%%%%%%%%%%%%%%%%%%%%%%%%%%%%%%%%%%%%%%%%%%%%%%%%%%%%%%%%%%

\begin{figure*}[h]
        \centering
        \includegraphics[width=1.0\textwidth]{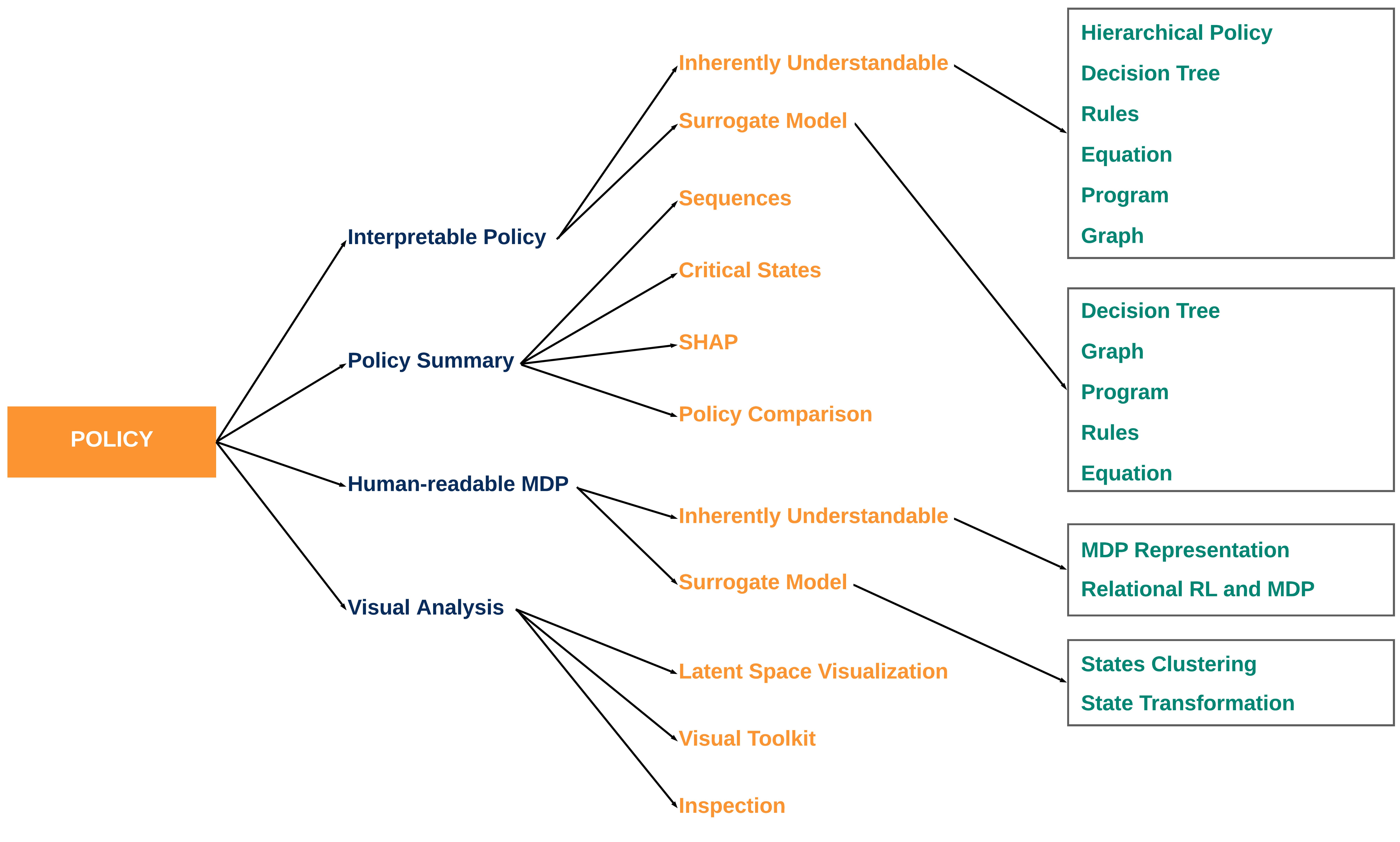}
        \caption[XRL taxonomy: policy-level methods]{Detailed taxonomy for policy-level methods.}
        \label{fig_ch1:Taxonomy_policy}
    \end{figure*}

% Explanation of Taxonomy figures + order groups of work
Before delving into the state of the art, we briefly explain how to interpret Figures \ref{fig_ch1:Taxonomy_policy}, \ref{fig_ch1:Taxonomy_sequence}, \ref{fig_ch1:Taxonomy_action}, which abstractly represent the taxonomy used for each target, as well as the order in which the groups of methods are presented in the paper.
Each figure represents a left-to-right hierarchy of the taxonomy. In each figure, the leftmost element denotes the target of the explanation. Moving to the right, blue nodes indicate high-level ways of explanation. These ways are refined into orange nodes which are then refined into green nodes, corresponding to more specific approaches of explanation. Branching indicates that a way of explanation can be divided into sub-groups of work.
In the remainder of the paper, for each target, the various groups of work are ordered in descending order according to the number of works identified per group.

As described in Figure~\ref{fig_ch1:Taxonomy_policy}, policy explanation is carried out using four ways of explanation.
Constructing \emph{Interpretable Policies} makes the agent’s reasoning transparent. 
\emph{Summarising Policies} provides an overview of the agent’s capabilities. 
The use of \emph{Human-readable MDP} allows the user to understand the information available to the agent, its actions and the dynamics of the environment. 
\emph{Visual analysis} helps the user to study the agent’s policy visually, using various tools and metrics. 
\emph{Interpretable Policies} are divided into two groups of work: learning an interpretable surrogate model of the policy and directly learning an interpretable policy. In this work, the policy can take different forms, such as a decision tree or an equation.
A policy summary can be composed of a set of state-action sequences or critical states, resulting from the use of SHAP, and can be used to compare policies.
\emph{Human-readable MDPs} are divided into two groups of work: Surrogate Model, which transforms the MDP to make it interpretable, and Inherently Understandable, which proposes an MDP that is interpretable by design.
\emph{Visual Analysis} brings together specific analysis of DQN latent spaces, visual interfaces and agent capacity inspection.

%------------------------------------------------
%------------------------------------------------
\subsection{Interpretable Policy}
%------------------------------------------------
%------------------------------------------------

A way of making the agent's behavior understandable to the user is to construct interpretable policies, thereby making the agent's decision-making transparent.
To do this, the agent's policy is directly learned in such a way as to be interpretable, or an interpretable surrogate model is learned to imitates the agent's opaque decision-making. 
To clarify, although these methods enable us to explain the agent's decision making (i.e. its choice of action in a particular situation), we choose to classify this type of approach as policy explanation, because the main result of these methods is an interpretable form of the policy, and the action explanation is simply derived from it.

    %------------------------------------------------
    \subsubsection{Inherently Understandable}
    %------------------------------------------------

    \begin{table*}[h]
        \centering
        \setlength\tabcolsep{5pt}
        \caption{Interpretable Policy works.}
        \vspace{2mm}
        \begin{tabular}{ c | c | l }
         \multicolumn{2}{c |}{\textbf{Type}} & \textbf{Refs}\\ 
         \hline
         
                  \multirow{19}{*}{Inherently Understandable {\scriptsize(54)}} &  \multirow{7}{*}{Hierarchical Policy {\scriptsize(20)}} & {\small Y. Jiang et al. \citeyear{DBLP:conf/nips/JiangGMF19}; Shu et al. \citeyear{DBLP:conf/iclr/ShuXS18}; D. Xu and Fekri \citeyear{DBLP:journals/corr/abs-2106-11417}} \rule{0pt}{2.6ex} \\
         & & {\small Chang et al. \citeyear{chang2020coactive}; Lyu et al. \citeyear{DBLP:conf/aaai/LyuYLG19}; B. Wu et al. \citeyear{DBLP:conf/atal/WuGK19} } \\
         & & {\small Beyret et al. \citeyear{DBLP:conf/iros/BeyretSF19}; Jin et al. \citeyear{DBLP:conf/aaai/JinMJZCY22}; Rossum et al. \citeyear{DBLP:journals/corr/abs-2104-06630} } \\
         & & {\small Andreas et al. \citeyear{DBLP:conf/icml/AndreasKL17}; Rietz et al. \citeyear{DBLP:journals/nca/RietzMHSWS23}; Tasse et al. \citeyear{DBLP:conf/nips/TasseJR20} } \\
         & & {\small Furelos-Blanco et al. \citeyear{DBLP:journals/jair/Furelos-BlancoL21}; Leonetti et al. \citeyear{DBLP:journals/ai/LeonettiIS16}; Sridharan et al. \citeyear{DBLP:journals/jair/SridharanGZW19} } \\
         & & {\small Eysenbach et al. \citeyear{DBLP:conf/nips/EysenbachSL19}; F. Yang, Lyu, et al. \citeyear{DBLP:conf/ijcai/YangLLG18}; A. Zhang et al. \citeyear{DBLP:conf/icml/ZhangSLSF18}} \\
         & & {\small Driessens and Blockeel \citeyear{driessens2001learning}; Kulkarni et al. \citeyear{DBLP:conf/nips/KulkarniNST16} } \\
         %\cline{2-3}
        \rule{0pt}{3ex}
         & \multirow{4}{*}{Rules {\scriptsize(12)}} & {\small Akrour et al. \citeyear{akrour2019towards}; Hein, Hentschel, et al. \citeyear{DBLP:journals/eaai/HeinHRU17}; J. Huang et al. \citeyear{DBLP:journals/eaai/HuangAY20}} \\
         & & {\small Z. Jiang and Luo \citeyear{DBLP:conf/icml/JiangL19}; Ou et al. \citeyear{DBLP:journals/tfs/OuCWL24}; L. Zhang et al. \citeyear{DBLP:conf/pkdd/ZhangLWT21} } \\
         & & {\small Hazra and Raedt \citeyear{DBLP:conf/pkdd/HazraR23}; Ma et al. \citeyear{DBLP:journals/corr/abs-2103-08228}; Payani and Fekri \citeyear{DBLP:journals/corr/abs-2003-10386} } \\
         & & {\small Dong et al. \citeyear{DBLP:journals/corr/abs-1904-11694}; Song et al. \citeyear{song2022interpretable}; Zimmer et al. \citeyear{DBLP:journals/corr/abs-2102-11529} } \\
        %\cline{2-3}
        \rule{0pt}{3ex}
         & \multirow{4}{*}{Decision Tree {\scriptsize(10)}} & {\small Roth et al. \citeyear{DBLP:journals/corr/abs-1907-01180}; Silva et al. \citeyear{silva2019optimization}; Topin, Milani, et al. \citeyear{DBLP:conf/aaai/TopinMFV21} } \\
         & & {\small Cole et al. \citeyear{cole2003symbolic}; Custode and Iacca \citeyear{DBLP:journals/access/CustodeI23}; Paleja et al. \citeyear{DBLP:conf/rss/PalejaNSRCG22} } \\
         & & {\small Gupta et al. \citeyear{DBLP:conf/aaai/GuptaTB15}; Likmeta et al. \citeyear{DBLP:journals/ras/LikmetaMTGRR20} } \\
         & & {\small Das et al. \citeyear{DBLP:journals/corr/abs-2006-05595}; Ernst et al. \citeyear{DBLP:journals/jmlr/ErnstGW05} } \\
         %\cline{2-3}
         \rule{0pt}{3ex}
         
         & \multirow{2}{*}{Graph {\scriptsize(5)}} & {\small Inala et al. \citeyear{DBLP:conf/iclr/InalaBTS20}; Y. Lin et al. \citeyear{DBLP:journals/ral/LinWUR22}; T. Wang et al. \citeyear{DBLP:conf/iclr/WangLBF18} } \\
         & & {\small Hasanbeig et al. \citeyear{DBLP:conf/aaai/HasanbeigJAMK21}; Varma \citeyear{varma2021interpretable} } \\
         \rule{0pt}{3ex}
         
         & \multirow{2}{*}{Equation {\scriptsize(4)}} & {\small Hein, Udluft, et al. \citeyear{DBLP:conf/gecco/0001UR19}; Landajuela et al. \citeyear{DBLP:conf/icml/LarmaPKSGMPF21} } \\
         & & {\small Ault et al. \citeyear{DBLP:conf/atal/AultHS20}; Maes et al. \citeyear{DBLP:conf/dis/MaesFWE12} } \\
         %\cline{2-3}
         \rule{0pt}{3ex}
         
         & Program {\scriptsize(3)} & {\small Gu et al. \citeyear{DBLP:conf/aaai/GuZLGLZ24}; Trivedi et al. \citeyear{DBLP:conf/nips/TrivediZSL21}; Wilson et al. \citeyear{DBLP:conf/gecco/WilsonCLM18} } \\
         %\cline{2-3}

         %\cline{2-3}
         %& Others {\scriptsize(1)} & \cite{jia2019advanced} \\
         \hline

        \multirow{13}{*}{Surrogate Model {\scriptsize(34)}} & \multirow{6}{*}{Decision Tree {\scriptsize(17)}} & {\small Coppens et al. \citeyear{coppens2019distilling}; Dai, Q. Chen, et al. \citeyear{dai2022enhanced}; W. Guo and Wei \citeyear{guo2022explainable}}  \rule{0pt}{2.6ex} \\
         && {\small Bastani, Pu, et al. \citeyear{DBLP:conf/nips/BastaniPS18}; Dhebar, Deb, et al. \citeyear{DBLP:journals/corr/abs-2009-09521}; Ding et al. \citeyear{DBLP:journals/corr/abs-2011-07553} } \\
         & & {\small Milani, Z. Zhang, et al. \citeyear{DBLP:conf/pkdd/MilaniZTSKPF22}; Sieusahai and Guzdial \citeyear{DBLP:conf/aiide/SieusahaiG21}; Vasic et al. \citeyear{DBLP:journals/nn/VasicPWNSK22} } \\
         & & {\small Acharya et al. \citeyear{DBLP:journals/corr/abs-2011-09004}; Cichosz and Pawelczak \citeyear{DBLP:journals/amcs/CichoszP14}; G. Liu, Sun, et al. \citeyear{DBLP:conf/nips/LiuSSP21} } \\
         & & {\small Carbone \citeyear{carbone2020explainable}; Løver et al. \citeyear{lover2021explainable}; Skirzynski et al. \citeyear{DBLP:journals/ml/SkirzynskiBL21} } \\
         & & {\small Jhunjhunwala et al. \citeyear{DBLP:conf/ijcnn/JhunjhunwalaLSA20}; G. Liu, Schulte, et al. \citeyear{DBLP:conf/pkdd/LiuSZL18} } \\
         %\cline{2-3}
         \rule{0pt}{3ex}
         
         & \multirow{4}{*}{Graph {\scriptsize(9)}} & {\small Domenech i Vila et al. \citeyear{domenech2024explaining}; McCalmon et al. \citeyear{DBLP:conf/atal/McCalmonLA022}} \\ 
         & & {\small Kazhdan et al. \citeyear{DBLP:conf/ijcnn/KazhdanSL20}; Koul et al. \citeyear{DBLP:conf/iclr/KoulFG19}; Vinzent et al. \citeyear{DBLP:conf/aips/VinzentS022} } \\
         & & {\small Danesh et al. \citeyear{DBLP:conf/icml/DaneshKFK21}; Hüyük et al. \citeyear{DBLP:journals/corr/abs-2310-19831}; J. H. Lee \citeyear{DBLP:journals/corr/abs-1901-00188} } \\
         & & {\small Topin and M. Veloso \citeyear{DBLP:conf/aaai/TopinV19}} \\
         %\cline{2-3}
         \rule{0pt}{3ex}
         & \multirow{2}{*}{Program {\scriptsize(5)}} & {\small Kholer, Delfosse, et al. \citeyear{DBLP:journals/corr/abs-2405-14956}; Verma, Murali et al. \citeyear{DBLP:conf/icml/VermaMSKC18}; H. Zhu et al. \citeyear{DBLP:conf/pldi/ZhuXMJ19}} \\
         & & {\small Burke et al. \citeyear{DBLP:conf/rss/BurkePR19}; Verma, H. M. Le, et al. \citeyear{DBLP:conf/nips/Verma0YC19} } \\
         %\cline{2-3}
         \rule{0pt}{3ex}
         & Rules {\scriptsize(2)} &  {\small Nageshrao et al. \citeyear{DBLP:conf/icmla/NageshraoCF19}; Soares et al. \citeyear{DBLP:journals/tfs/SoaresACCNF21} } \\
         %\cline{2-3}
        \rule{0pt}{3ex}
         & Equation {\scriptsize(2)} & {\small Kubalík et al. \citeyear{kubalik2017optimal}; H. Zhang et al. \citeyear{zhang2020interpretable} }  \\
         
        \hline
        \end{tabular}
        \label{tab_ch1:interpretable-policy_works}
    \end{table*}
    
A policy is said to be inherently understandable when it is interpretable by its form. In other words, each decision made by the agent can be explained by looking at the policy, which takes the form of a set of rules or an equation, for example. Unlike the surrogate model approach, which requires learning a second understandable policy that best imitates the first, which is opaque, the inherently understandable approach brings together a set of works that learn a single interpretable policy. We categorise these works according to the form of the policy, which can be a hierarchical policy, rules, a decision tree, a graph, an equation or a program. These different policy formats are presented below, along with some illustrative works. %For a more detailed summary of the various works on inherently understandable policies, please refer to Appendix A.

                \paragraph{Hierarchical Policy.}
                \label{par:HRL}
                % Hierarchical RL
Hierarchical RL (HRL) is a sub-domain of RL consisting in learning a policy which is structured in a hierarchical manner, generally comprising 2 levels, so as to break down the task to be performed into a set of sub-tasks. 
For each sub-task, a \emph{low-level} policy is learned. A \emph{high-level} policy is learned using the various \emph{low-level} policies as an action.
This approach was not originally designed to explain a policy, but to improve the performance and generalisation of policies on tasks that are decomposable. Therefore, in this section we do not pretend to provide an overview of HRL%\todoflo{Il aurait fallu introduire ce domaine en début de paragraphe}
, but rather to present work that has been designed for explainability or that seems relevant to this purpose. 

Of the 20 works in this group (cf. Table~\ref{tab_ch1:interpretable-policy_works}), 12 focus on the interpretability of the policies provided. 
(Andreas et al. \citeyear{DBLP:conf/icml/AndreasKL17}; Y. Jiang et al. \citeyear{DBLP:conf/nips/JiangGMF19}; Rossum et al. \citeyear{DBLP:journals/corr/abs-2104-06630}; Shu et al. \citeyear{DBLP:conf/iclr/ShuXS18}) propose that low-level or high-level policies are explicitly represented using human-understandable symbolic labels or natural language descriptions.
%As an example, for Minecraft games, agents learn to determine when it is necessary to learn a new policy in \cite{DBLP:conf/iclr/ShuXS18}. If not, the agent chooses one of the policies it has already learned. These policies are interpretable  due to their description, such as \emph{`Get object o'}. 
%To learn the dependencies between different tasks/policies (for example, it is necessary to use \emph{`Get object o'} before \emph{`Put object o'}), a stochastic temporal grammar model is learned.
Model-based RL methods use the dynamics of the environment (i.e. the transition function and the reward function) to obtain an agent’s policy. In a model-based HRL context, D. Xu and Fekri (\citeyear{DBLP:journals/corr/abs-2106-11417}) learn symbolic logical transition rules between sub-goals and B. Wu et al. (\citeyear{DBLP:conf/atal/WuGK19}) learn multiple imperfect environment models and use a gating controller to compose modular interpretable policies.
Three works (Chang et al. \citeyear{chang2020coactive}; Jin et al. \citeyear{DBLP:conf/aaai/JinMJZCY22}; Lyu et al. \citeyear{DBLP:conf/aaai/LyuYLG19}) combine RL with planning, where a high-level policy composed of symbolic actions is generated by a planner, and low-level policies are learned for each symbolic action.
Rietz et al. (\citeyear{DBLP:journals/nca/RietzMHSWS23}) combines HRL with reward decomposition (which will be introduced in Section~\ref{sec:reward_decomp}) to obtain both policy-level and action-level explanations. Beyret et al. (\citeyear{DBLP:conf/iros/BeyretSF19}) proposes an approach where the high-level agent iteratively generates sub-goals to guide the low-level agent,  and Tasse et al. (\citeyear{DBLP:conf/nips/TasseJR20}) draws inspiration from a slightly different approach to HRL to obtain a new skill without learning, through the use of Boolean algebra on tasks.

In this paragraph, we mention some HRL methods or methods close to the domain which are not focused on explainability, but which are nonetheless worth mentioning.
(Furelos-Blanco et al. \citeyear{DBLP:journals/jair/Furelos-BlancoL21}; Leonetti et al. \citeyear{DBLP:journals/ai/LeonettiIS16}; Sridharan et al. \citeyear{DBLP:journals/jair/SridharanGZW19}) use logic programming, e.g. Furelos-Blanco et al. mix Inductive Logic Programming (ILP) and RL by learning an automaton composed of edges which represent a sub-goal by a logic formula.
(Eysenbach et al. \citeyear{DBLP:conf/nips/EysenbachSL19}: A. Zhang et al. \citeyear{DBLP:conf/icml/ZhangSLSF18}) build a graph to obtain high-level policies, and (Driessens and Blockeel \citeyear{driessens2001learning}; Kulkarni et al. \citeyear{DBLP:conf/nips/KulkarniNST16}) use variants of Q-learning in a hierarchical context.

We refer the reader to Appendix~\ref{par:App_A_IU_Hierarchical} for more detailed summaries of these works.

                \paragraph{Rules}
The following works represent the policy of an agent by a set of rules.
The 12 works identified (cf. Table~\ref{tab_ch1:interpretable-policy_works}) can be divided according to the type of rules used as policy. These rules are either fuzzy rules or logic-based rules.

A fuzzy rule describes a decision using membership functions that represent the degree to which a state satisfies a condition. 
A total of four works (Akrour et al. \citeyear{akrour2019towards}; Hein, Hentschel, et al. \citeyear{DBLP:journals/eaai/HeinHRU17}; J. Huang et al. \citeyear{DBLP:journals/eaai/HuangAY20}; Ou et al. \citeyear{DBLP:journals/tfs/OuCWL24}) represent policies as if–then rules, where states belong to fuzzy sets through membership functions.
To give an example, a model-based approach is presented in Hein, Hentschel, et al., entitled Fuzzy Particle Swarm RL. The aim is to learn a set of fuzzy rules offline with particle swarm optimization (Kennedy and Eberhart \citeyear{DBLP:conf/icnn/KennedyE95}), using NN's to represent world models. The produced rules take the form: \emph{`if s is m then o'} where $s$ is a state, $m$ a Gaussian membership function and $o$ a real number. The rules are visualised by presenting the membership functions and an example state. The learning of world models has an impact on the production of efficient policies: models that do not approximate the real dynamics correctly can cause the agent to learn the wrong policy. 

A first-order logic rule describes a decision using symbolic predicates and logical relations. It takes the form a clause $a \leftarrow a_1, a_2, \dots a_n$ where $a$ is the head atom (i.e. the action to be performed) and $a_1, a_2, \dots a_n$ the body atoms (i.e. the preconditions to be met).
A total of eight works (Dong et al. \citeyear{DBLP:journals/corr/abs-1904-11694}; Hazra and Raedt \citeyear{DBLP:conf/pkdd/HazraR23}; Z. Jiang and Luo \citeyear{DBLP:conf/icml/JiangL19}; Ma et al. \citeyear{DBLP:journals/corr/abs-2103-08228}; Payani and Fekri \citeyear{DBLP:journals/corr/abs-2003-10386}; Song et al. \citeyear{song2022interpretable}; L. Zhang et al. \citeyear{DBLP:conf/pkdd/ZhangLWT21}; Zimmer et al. \citeyear{DBLP:journals/corr/abs-2102-11529}) represent policies as first-order logic if–then rules.  Most of the works (Hazra and Raedt; Z. Jiang and Luo; Payani and Fekri; L. Zhang et al.) is based on DILP (Evans and Grefenstette \citeyear{DBLP:journals/jair/EvansG18}) to produce interpretable policies. This
neuro-symbolic approach makes it possible to learn a set of symbolic rules weighted by a confidence score, while
still having the advantages of using NNs.
As an example, Deep Explainable Relational RL (Hazra and Raedt) represents an interpretable policy by a set of if-then rules. The objective is to find a policy such that given a state and background knowledge, an action is predicted. An NN is learned to produce rules and semantic constraints are proposed to avoid redundancy between rules.

Refer to Appendix~\ref{par:App_A_IU_Rules} for more detailed summaries of the mentioned works.
                        
                \paragraph{Decision Tree}
The interpretable policy learnt in the following works takes the form of a Decision Tree (DT).
A DT is a hierarchical structure composed of nodes and branches that allow instances to be classified. In our context, instances are the agent's states, and classes correspond in most works to actions. Thus, an internal node in a DT allows the state space to be split into two sub-regions, and a terminal node, called a leaf, associates a state with an action.
The main difference in the work lies in the output, i.e. the leaves of the DT. 

\begin{figure}[]
        \centering
        \includegraphics[width=0.5\linewidth]{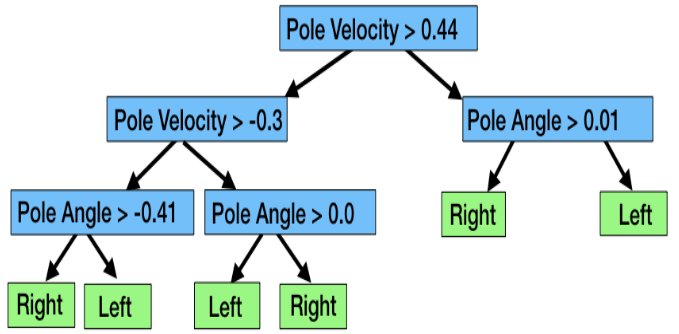}
        \caption[Example of discrete decision tree]{A discrete decision tree for the CartPole environment (Silva et al. \citeyear{silva2019optimization}). Each internal node of the tree (in blue) represents a condition on a feature of the agent's state, and each leaf of the tree (in green) corresponds to an action.}
        \label{fig_ch1:DDT}
    \end{figure}

(Custode and Iacca \citeyear{DBLP:journals/access/CustodeI23}; Roth et al. \citeyear{DBLP:journals/corr/abs-1907-01180}; Silva et al. \citeyear{silva2019optimization}) methods represent Q-values in the leaves, making the DT a structured approximation of the Q-function, from which the policy is derived.
(Gupta et al. \citeyear{DBLP:conf/aaai/GuptaTB15}; Likmeta et al. \citeyear{DBLP:journals/ras/LikmetaMTGRR20}; Silva et al. \citeyear{silva2019optimization}; Topin, Milani et al. \citeyear{DBLP:conf/aaai/TopinMFV21}) methods approximate the policy directly, with leaves representing action distributions or actions.
As an example, Differentiable DT (DDT) is used in Silva et al. The DDT can be used for Q-learning, where each leaf contains the Q-values of the actions, or for a gradient-based approach, where each leaf contains the distribution of the actions. 
The tree is then discretised to obtain an interpretable DT (an example for the CartPole environment (Brockman et al. \citeyear{brockman2016openai}) is shown in Figure \ref{fig_ch1:DDT}).  
In addition, a decision list can be extracted. 
The results show that these trees have a good performance compared to an MLP, while being more interpretable, according to the user study carried out. 
Alongside this work, Paleja et al. (\citeyear{DBLP:conf/rss/PalejaNSRCG22}) is learning about sparse linear controllers (i.e. based on a few features).

Among the 10 papers in this section (cf. Table~\ref{tab_ch1:interpretable-policy_works}), the following four works do not aim for interpretability, but we think it is interesting to mention them, because of their representation of the Q-function as a tree. Only a few works are presented, to give an overview.
Ernst et al. (\citeyear{DBLP:journals/jmlr/ErnstGW05}) approximates the Q-function via ensemble tree methods, (Cole et al. \citeyear{cole2003symbolic}; Das et al. \citeyear{DBLP:journals/corr/abs-2006-05595}) approximates it by leveraging relational predicates and background knowledge (in a Relational MDP setting, described in more detail in Section~\ref{par:RRL_RMDP}), and K. S. Ng (\citeyear{ng2004alkemy}) transforms the approximated Q-function into a compact Boolean representation of the policy.

For more details on this line of work, we refer the reader to Appendix~\ref{par:App_A_IU_DT}.

                \paragraph{Graph}
Five works represent policy using a graph (which is not a decision tree). 
A graph is a structure composed of a set of nodes and edges, allowing relationships between nodes to be represented via edges.
Among these works, (Y. Lin et al. \citeyear{DBLP:journals/ral/LinWUR22}; T. Wang et al. \citeyear{DBLP:conf/iclr/WangLBF18}) represent the policy using a graph structure reflecting the physical or relational structure of the task (robot skeleton or task entities), with interpretability obtained through graph analysis or visualization of learned representations.
(Hasanbeig et al. \citeyear{DBLP:conf/aaai/HasanbeigJAMK21}; Inala et al. \citeyear{DBLP:conf/iclr/InalaBTS20}) represent the agent’s policy as an explicit state machine (automaton), assimilated to a graph, where nodes are akin to sub-policies and transitions encode switching conditions or task progression, yielding interpretable control logic.
Varma (\citeyear{varma2021interpretable}) replaces neural networks with a Tsetlin Machine (Granmo \citeyear{DBLP:journals/corr/abs-1804-01508}) that learns policies as sets of logical clauses. Although symbolic in nature, scalability issues (many clauses required) reduce interpretability.
We refer the reader to Appendix~\ref{par:App_A_IU_Graph} for more details on this line of work.

                \paragraph{Equation}
The policy takes the form of an equation. This form is used when the agent's actions can be expressed as a function of the features of the agent's state.
A total of four works fall into this group (cf. Table~\ref{tab_ch1:interpretable-policy_works}) and are briefly described below.
In a model-based RL setting, the work of Hein et al. \citeyear{DBLP:conf/gecco/0001UR19} uses genetic programming to generate policy function trees, evaluating interpretability via a node-weighted complexity measure.
Landalueja et al. \citeyear{DBLP:conf/icml/LarmaPKSGMPF21} propose a policy generator-evaluator framework: the RNN-based generator produces expression trees for continuous-control policies, trained via reward feedback from the policy evaluator.
In Maes et al. (\citeyear{DBLP:conf/dis/MaesFWE12}), policy search is framed as selecting the best small-sized formula from clustered candidates in a multi-armed bandit setting.  
Assuming that monotonicity over state features ensures interpretability in traffic control tasks, Ault et al. (\citeyear{DBLP:conf/atal/AultHS20}) jointly learns Q-values and a monotonic (regulatable) precedence function.
For more details, refer to Appendix~\ref{par:App_A_IU_Equation}.
                    
                \paragraph{Program}
A policy in the form of a program is proposed in the three following works (cf Table~\ref{tab_ch1:interpretable-policy_works}).
For the remainder of the survey, a program is defined as a symbolic, executable, and understandable policy representation expressed in a programming (or environment-specific) language, composed of control-flow constructs (e.g. conditionals, loops, sequencing) and interpretable primitives.

In order to obtain an interpretable policy based on a domain-specific language, Trivedi et al. (\citeyear{DBLP:conf/nips/TrivediZSL21}) learns a latent embedding space of programs, then searches this space (via Cross Entropy Method (Rubinstein \citeyear{rubinstein1997optimization}) a latent program solving a given task.
In Wilson et al. (\citeyear{DBLP:conf/gecco/WilsonCLM18}), policies are obtained via Cartesian Genetic Programming (Miller and P. Thomson \citeyear{DBLP:conf/eurogp/MillerT00}), where programs are represented as directed graphs of predefined function nodes.
A program expressed in a domain-specific language is constructed in Gu et al. (\citeyear{DBLP:conf/aaai/GuZLGLZ24}) using Monte Carlo Tree Search and refined via Bayesian optimization (Snoek et al. \citeyear{DBLP:conf/nips/SnoekLA12}), resulting in small and interpretable programs.
Refer to Appendix~\ref{par:App_A_IU_Program} for a more detailed summary of this work.

    %------------------------------------------------
    \subsubsection{Surrogate Model}
    %------------------------------------------------

A surrogate model is an interpretable substitute for an agent's opaque policy. It is used to understand the agent's behavior. This section organises the works according to the form taken by the surrogate model.

                \paragraph{Decision Tree}
The following works propose a surrogate model in the form of a DT.
This surrogate model format is the most widely used in the XRL literature, with a total of 17 works identified (cf. Table~\ref{tab_ch1:interpretable-policy_works}).

Most of the work approximates the agent policy with a DT (i.e., the output of the DT is an action), whereas (Jhunjhunwala et al. \citeyear{DBLP:conf/ijcnn/JhunjhunwalaLSA20}; G. Liu, Schulte, et al. \citeyear{DBLP:conf/pkdd/LiuSZL18}) are interested in approximating Q values (the output of the DT is a Q value).
As an example, Jhunjhunwala et al. proposes a new structure, called Q-BSP Tree, which models the Q-value in an interpretable way and an Ordered Sequential Monte Carlo algorithm which learns to approximate the Q-value of the DQN. 
For a problem comprising $n$ actions, the Q-BSP forest is composed of $n$ Q-BSP trees, each learning the Q-value of one action.

In order to make nodes interpretable, two approaches stand out. 
The first approach consists of representing nodes by linear functions on state features, thereby enabling an understanding of the influence of features on the action choice (Carbone \citeyear{carbone2020explainable}; Coppens et al. \citeyear{coppens2019distilling}; Dai, Q. Chen, et al. \citeyear{dai2022enhanced}; W. Guo and Wei \citeyear{guo2022explainable}; Jhunjhunwala et al.; G. Liu, Schulte, et al.; Løver et al. \citeyear{lover2021explainable}).
As an example, Dai et al. propose an \emph{information gain rate weighted oblique DT} (IGR-WODT) to extract an interpretable policy from a DRL agent. Based on three defined metrics, IGR-WODT provides a policy that requires a much smaller number of parameters (only 1.1\% of the number of DRL model parameters). 
An \emph{oblique DT} is composed of internal nodes that separate data based on linear combinations of features (i.e. of the form $w_1f_1 + w_2f_2 + ... + w_nf_n > 0$ where $w_i$ are weights, $f_i$ are state features and $n$ is the number of features).  
The WODT (B. Yang et al. \citeyear{DBLP:conf/aaai/YangSG19}), on which IGR-WODT is based, uses a logistic regression model to determine the probability of an instance belonging to the left or right child node. The extension provided by IGR-WODT consists of using a different objective function inspired by the information gain (Quinlan \citeyear{DBLP:journals/ml/Quinlan86}).
The second approach to making nodes interpretable consists of modifying the representation of states so as to learn a DT based on interpretable features rather than raw features (Acharya et al. \citeyear{DBLP:journals/corr/abs-2011-09004}; G. Liu, Sun, et al. \citeyear{DBLP:conf/nips/LiuSSP21}; Sieusahai and Guzdial \citeyear{DBLP:conf/aiide/SieusahaiG21}; Skirzynski et al. \citeyear{DBLP:journals/ml/SkirzynskiBL21}). These interpretable features can take the form of objects or sprites extracted from an image (G. Liu, Sun, et al.; Sieusahai and Guzdial) or predicates (Acharya et al.; Skirzynski et al.). 
To illustrate, the method proposed by Sieusahai et al. involves learning a DT based not on the image but on a set of \emph{sprites} associated with their position in the image. A sprite is an image representing a character or object within a game. To learn the DT, a function transformation is used to extract a set of sprites with their coordinates from the image representing the state of the DRL agent. The DT learns to approach the DRL agent based on this more compact representation of the agent's state, which makes the internal nodes of the tree interpretable.

Among the works, only one focuses on the multi-agent aspect (Milani, Z, Zhang, et al. \citeyear{DBLP:conf/pkdd/MilaniZTSKPF22}).

For a more detailed summary of the 17 works, please refer to Appendix~\ref{par:App_A_SM_DT}.

                \paragraph{Graph}  
The surrogate models described in this section are an abstract policy represented by a graph.
In most of the work in this section, a node of the graph represents an (abstract) state of the environment, and an edge between two nodes corresponds to an agent's action coupled with an environment transition. 
The aim of the nine works (Danesh et al. \citeyear{DBLP:conf/icml/DaneshKFK21}; Domenech i Vila et al. \citeyear{domenech2024explaining}; Hüyük et al. \citeyear{DBLP:journals/corr/abs-2310-19831}; Kazhdan et al. \citeyear{DBLP:conf/ijcnn/KazhdanSL20}; Koul et al. \citeyear{DBLP:conf/iclr/KoulFG19}; J. H. Lee \citeyear{DBLP:journals/corr/abs-1901-00188}; McCalmon et al. \citeyear{DBLP:conf/atal/McCalmonLA022}; Topin and M. Veloso \citeyear{DBLP:conf/aaai/TopinV19}; Vinzent et al. \citeyear{DBLP:conf/aips/VinzentS022}) (cf. Table~\ref{tab_ch1:interpretable-policy_works}) is to build a graph that zooms out the agent’s policy.

\begin{figure}[]
        \centering
        \includegraphics[width=0.8\linewidth]{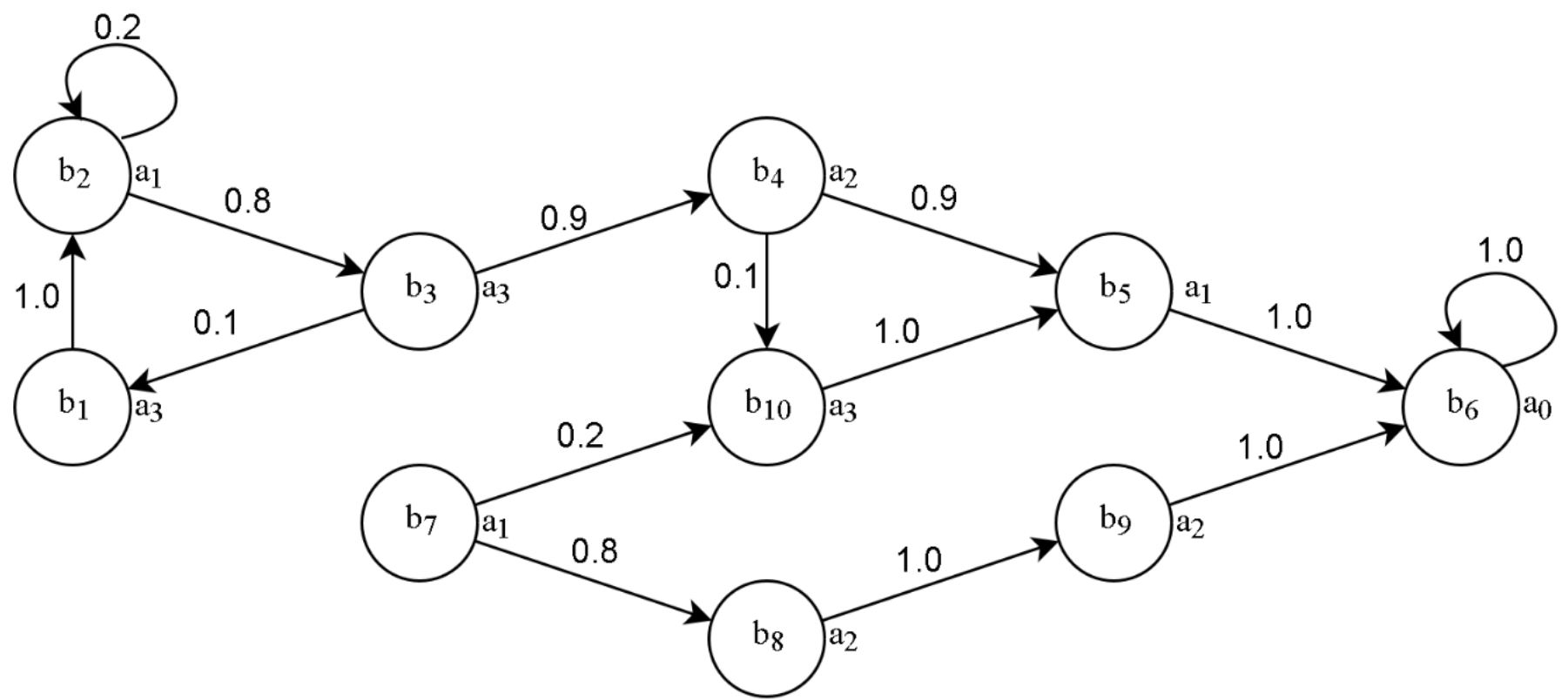}
        \caption[Example of Abstract Policy Graph]{Example of Abstract Policy Graph (Topin and M. Veloso \citeyear{DBLP:conf/aaai/TopinV19}). Each vertex is an abstract state with an associated action, and the edges between two vertices represent transitions (weighted by their probability of occurring).}
        \label{fig_ch1:APG}
    \end{figure}

Five of these works (Domenech i Vila et al.; Kazhdan et al.; McCalmon et al.; Topin and M. Veloso; Vinzent et al.) propose to calculate abstract states to build a graph modelling the agent's high-level policy. The method used to produce abstract states differs: Topin and M. Veloso groups states according to the action performed by the agent from a state and the importance of the state features;  McCalmon et al. calculates abstract states using a clustering algorithm (B. Liu et al. \citeyear{liu2005clustering}) and then describes them using a natural language sentence; (Domenech i Vila et al.; Kazhdan et al.; Vinzent et al.) are based on a set of predefined predicates to determine abstract states. 
To illustrate, Topin and M. Veloso presents \emph{abstracted policy graphs} (APG), which are Markov chains of abstract states. 
In an APG, edges are transitions labelled by their probability of occurring, nodes are abstract states labelled by the agent action performed from that state. An abstract state represents a grouping of similar states that is constructed with respect to the actions and feature importance of the states, calculated via FIRM (Zien et al. \citeyear{DBLP:conf/pkdd/ZienKSR09}). 
The transition probability from node $S$ to node $S'$ is an average of the transition probabilities on the states represented by $S$ and $S'$. To build the APG, a policy, a value function and a set of data on the agent's interactions within the environment are required. An example of APG is described in Figure \ref{fig_ch1:APG}.

Two works (Danesh et al.; Koul et al.) focuses on transforming a policy that takes the form of a Recurrent NN (RNN) into a Moore
Machine (MM) (Moore \citeyear{moore1956gedanken}), which is a finite state machine whose output depends only on the current state and where each state is labelled by an output value (an action in this context).
%As an example, \cite{DBLP:conf/iclr/KoulFG19} first propose an auto-encoder called Quantized Bottleneck Network (QBN), which learns to quantize the agent’s observations and the RNN’s memory states for a MM representation. Next, the QBN is combined with the RNN to create a Moore Machine Network (MMN). Finally, a MM is extracted from the MMN and minimised to be interpretable. In the MM, the nodes represent the states of the memory, and the edges represent the agent’s observations. Each node is labelled with the action to be performed from a memory state. In some environments, it has been observed that the number of memory states and observations to represent the agent’s policy is small.

While most work focuses on explaining policies in a single-agent context, (Domenech i Vila et al.; Kazhdan et al.)  focuses on the multi-agent context.

For further details on the works of this section, we refer the reader to Appendix~\ref{par:App_A_SM_Graph}.

                \paragraph{Program}
Using an environment-specific language, the program is produced based directly on DRL policies in (Verma, H. M. Le, et al. \citeyear{DBLP:conf/nips/Verma0YC19}; Verma, Murali, et al. \citeyear{DBLP:conf/icml/VermaMSKC18}; H. Zhu et al. \citeyear{DBLP:conf/pldi/ZhuXMJ19}). 
Verma, H. M. Le, et al. iteratively search for a policy represented by a neural network using policy-gradient methods and to project the policy into the space of programmatic policies (i.e. policies representable by programs). 
Based on a fixed DRL agent's policy, (Verma, Murali, et al.; H. Zhu et al.) constrain the program's search space in order to provide an interpretable policy within a reasonable time.

An intermediate step is used in (Burke et al. \citeyear{DBLP:conf/rss/BurkePR19}; Kohler, Delfosse, et al. \citeyear{DBLP:journals/corr/abs-2405-14956}) to obtain a program expressed via Python programming language: Kohler, Delfosse, et al. proposes learning a policy by imitation in the form of a regularised oblique tree (Murthy et al. \citeyear{DBLP:journals/jair/MurthyKS94}), then converting it into a program, and Burke et al. calculates a sequence expressed by interpretable low-level motion primitives in a robot manipulation task context, then uses program induction to simplify the sequence into a program.

Refer to Appendix~\ref{par:App_A_SM_Program} for a more detailed description of this work.
Moreover, for a detailed overview of program synthesis for interpreting DRL agents, we recommend Bastani, Inala, et al. (\citeyear{DBLP:conf/icml/BastaniIS20}) to the interested reader. 

                \paragraph{Rules}
Two works (cf. Table~\ref{tab_ch1:interpretable-policy_works}) present a set of if-then rules that act as a surrogate model.
In this work, an if-then rule associates an action or a model that returns an action with a region of the state space. The aggregation of these rules makes it possible to approximate the policy of the DRL agent.
In Nageshrao et al. (\citeyear{DBLP:conf/icmla/NageshraoCF19}), a learned rule represents a region of the state space and outputs an action via a linear model, whereas in Soares et al.  (\citeyear{DBLP:journals/tfs/SoaresACCNF21}), a learned rule associates an action with each state similar to a prototype.

For more details, refer to Appendix~\ref{par:App_A_SM_Rules}.

                \paragraph{Equation}
The substitute policy takes the form of equations in the following two works (cf. Table~\ref{tab_ch1:interpretable-policy_works}). It is in a context where the agent's action is determined as a function of continuous features of the agent's state. Genetic algorithms are used to find such policies: H. Zhang et al. (\citeyear{zhang2020interpretable}) use the evolution feature synthesis (Arnaldo et al. \citeyear{DBLP:conf/gecco/ArnaldoOV15}) and Kubalík et al. (\citeyear{kubalik2017optimal}) a variant of Single Node Genetic Programming (Jackson \citeyear{DBLP:conf/eurogp/Jackson12}).
To obtain concise equations, Kubalík et al. determines a complexity limit (defined by the number of features used in the equation) and H. Zhang et al. proposes a method based on the complexity-performance trade-off.

We refer the reader to Appendix~\ref{par:App_A_SM_Equation} for more details on this work.

\vspace{0.5em}

Table~\ref{tab_ch1:interpretable-policy_works} lists \emph{Interpretable Policy} works. With a total of 88 works mentioned (including 34 in \emph{Surrogate Model} and 54 in \emph{Inherently Understandable}), \emph{Interpretable Policy} is the most widely used ways of explaining policies. The most commonly used and least commonly used policy forms in the works are decision trees and equations, respectively.
                    
%------------------------------------------------
%------------------------------------------------
\subsection{Policy Summary}
\label{sec:SOTA_policy_summary}
%------------------------------------------------
%------------------------------------------------

        % Blue sky strategy summarization Amir / Amir
In the blue sky papers (O. Amir et al. \citeyear{DBLP:conf/atal/AmirDS18,DBLP:journals/aamas/AmirDS19}), O. Amir \emph{et al.} propose a framework for the Strategy Summarization problem that can be decomposed into three parts: World States Representation, Intelligent States Extraction and Strategy Summary Interface. 
World States Representation is concerned with encoding the state space to make it comprehensible to the user, for example by grouping states into abstract states. 
Intelligent States Extraction presents different directions for extracting states that are useful for the user's purpose. The summary should be of reasonable length but still provide sufficient information to the user. 
Strategy Summary Interface provides considerations for user interface design. In addition, this work presents ideas for methods of evaluating summaries, such as domains or metrics. 

    \begin{table}[h]
        \centering
        \setlength\tabcolsep{5pt}
        \caption{Policy Summary works.}
        \vspace{2mm}
        \begin{tabular}{c | l }
         \textbf{Type} & \textbf{Refs}\\ 
         \hline
         
         \multirow{5}{*}{Sequences {\scriptsize(13)}} & D. Amir and O. Amir \citeyear{DBLP:conf/atal/AmirA18}; Huber, Weitz, et al. \citeyear{DBLP:journals/ai/HuberWAA21}; Septon et al. \citeyear{DBLP:conf/paams/SeptonHAA23} \rule{0pt}{2.6ex} \\ 
         & Amitai, Septon, et al. \citeyear{DBLP:conf/aaai/AmitaiSA24}; Feit et al. \citeyear{DBLP:conf/acsos/FeitMP22}; Sequeira and Gervasio \citeyear{DBLP:journals/ai/SequeiraG20} \\ 
         & S. H. Huang, Held, et al. \citeyear{DBLP:journals/arobots/HuangHAD19}; Lage et al. \citeyear{DBLP:conf/ijcai/LageLDA19,DBLP:conf/atal/LageLDA19}; Sequeira and Gervasio \citeyear{DBLP:conf/xai/SequeiraG23} \\ 
         & Altmann et al. \citeyear{DBLP:journals/corr/abs-2404-03359}; Frost el al. \citeyear{DBLP:journals/corr/abs-2201-12462}; Gottesman et al. \citeyear{DBLP:conf/icml/GottesmanF0PCBD20} \\
        \rule{0pt}{3ex}
         
         \multirow{3}{*}{Critical States {\scriptsize(8)}} & Dao, Huff, et al. \citeyear{DBLP:conf/ssci/DaoHL21}; S. H. Huang, Bhatia, et al. \citeyear{DBLP:conf/iros/HuangBAD18}; Rupprecht et al. \citeyear{DBLP:conf/iclr/0001IP20} \\ 
         & Dao, I. Mishra, et al. \citeyear{DBLP:conf/icmla/DaoML18}; Dinu et al. \citeyear{DBLP:conf/icml/DinuHPDBBAH20}, I. Mishra et al. \citeyear{DBLP:conf/ssci/MishraDL18} \\
         & Kazak et al. \citeyear{DBLP:conf/sigcomm/KazakBKS19}; Sreedharan, Srivatsava, and Kambhampati \citeyear{DBLP:conf/aips/SreedharanSK20} \\
         \rule{0pt}{3ex}
         
         \multirow{2}{*}{SHAP {\scriptsize(6)}} & Sequeira and Gervasio \citeyear{DBLP:conf/xai/SequeiraG23}; Y. Wang et al. \citeyear{DBLP:conf/aaaiss/WangME20}; K. Zhang et al. \citeyear{DBLP:journals/tcss/ZhangZXGG22} \\
         & Angelotti and Díaz-Rodríguez \citeyear{angelotti2023towards}; Beechey et al. \citeyear{DBLP:conf/icml/BeecheySS23}; Heuillet et al. \citeyear{DBLP:journals/cim/HeuilletCR22} \\
         \rule{0pt}{3ex}
         
         Policy Comparison {\scriptsize(2)}  & Amitai and O. Amir \citeyear{DBLP:conf/aaai/AmitaiA22}; Gajcin, Nair, et al. \citeyear{DBLP:journals/corr/abs-2112-09462} \\
        \hline
        \end{tabular}
        \label{tab_ch1:policy-summary_works}
    \end{table}

    %------------------------------------------------
    \subsubsection{Sequences}
    \label{sec:SOTA_summary_sequences}
    %------------------------------------------------

The 13 works (cf. Table~\ref{tab_ch1:policy-summary_works}) in this section provide the user with a summary in the form of a set of sequences, composed of state-action pairs. These sequences are the result of the agent's interaction (via its policy) within the environment and thus reflect its behavior. The main objective of (Altmann et al. \citeyear{DBLP:journals/corr/abs-2404-03359}; D. Amir and O. Amir \citeyear{DBLP:conf/atal/AmirA18}; Amitai, Septon, et al. \citeyear{DBLP:conf/aaai/AmitaiSA24}; Feit et al. \citeyear{DBLP:conf/acsos/FeitMP22}; Frost et al. \citeyear{DBLP:journals/corr/abs-2201-12462}; S. H. Huang, Held, et al. \citeyear{DBLP:journals/arobots/HuangHAD19}; Huber, Weitz, et al. \citeyear{DBLP:journals/ai/HuberWAA21}; Lage et al. \citeyear{DBLP:conf/ijcai/LageLDA19,DBLP:conf/atal/LageLDA19}; Septon et al. \citeyear{DBLP:conf/paams/SeptonHAA23}; Sequeira and Gervasio \citeyear{DBLP:journals/ai/SequeiraG20,DBLP:conf/xai/SequeiraG23}) is to provide a limited number of informative sequences for the user, with the aim of producing an overview of the strengths and/or weaknesses of the agent's policy. 

As an example, the HIGHLIGHTS algorithm (D. Amir and O. Amir) provides a set of sequences that summarise the agent's behavior. To select sequences, D. Amir and O. Amir focus on important states using the notion of \emph{state importance} (Clouse \citeyear{clouse1996integrating}) (which uses the agent's Q-values). From the important states, sequences are created to provide the user with more context. Thus, a sequence includes states that precede the important state, the state itself, and states that follow it. In the same work, the HIGHLIGHTS-DIV algorithm is proposed to provide more varied sequences as a summary.

For a more detailed summary of the various works on policy summaries via sequences, refer to Appendix~\ref{sec:App_A_summary_sequences}.

    %------------------------------------------------
    \subsubsection{Critical States}
    %------------------------------------------------

An agent's policy can be summarised simply as a set of so-called critical states in the agent's interactions with the environment.  The definition of a critical state varies according to the 8 works identified (cf. Table~\ref{tab_ch1:policy-summary_works}), but the overall idea is a state that has a significant impact on the agent's behavior and allows the policy to be understood through examples. Critical states are identified by constructing a dense reward function (Arjona-Medina et al. \citeyear{DBLP:conf/nips/Arjona-MedinaGW19}), using Q-values (S. H. Huang, Bhatia, et al. \citeyear{DBLP:conf/iros/HuangBAD18}; Rupprecht et al. \citeyear{DBLP:conf/iclr/0001IP20}), via compilation and then constructing a sequence of states characterising the policy (Sreedharan, Srivastava, and Kambhampati \citeyear{DBLP:conf/aips/SreedharanSK20}).

\begin{figure*}[]
        \centering
        \includegraphics[width=1.0\linewidth]{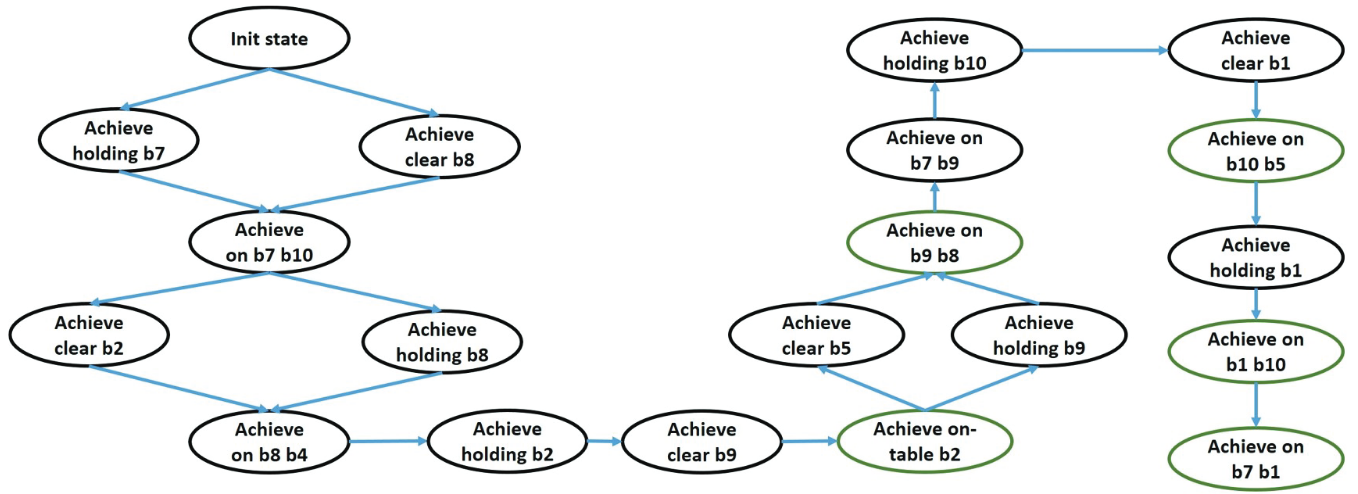}
        \caption[Example of the use of TLdR]{Landmarks and their ordering obtained using TLdR (Sreedharan, Srivastava, and Kambhampati \citeyear{DBLP:conf/aips/SreedharanSK20}) for the exploding blocks-world problem (Younes \citeyear{younesexploding}). Vertices are landmarks and each blue edge defines an order between two landmarks. The green nodes are goals that the agent must reach.}
        \label{fig_ch1:TLdR}
    \end{figure*}

To illustrate this line of work, Sreedharan, Srivastava, and Kambhampati propose a policy summary in the form of milestones, called landmarks, that the agent seeks to reach. The summary is a partial order of landmarks. 
For the exploding blocks-world problem (Younes \citeyear{younesexploding}), an ordering of landmarks is described in Figure \ref{fig_ch1:TLdR}.
The identification of landmarks is done using compilation procedures and the ordering between these landmarks using methods for reasoning about task hierarchies in MDP's.

Other work focuses on summarising learning by identifying the states that had the greatest impact on agent training (Dao, Huff, et al. \citeyear{DBLP:conf/ssci/DaoHL21}; Dao, I. Mishra, et al. \citeyear{DBLP:conf/icmla/DaoML18}; I. Mishra et al. \citeyear{DBLP:conf/ssci/MishraDL18}), and Kazak et al. (\citeyear{DBLP:conf/sigcomm/KazakBKS19}) proposes a state to the user as a counterexample of an unverified requirement in the context of formal NN verification. We refer the reader to Appendix~\ref{sec:App_A_critical_states} for more details on work summarising policy by critical states.

    %------------------------------------------------
    \subsubsection{SHAP}
    \label{sec:policy_SHAP}
    %------------------------------------------------

       % SHAP (Kernel SHAP and DeepSHAP Lundberg) 
This work is based on the model agnostic method, called SHAP (Lundberg and S. Lee \citeyear{DBLP:conf/nips/LundbergL17}). It is interesting to note that this method is mainly used to explain classifiers. 
SHAP consists of determining the impact of each feature of an input on the output of the model. In our context, the input is a state, the output an action, and the model the agent's policy. This method is an approximation to the computation of Shapley values (Shapley et al. \citeyear{shapley1953value}). 
Originally from the field of game theory, Shapley values are used to determine the extent to which a member of a coalition contributes to the final value obtained. To determine Shapley values, it is necessary to calculate for each player $m$ of the game, the set of possible coalitions $c$ containing $m$, and to see the impact of its removal from $c$ on the final value.
SHAP proposes different approximations to Shapley values, which consist in limiting the number of coalition samples. 
Two approximations have been used in the work presented in this survey: Kernel SHAP, a model-agnostic estimation, and DeepSHAP, an estimation specific to neural networks. Kernel SHAP is based on Linear LIME (Ribeiro et al. \citeyear{DBLP:conf/kdd/Ribeiro0G16}) (this method will be described in Section~\ref{sec:SOTA_model-agnostic_approach}) and DeepSHAP on DeepLIFT (Shrikumar et al. \citeyear{DBLP:journals/corr/ShrikumarGSK16}).

\begin{figure}[]
        \centering
        \includegraphics[width=0.6\linewidth]{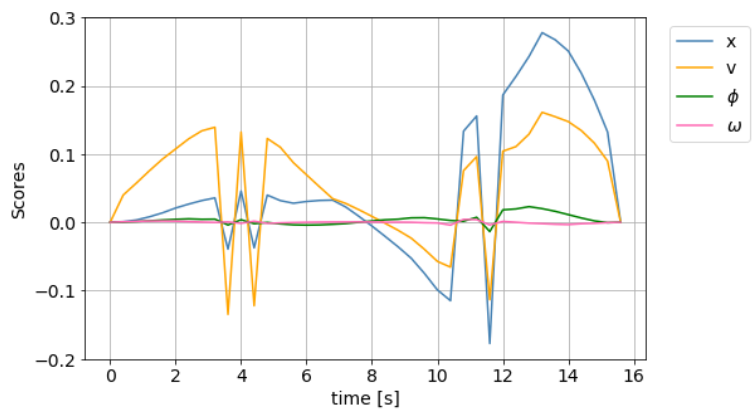}
        \caption[SHAP for a machine control use case]{SHAP for a set of states along an episode for a machine control use case (Y. Wang et al. \citeyear{DBLP:conf/aaaiss/WangME20}). %The base state used for the calculation is the start position, i.e. the initial state. 
        The evolution of the importance score of each feature is represented by a colored curve in the chart, this importance varying over time.}
        \label{fig_ch1:SHAP}
    \end{figure}

As this section is concerned with the explanation of policies, SHAP is used globally: it identifies the impact of each feature on the agent's policy.
To clarify, SHAP values are calculated for each state of an episode, or even a set of episodes of interactions between the agent and the environment. These values are then usually provided to the user in plots, such as the one shown in Figure \ref{fig_ch1:SHAP} from Y. Wang et al. (\citeyear{DBLP:conf/aaaiss/WangME20}). 
Y. Wang et al.; K. Zhang et al. (\citeyear{DBLP:journals/tcss/ZhangZXGG22}) use the method as is (K. Zhang et al. use DeepSHAP and Y. Wang et al. use Kernel SHAP), while other works (Angelotti and Díaz-Rodríguez \citeyear{angelotti2023towards}; Beechey et al. \citeyear{DBLP:conf/icml/BeecheySS23}; Heuillet et al. \citeyear{DBLP:journals/cim/HeuilletCR22}; Sequeira and Gervasio \citeyear{DBLP:conf/xai/SequeiraG23}) calculate SHAP values differently. 
For example, in Sequeira and Gervasio, SHAP values are calculated using XGBoost machines (T. Chen and Guestrin \citeyear{DBLP:conf/kdd/ChenG16}) to simplify the analysis, and in Heuillet et al., they are calculated via Monte Carlo sampling in a cooperative MARL context. For a more detailed summary of the 6 works (cf. Table~\ref{tab_ch1:policy-summary_works}), refer to Appendix~\ref{sec:App_A_policy_SHAP}.

    %------------------------------------------------ 
    \subsubsection{Policy Comparison}
    %------------------------------------------------

        % Compare policies
Two works have been found to compare two policies. This type of method makes it possible to evaluate two agents by identifying the differences in their behavior. Amitai and O. Amir (\citeyear{DBLP:conf/aaai/AmitaiA22}) identifies important states in terms of performance differences between two policies, while Gajcin, Nair, et al. (\citeyear{DBLP:journals/corr/abs-2112-09462}) identifies them in terms of preference differences. We refer the reader to Appendix~\ref{sec:App_A_policy_comparison} for more details. 
    
\vspace{0.5em}

Table~\ref{tab_ch1:policy-summary_works} lists the Policy Summary works. Among the 29 works, the majority propose summaries composed of state-action sequences, and only two works focused on policy comparisons. 

%------------------------------------------------
%------------------------------------------------
\subsection{Human-readable MDP}
%------------------------------------------------
%------------------------------------------------

This section brings together a set of works explaining policy using an environment that is (partially) interpretable by a user. Since the MDP is a tuple composed of a state space, an action space, a transition function and a reward function, the works focus on making one or more elements of the MDP interpretable, either directly (i.e. the component is interpretable by design) or via a surrogate model. 

    \begin{table*}[h]
        \centering
        \setlength\tabcolsep{5pt}
        \caption{Human-readable MDP works.}
        \vspace{2mm}
        \begin{tabular}{ c | c | l }
         \multicolumn{2}{c|}{\textbf{Type}} & \textbf{Refs}\\ 
         \hline
         
          \multirow{13}{*}{Inherently Understandable {\scriptsize(33)}} 
         &  \multirow{8}{*}{MDP Representation {\scriptsize(20)}} & {\small Bica et al. \citeyear{DBLP:conf/iclr/BicaJHS21}; Jenner and Gleave \citeyear{DBLP:journals/corr/abs-2203-13553}; Saldiran et al. \citeyear{saldiran2024towards} } \rule{0pt}{2.6ex}\\
         & & {\small Devidze et al. \citeyear{DBLP:conf/nips/DevidzeRKS21}; Srinivasan and Doshi-Velez \citeyear{srinivasan2020interpretable}} \\
         & & {\small Bewley and Lécué \citeyear{DBLP:conf/atal/BewleyL22}; Camacho et al. \citeyear{DBLP:conf/ijcai/CamachoIKVM19}} \\
         & & {\small Gaon and Brafman \citeyear{DBLP:conf/aaai/GaonB20}; Icarte, Klassen, et al. \citeyear{DBLP:conf/icml/IcarteKVM18}} \\
         & & {\small Kaiser et al. \citeyear{DBLP:conf/esann/KaiserORE19}; Kasenberg and Scheutz \citeyear{DBLP:conf/cdc/KasenbergS17}; X. Li et al. \citeyear{DBLP:conf/iros/LiVB17}} \\
         & & {\small Degris et al. \citeyear{DBLP:conf/icml/DegrisSW06}; Goel et al. \citeyear{DBLP:conf/nips/GoelWP18}; Metzen \citeyear{DBLP:conf/pkdd/Metzen13}} \\
         & & {\small Finkelstein et al. \citeyear{DBLP:conf/aaai/FinkelsteinLKPR22}; Jacq et al. \citeyear{DBLP:journals/corr/abs-2203-08542}; Tabrez and Hayes \citeyear{DBLP:conf/hri/TabrezH19}} \\
         & & {\small Icarte, Waldie, et al. \citeyear{DBLP:conf/nips/IcarteWKVCM19}; Z. Xu et al. \citeyear{DBLP:conf/aips/0005GAMNT020}} \\
        \rule{0pt}{3ex}
         
         & \multirow{5}{*}{Relational RL and MDP {\scriptsize(13)}} & 
         {\small Diuk et al. \citeyear{DBLP:conf/icml/DiukCL08}; Marom and Rosman \citeyear{DBLP:conf/nips/MaromR18}; Scholz et al. \citeyear{DBLP:conf/icml/ScholzLIW14}} \\
         & & {\small Kansky et al. \citeyear{DBLP:conf/icml/KanskySMELLDSPG17}; Veerapaneni et al. \citeyear{DBLP:conf/corl/VeerapaneniC0JF19}; Walker et al. \citeyear{DBLP:conf/ilp/WalkerTSM07}} \\
         & & {\small Martínez et al. \citeyear{DBLP:journals/ai/MartinezAT17}; Walsh \citeyear{walsh2010efficient}; G. Zhu, Z. Huang, et al. \citeyear{DBLP:conf/nips/ZhuHZ18}} \\
         & & {\small Agnew and Domingos \citeyear{agnew2018unsupervised}; G. Zhu, J. Wang, et al. \citeyear{DBLP:conf/aaai/ZhuWRLZ20}} \\
         & & {\small Adjodah et al. \citeyear{adjodah2018symbolic}; Zambaldi et al. \citeyear{DBLP:journals/corr/abs-1806-01830}} \\
         
         \hline
                 \multirow{8}{*}{Surrrogate Model {\scriptsize(16)}} & 
         \multirow{4}{*}{States Clustering {\scriptsize(8)}} & {\small Bekkemoen and Langseth \citeyear{DBLP:conf/acml/BekkemoenL23}; Hayes and Shah \citeyear{DBLP:conf/hri/HayesS17}} \rule{0pt}{2.6ex}\\
         & & {\small Climent et al. \citeyear{DBLP:conf/ccia/ClimentGA21}; Iucci et al. \citeyear{DBLP:conf/icar/IucciHTIL21}}\\
         & & {\small Booth et al. \citeyear{DBLP:conf/ijcai/BoothMS19}; Wollenstein-Betech et al. \citeyear{DBLP:conf/itsc/Wollenstein-Betech20a}} \\
         & & {\small Bewley and Lawry \citeyear{DBLP:conf/aaai/BewleyL21}; Bewley, Lawry, and Richards \citeyear{DBLP:journals/corr/abs-2201-07749}} \\
         \rule{0pt}{3ex}
         
         & \multirow{4}{*}{State Transformation {\scriptsize(8)}} & {\small Garnelo et al. \citeyear{DBLP:journals/corr/GarneloAS16}; Raffin et al. \citeyear{DBLP:journals/corr/abs-1809-09369} } \\
         & & {\small Dai, Ouyang, et al. \citeyear{DBLP:journals/apin/DaiOZLD23}; Garcez et al. \citeyear{DBLP:journals/corr/abs-1804-08597}} \\
         & & {\small François-Lavet et al. \citeyear{DBLP:conf/aaai/Francois-LavetB19}; A. Zhang et al. \citeyear{DBLP:conf/icml/ZhangSLSF18}} \\
         & & {\small J. Chen et al. \citeyear{DBLP:journals/tits/ChenLT22}; Gajcin and Dusparic \citeyear{DBLP:conf/atal/GajcinD22}} \\

         \hline
        \end{tabular}
        \label{tab_ch1:human_readable-MDP_works}
    \end{table*}

    %------------------------------------------------
    \subsubsection{Inherently Understandable}
    %------------------------------------------------
    
The following works directly represent the MDP in order for it to be (partially) interpretable. In comparison with the next section, the MDP is interpretable from its definition. The following works focus on the reward function, the transition function and the state space. 

            \paragraph{MDP Representation.}

This section groups works that represents the MDP, especially the dynamics of the environment (i.e. the reward function and the transition function) in an interpretable way.
11 of the 20 works (cf. Table~\ref{tab_ch1:human_readable-MDP_works}) do not have explainability as their primary objective, but are mentioned because of the interpretable form of the reward and/or transition function proposed. 

\begin{figure}[]
        \centering
        \includegraphics[width=0.5\linewidth]{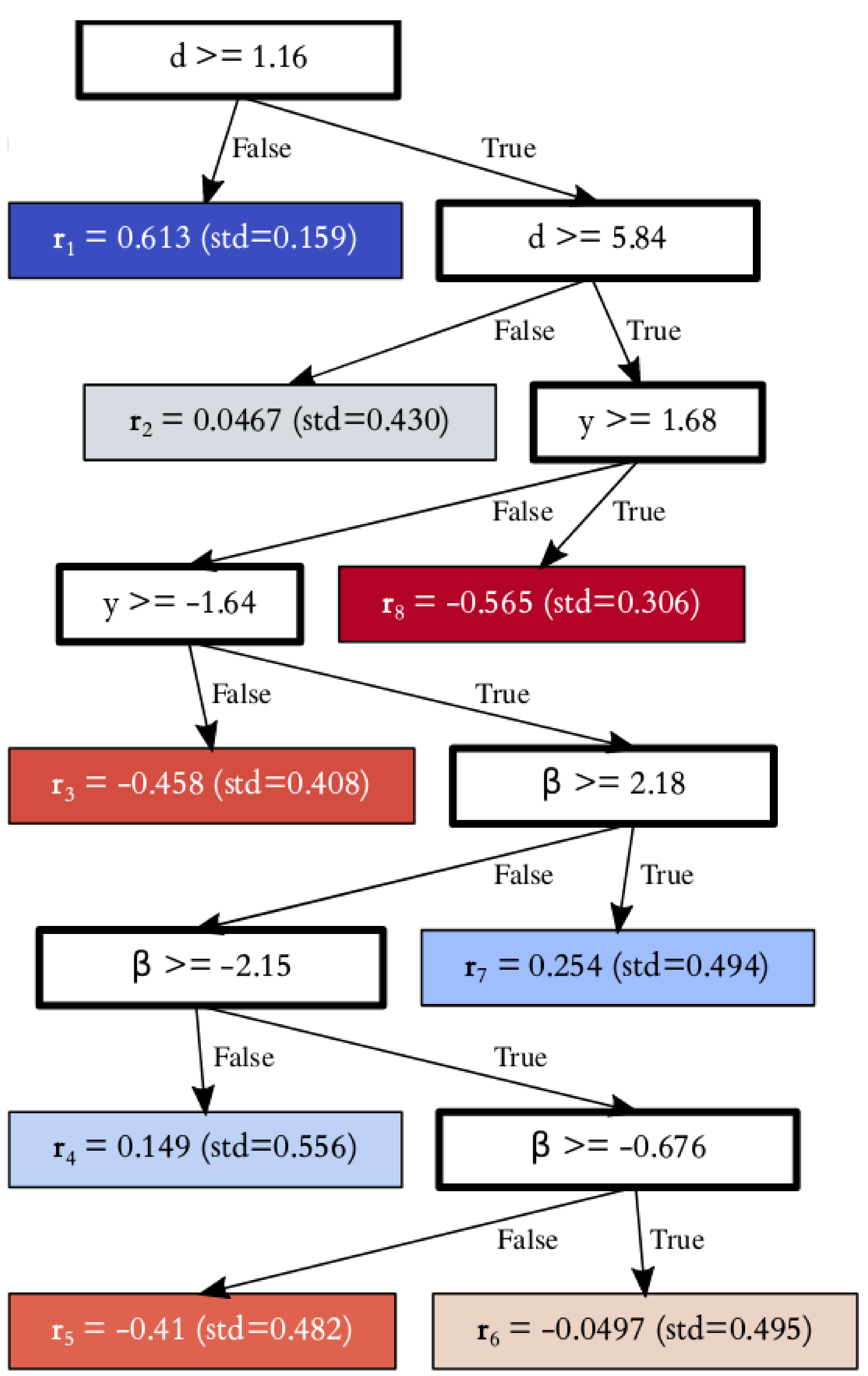}
        \caption[A reward function in the form of a binary tree structure]{A reward function as a binary tree structure for the original RoboCar environment consisting of a 4-wheeled vehicle whose goal is to reach a certain area while avoiding obstacles (Bewley and Lécué \citeyear{DBLP:conf/atal/BewleyL22}). In this reward function, each internal node represents a test on a feature of an agent state, and each leaf a reward. The color of the leaves reflects the amount of reward the agent can obtain (blue means a high amount, red a low one).}
        \label{fig_ch1:reward_function_tree}
    \end{figure}

                % Reward
In (Bewley and Lécué \citeyear{DBLP:conf/atal/BewleyL22}; Bica et al. \citeyear{DBLP:conf/iclr/BicaJHS21}; Camacho et al. \citeyear{DBLP:conf/ijcai/CamachoIKVM19}; Devidze et al. \citeyear{DBLP:conf/nips/DevidzeRKS21}; Gaon and Brafman \citeyear{DBLP:conf/aaai/GaonB20}; Icarte, Klassen, et al. \citeyear{DBLP:conf/icml/IcarteKVM18}; Icarte, Waldie, et al. \citeyear{DBLP:conf/nips/IcarteWKVCM19}; Jenner and Gleave \citeyear{DBLP:journals/corr/abs-2203-13553}; Kasenberg and Scheutz \citeyear{DBLP:conf/cdc/KasenbergS17}; X. Li et al. \citeyear{DBLP:conf/iros/LiVB17}; Saldiran et al. \citeyear{saldiran2024towards}; Srinivasan and Doshi-Velez \citeyear{srinivasan2020interpretable}; Tabrez and Hayes \citeyear{DBLP:conf/hri/TabrezH19}; Z. Xu et al. \citeyear{DBLP:conf/aips/0005GAMNT020}), interpretability is based solely on the reward function. 
Decomposition methods (Bica et al.; Saldiran et al.) propose to express reward as (weighted) interpretable components (reward objectives (Saldiran et al.) or state features (Bica et al.)) whose relative importance can be visualized or analyzed.
To simplify dense rewards into sparse, interpretable ones while preserving optimal policies, transformation methods are proposed (Devidze et al.; Jenner and Gleave).
Tree-based reward models (Bewley and Lécué; Degris et al. \citeyear{DBLP:conf/icml/DegrisSW06}; Srinivasan and Doshi-Velez) represent (part) of the reward function through tree structures. 
Preference-based RL (PbRL) consists of solving a problem modelled by an MDP where the reward function is not provided. A fitness function reflecting the user's preference on pairs of trajectories is used to learn a reward function upstream of the agent's policy. 
As an example of tree-based reward model, Bewley and Lécué propose a PbRL algorithm that represents the interpretable reward function as a binary tree structure where each internal node is a test on a state-action pair and each leaf returns a reward. An example is shown in Figure \ref{fig_ch1:reward_function_tree}.
Although not directly focused on explainability, some works propose representing rewards in an interpretable manner, such as a finite state automaton (Camacho et al.; Gaon and Brafman; Icarte, Klassen, et al.; Icarte, Waldie, et al.; Z. Xu et al.) or a logical specification (Kasenberg and Scheutz; X. Li et al.) capturing the temporal structure of the objective.

                % Transition
The transition function is interpretable in Degris et al.; Goel et al. \citeyear{DBLP:conf/nips/GoelWP18}; Kaiser et al. \citeyear{DBLP:conf/esann/KaiserORE19}; Metzen \citeyear{DBLP:conf/pkdd/Metzen13}.
Kaiser et al. proposes to learn a transition function using a Bayesian approach where the causal sources influencing transitions are explicit, and Goel et al. proposes to learn a transition function in terms of the movement of objects on states that are images.
Precisely, Motion-Oriented REinforcement Learning (MOREL) (Goel et al. \citeyear{DBLP:conf/nips/GoelWP18}) improves the interpretability and, depending on the problem, the sample complexity of the RL algorithm used, by tracking moving objects in states. This approach is tested for a set of 59 Atari Games, where the state of an agent takes the form of an image. 
First, an unsupervised method is applied to learn to predict object and camera motions as well as object masks using a random policy to generate the data. 
Then, the agent uses and refines the previously learned model to learn its policy. Users can view the object masks to understand the agent's policy.
Metzen and Degris et al. represent the function respectively by a graph and a DT.
Finkelstein et al. (\citeyear{DBLP:conf/aaai/FinkelsteinLKPR22}) proposes a contrastive explanation based on a sequence of MDP transformations, and Jacq et al. (\citeyear{DBLP:journals/corr/abs-2203-08542}) proposes a particular MDP that allows for obtaining a sparse policy.

We refer the reader to Appendix~\ref{par:App_A_MDP_Representation} for more detailed summaries of this work.

        \paragraph{Relational RL and MDP}
        \label{par:RRL_RMDP}
        % Merge relational MDP and RL?
        
        % Relational RL (Dzeroski): (non-exhaustive set of papers) (not directly targeted for explanations) (link with inherently interpretable policies)
        % Relational MDP: Guestrin (paper:'Generalizing plans to new environments in relational MDPs')
The following works provide an interpretable representation of states and transitions and are based on works presenting relational MDPs (Guestrin et al. \citeyear{DBLP:conf/ijcai/GuestrinKGK03}) and Relational RL (Dzeroski et al. \citeyear{DBLP:journals/ml/DzeroskiRD01}). 
As a reminder, the objective is to be able to learn a generalizable and efficient policy based on a representation of the environment by a set of objects and relations between them. In this section, we present only an overview of work in these domains.
        % Survey: van Otterlo (logic-based and relational MDP)
For more details, we recommend reading the state of the art Otterlo (\citeyear{DBLP:books/sp/12/Otterlo12}) dating from 2012, which does not therefore include recent work.

Out of the 13 works in this group (cf. Table~\ref{tab_ch1:human_readable-MDP_works}), 7 are not proposed as methods for explainability, although their approach makes the MDP interpretable. Object-Oriented MDPs (Diuk et al. \citeyear{DBLP:conf/icml/DiukCL08}) and their extensions (Marom and Rosman \citeyear{DBLP:conf/nips/MaromR18}; Scholz et al. \citeyear{DBLP:conf/icml/ScholzLIW14}) explicitly encode states as objects and relations and define transitions relationally, improving abstraction and generalisation. In line with this idea, (Kansky et al. \citeyear{DBLP:conf/icml/KanskySMELLDSPG17}; Veerapaneni et al. \citeyear{DBLP:conf/corl/VeerapaneniC0JF19}; Walker et al. \citeyear{DBLP:conf/ilp/WalkerTSM07}; Walsh \citeyear{walsh2010efficient}) propose learning a transition function based on objects (Kansky et al.; Veerapaneni et al.; Walker et al.) or relational action schemas (Walsh). 
To illustrate, in a model-based RL context, Veerapaneni et al. approximate the observation model which represents an image (i.e. agent state) by a set of semantic masks representing objects and the dynamics model which describes the temporal evolution of the different objects for planning.

Among other works (Adjodah et al. \citeyear{adjodah2018symbolic}; Agnew and Domingos \citeyear{agnew2018unsupervised}; Martínez et al. \citeyear{DBLP:journals/ai/MartinezAT17}; Zambaldi et al. \citeyear{DBLP:journals/corr/abs-1806-01830}; G. Zhu, Z. Huang, et al. \citeyear{DBLP:conf/nips/ZhuHZ18}; G. Zhu, J. Wang, et al. \citeyear{DBLP:conf/aaai/ZhuWRLZ20}), (Agnew and Domingos; Zambaldi et al.; G. Zhu, Z. Huang, et al.;  G. Zhu, J. Wang, et al.) also learn to extract from a raw state (usually an image) a set of objects and relations between them in order to learn the transition function.  
For example, G. Zhu, J. Wang, et al. proposes a three-level learning architecture for learning model dynamics from the most to the least abstract. To aid learning, the output of an abstract level is provided to the level below. 
In order from most to least abstract, motion detection detects regions containing dynamic objects from an image sequence, instance segmentation produces coarse masks of dynamic objects and dynamics learning learns the transition function based on agent action and object relationships.

For more details on the works mentioned, refer to Appendix~\ref{par:App_A_RRL_RMDP}.

    %------------------------------------------------
    \subsubsection{Surrogate Model}
    \label{sec:SOTA_human-readable_MDP}
    %------------------------------------------------
    
The following works propose to obtain, from a non-interpretable component of the MDP, in this case the state space, a substitute model of the MDP that is understandable to the user. The 16 works in this section (cf. Table ~\ref{tab_ch1:human_readable-MDP_works}) focus on obtaining interpretable states by performing \emph{state clustering} (8 works) or \emph{state transformation} (8 works). 

            \paragraph{States Clustering.}
            % State clustering
The following work clusters the states to make the MDP interpretable.
The aim of these methods is to obtain, from the state space, groups of states called abstract states that are easy for a user to describe and understand. 
The work uses different approaches to derive these abstract states. For example, Bekkemoen and Langseth (\citeyear{DBLP:conf/acml/BekkemoenL23}) is based on attention maps (a concept described in more detail in Section~\ref{sec:saliency_maps}) and Bewley and Lawry (\citeyear{DBLP:conf/aaai/BewleyL21}) uses a combination of agent action, value function and temporal dynamics. 
Depending on the work, an abstract state takes the form of a natural language description  (Booth et al. \citeyear{DBLP:conf/ijcai/BoothMS19}; Climent et al. \citeyear{DBLP:conf/ccia/ClimentGA21}; Hayes and Shah \citeyear{DBLP:conf/hri/HayesS17}; Wollenstein-Betech et al. \citeyear{DBLP:conf/itsc/Wollenstein-Betech20a}), a set of region limits of a state space (Bewley and Lawry), a representative state and an associated action (Bekkemoen and Langseth), or can be grouped by a tree diagram (Bewley, Lawry, and Richards \citeyear{DBLP:journals/corr/abs-2201-07749}).

\begin{figure*}[h]
        \centering
        \includegraphics[width=1.0\textwidth]{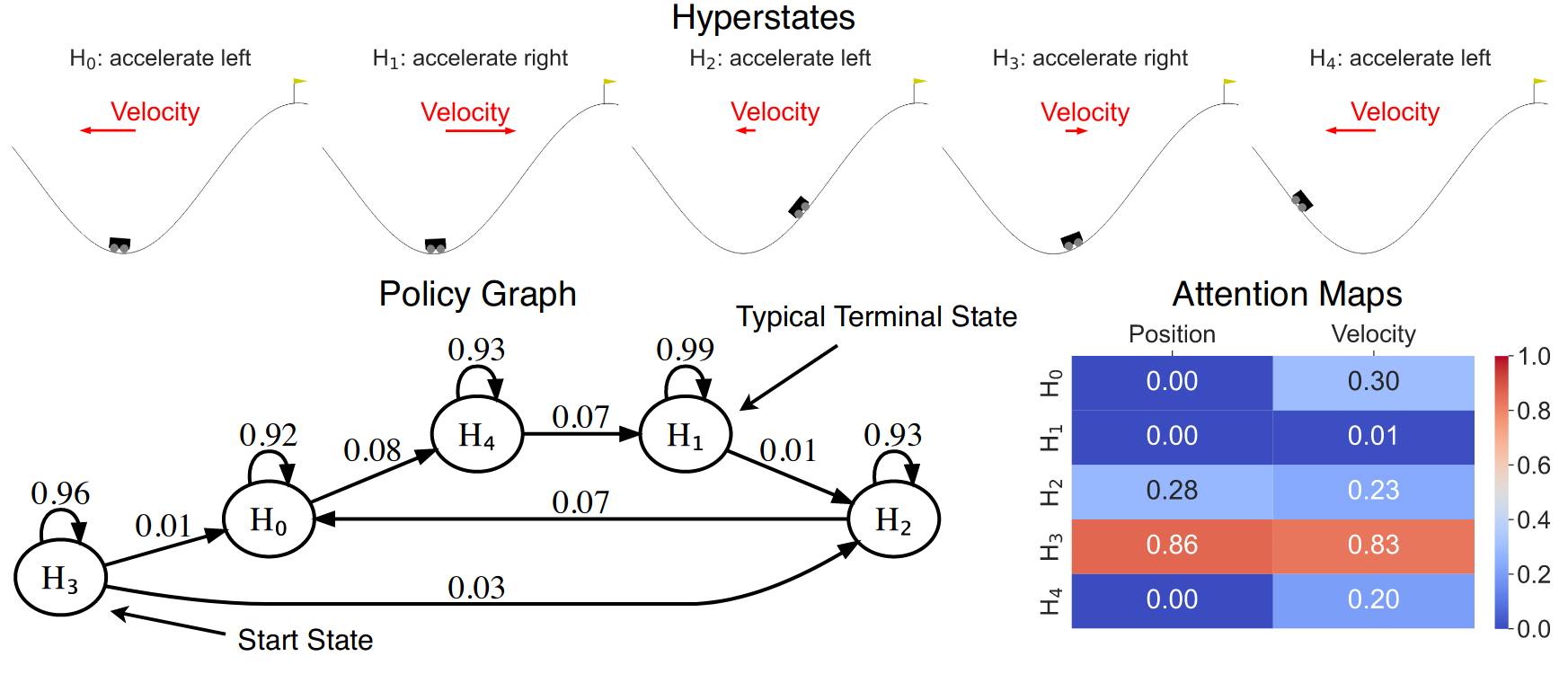}
        \caption[Explanation via ASAP of the agent's policy for the MountainCar environment]{Explanation via ASAP (Bekkemoen and Langseth \citeyear{DBLP:conf/acml/BekkemoenL23}) of the agent's policy for the Mountain Car environment. The first line shows a representative state for each hyperstate with its associated action. The second line is divided into two parts. On the left, a policy graph models the policy based on the hyperstates. Each vertex is a hyperstate and each edge represents a transition between two hyperstates (weighted by its probability of occurrence). On the right, the attention maps of each hyperstate are displayed (where an attention score is associated with each feature, determining the impact of the features on the agent's decision-making).}
        \label{fig_ch1:ASAP}
    \end{figure*}

As an example of this line of work, Bekkemoen and Langseth propose the ASAP method, which groups states into clusters representing abstract states. 
The explanation consists of two parts: a display of a representative state and its associated action for each abstract state identified and a modelling of a policy graph, which is a Markov chain showing the agent's behavior in the abstract state space. 
Abstract states are identified using attention maps (which can be likened to saliency maps).    
ASAP is illustrated with the Mountain Car environment (Brockman et al. \citeyear{brockman2016openai}) in Figure \ref{fig_ch1:ASAP}.

For more details on the work on states clustering, refer to Appendix~\ref{par:App_A_States_Clustering}.

            \paragraph{State Transformation.}% Peut etre à dispatché aussi
            
A state transformation is a function mapping an orginal state, observation or set of observations into an alternative representation whose structure, dimensionality, or semantics improve the interpretability.
The field of State Representation Learning (SRL) consists of learning a low-dimensional representation of the state, which evolves over time and reflects the change induced by the agent's actions.
As stated in the survey Lesort et al. (\citeyear{DBLP:journals/nn/LesortRGF18}), in addition to the learning efficiency aspect, methods in SRL enable low-dimensional states composed of understandable features to be obtained (see Raffin et al. \citeyear{DBLP:journals/corr/abs-1809-09369} for a toolbox of SRL methods).

In line with this field, François-Lavet et al. (\citeyear{DBLP:conf/aaai/Francois-LavetB19}) propose to combine the model-based and model-free approaches of RL algorithms to learn an abstract state representation. 
A specific loss is added for learning the representation of an abstract state so that features of the abstract state are impacted by the agent's actions. This makes the representation interpretable. 
The method is applied in a deterministic environment, but can be adapted for stochastic environments.

The other works convert raw states (which are generally images) into object-level states (Garcez et al. \citeyear{DBLP:journals/corr/abs-1804-08597}; Garnelo et al. \citeyear{DBLP:journals/corr/GarneloAS16}), binary attribute states (A. Zhang et al. \citeyear{DBLP:conf/icml/ZhangSLSF18}) or symbolic image states (J. Chen et al.  \citeyear{DBLP:journals/tits/ChenLT22}). 
Dai, Ouyang, et al. (\citeyear{DBLP:journals/apin/DaiOZLD23}) stands out for its use of prior knowledge to improve interpretability, and Gajcin and Dusparic (\citeyear{DBLP:conf/atal/GajcinD22}) for its use of state transformation for the detection of causal confusion.
For more information, refer to Appendix~\ref{par:App_A_States_Transformation}.
    
\vspace{0.5em}
Table~\ref{tab_ch1:human_readable-MDP_works} lists \emph{Human-readable MDP} works. Of the 49 works mentioned, the 16 \emph{Surrogate Model} methods focus exclusively on modifying the MDP state space, while the 33 \emph{Inherently Understandable} methods focus on the use of transition functions, reward functions, and state spaces that are interpretable by design.
    
%------------------------------------------------
%------------------------------------------------
\subsection{Visual Analysis}
%------------------------------------------------
%------------------------------------------------

This section brings together a body of work focusing on the extraction of agent policy information to be displayed visually to the user during of after the learning phase.% for the user via a visual intermediary.

    \begin{table}[h]
        \centering
        \setlength\tabcolsep{5pt}
        \caption{Visual Analysis works.}
        \vspace{2mm}
        \begin{tabular}{ c | l }
         \textbf{Type} & \textbf{Refs}\\ 
         \hline
         
         \multirow{2}{*}{Visual Toolkit {\scriptsize(5)}} & Jaunet et al. \citeyear{DBLP:journals/cgf/JaunetVW20}; A. Mishra et al. \citeyear{DBLP:journals/corr/abs-2104-02818}; J. Wang et al. \citeyear{DBLP:journals/tvcg/WangGSY19} \\
         & W. He et al. \citeyear{DBLP:conf/apvis/HeLBWS20}; McGregor et al. \citeyear{DBLP:journals/vlc/McGregorBDHMM17} \rule{0pt}{2.6ex} \\
         \rule{0pt}{3ex}

         Latent Space Visualization {\scriptsize(3)} & Annasamy and Sycara \citeyear{DBLP:conf/aaai/AnnasamyS19}; Ben-Zrihem et al. \citeyear{DBLP:journals/corr/ZrihemZM16}; Zahavy et al. \citeyear{DBLP:conf/icml/ZahavyBM16} \\
         \rule{0pt}{3ex}
         
        Inspection {\scriptsize(2)} & Dethise et al. \citeyear{DBLP:conf/sigcomm/DethiseCK19}; Ullauri et al. \citeyear{DBLP:journals/sosym/UllauriGBZZBOY22} \\
        \hline
        \end{tabular}
        \label{tab_ch1:visual-analysis_works}
    \end{table}

    %------------------------------------------------
    \subsubsection{Visual Toolkit}
    %------------------------------------------------

            % User Interface
A user interface including a set of tools for visual analysis of the agent's behavior is proposed in W. He et al. (\citeyear{DBLP:conf/apvis/HeLBWS20}); Jaunet et al. (\citeyear{DBLP:journals/cgf/JaunetVW20}); McGregor et al. (\citeyear{DBLP:journals/vlc/McGregorBDHMM17}); A. Mishra et al. (\citeyear{DBLP:journals/corr/abs-2104-02818}); J. Wang et al. (\citeyear{DBLP:journals/tvcg/WangGSY19}).

\begin{figure*}[h]
        \centering
        \includegraphics[width=1.0\linewidth]{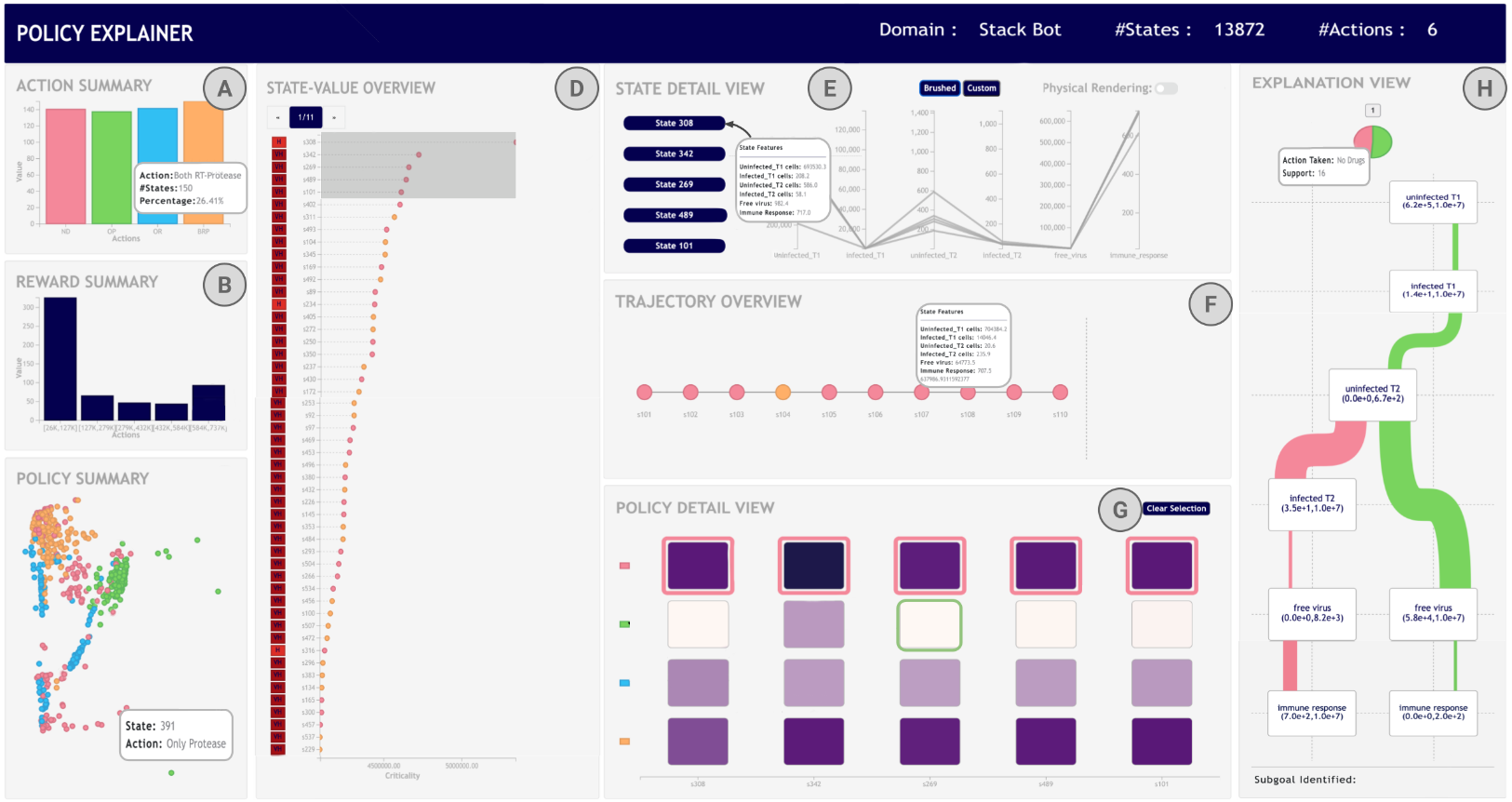}
        \caption[PolicyExplainer visual interface]{PolicyExplainer visual interface (A. Mishra et al. \citeyear{DBLP:journals/corr/abs-2104-02818}). The different tools are divided into three parts. Panels A-C provide a global view of the frequency of actions and rewards, the policy, the value function. Panels D-G provide a detailed analysis of particular states and their associated Q-value, or a particular trajectory. Panel H answers the questions ‘Why?’, ‘Why not?’ and ‘When?’ using state features.}
        \label{fig_ch1:policyExplainer}
    \end{figure*}

                % Policy Explainer Mishra
To illustrate, A. Mishra et al. propose a total of 8 tools to be used to understand the agent's policy. This interface, which is described in Figure \ref{fig_ch1:policyExplainer}, requires the features describing a state to be interpretable. 
PolicyExplainer provides a global view of the frequency of actions and rewards, the policy, the value function  (cf. panels A-C  in Figure \ref{fig_ch1:policyExplainer}) and can be used to look at particular states and their associated Q-value, or a particular trajectory (cf. panels D-G in Figure \ref{fig_ch1:policyExplainer}). In addition, the \emph{`Why?'}, \emph{`Why not?'} and \emph{`When?'} questions are answered by providing the state features that have an impact on the choice of action (cf. panel H in Figure \ref{fig_ch1:policyExplainer}). 

The other visualisation interfaces (W. He et al.; Jaunet et al.; McGregor et al.; J. Wang et al.) are designed for experts, to help them with debugging. 
For more details on these interfaces, we refer the reader to Appendix~\ref{sec:App_A_visual_toolkit}.

    %------------------------------------------------
    \subsubsection{Latent Space Visualization}
    %------------------------------------------------

            % Interpret DQNs
Three works are interested in visually analysing a DQN that has learned to play Atari games in order to understand its internal representation of states.
To do this, the main idea is to cluster the internal representations of states so that they can then be analysed and displayed to the user. 

The first work of this group, Zahavy et al. (\citeyear{DBLP:conf/icml/ZahavyBM16}),  propose to visualise the activations of the DQN by applying t-Distributed Stochastic Neighbor Embedding (t-SNE) (Van der Maaten and G. Hinton \citeyear{van2008visualizing}) to reduce the dimension. 
Before visualisation, a set of data is collected, including hand-crafted features such as the agent's position, so that clusters can be identified and analysed. In addition, saliency maps (a concept described in more detail in Section~\ref{sec:saliency_maps}) are also generated on the different states collected. By analysing the dynamics between clusters, the authors identified that the DQN has learned a hierarchical aggregation of the state space.

\begin{figure*}[h]
        \centering
        \includegraphics[width=1.0\textwidth]{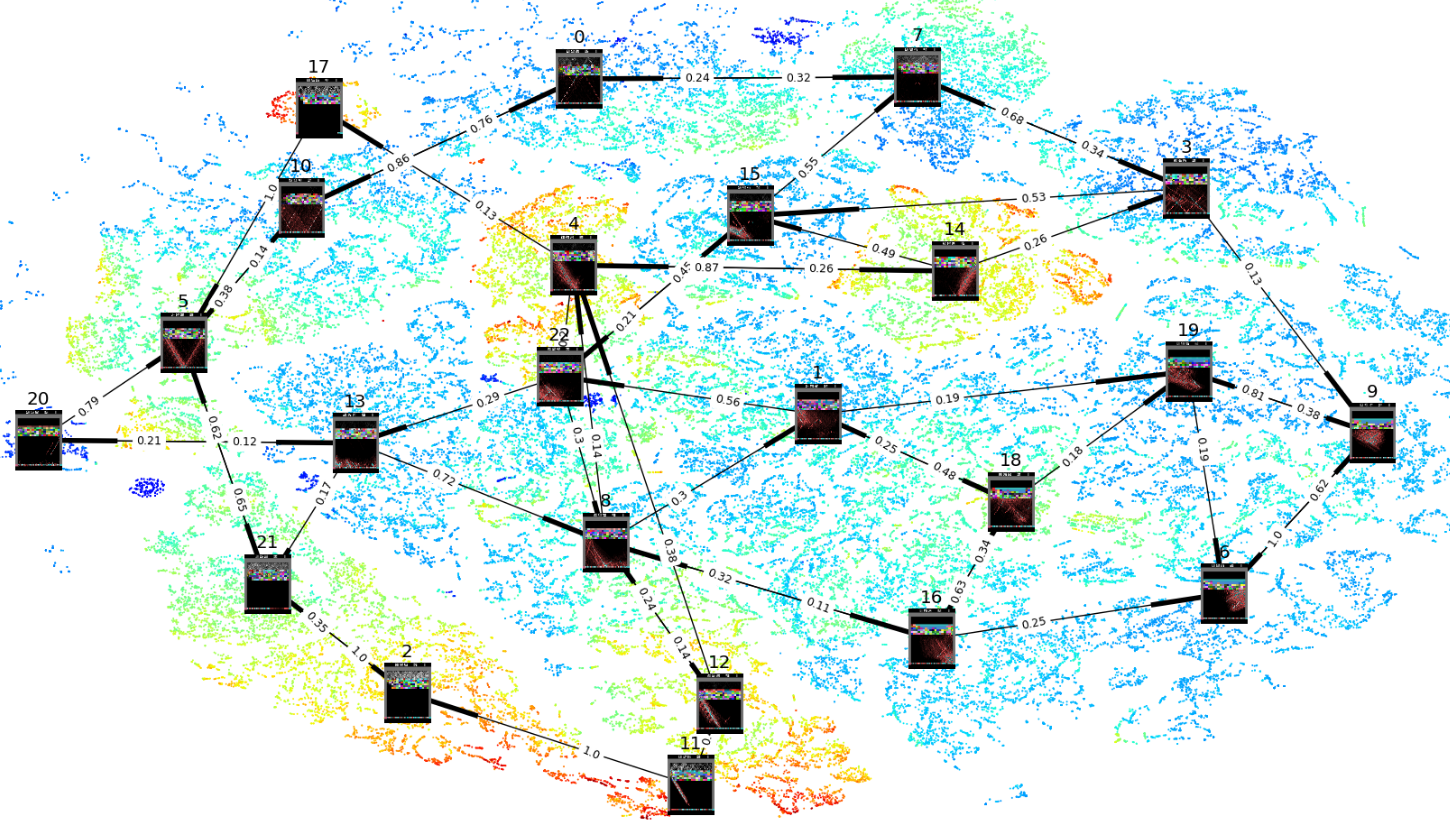}
        \caption[Semi-MDP for the Atari 2600 Breakout environment]{Semi-MDP built on top of a t-SNE map for the Atari 2600 Breakout environment (Ben-Zrihem et al. \citeyear{DBLP:journals/corr/ZrihemZM16}). The t-SNE map is the set of points on which the Semi-MDP is based. It is made up of states (or clusters of MDP states), linked by transitions weighted by their probability of occurrence. Each cluster is modelled by an image of the game where the position of the ball in each state of the cluster is displayed in red.}
        \label{fig_ch1:visual_analysis}
    \end{figure*}

In addition to identifying clusters, Ben-Zrihem et al. (\citeyear{DBLP:journals/corr/ZrihemZM16}) constructs a semi-aggregated MDP that allows transitions between clusters to be obtained and studied, as shown in Figure~\ref{fig_ch1:visual_analysis}.
For more information on (Annasamy and Sycara \citeyear{DBLP:conf/aaai/AnnasamyS19}; Ben-Zrihem et al.), refer to Appendix~\ref{sec:App_A_interpreting_DQNs}.   

    %------------------------------------------------
    \subsubsection{Inspection}
    \label{sec:policy_inspection}
    %------------------------------------------------

            % Track metrics during training
The next two works simply propose to study the agent's behavior by analysing the metrics and displaying them in the form of plots. 
The ETeMoX framework (Ullauri et al. \citeyear{DBLP:journals/sosym/UllauriGBZZBOY22}) tracks the overall reward, the exploration/exploitation trade-off, and a feature specific to the environment under study during training, and Dethise et al.  (\citeyear{DBLP:conf/sigcomm/DethiseCK19}) focuses, among other metrics, on the probability of action choice and confidence in the choice of actions.
For more details, we refer the reader to Appendix~\ref{sec:App_A_inspection}.
    
\vspace{0.5em}

Table~\ref{tab_ch1:visual-analysis_works} summarises the \emph{Visual Analysis} works. With only 10 works, this ways of explanation is the least studied for explaining policies among the ways identified via our taxonomy.
    
%%%%%%%%%%%%%%%%%%%%%%%%%%%%%%%%%%%%%%%%%%%%%%%%%%%%%%%%%%%%%%%%%%%%%%%%
\section{Sequence-level Methods}
\label{sec:sequence_SOTA}
%%%%%%%%%%%%%%%%%%%%%%%%%%%%%%%%%%%%%%%%%%%%%%%%%%%%%%%%%%%%%%%%%%%%%%%%

    \begin{figure}[h]
        \centering
        \includegraphics[width=0.7\linewidth]{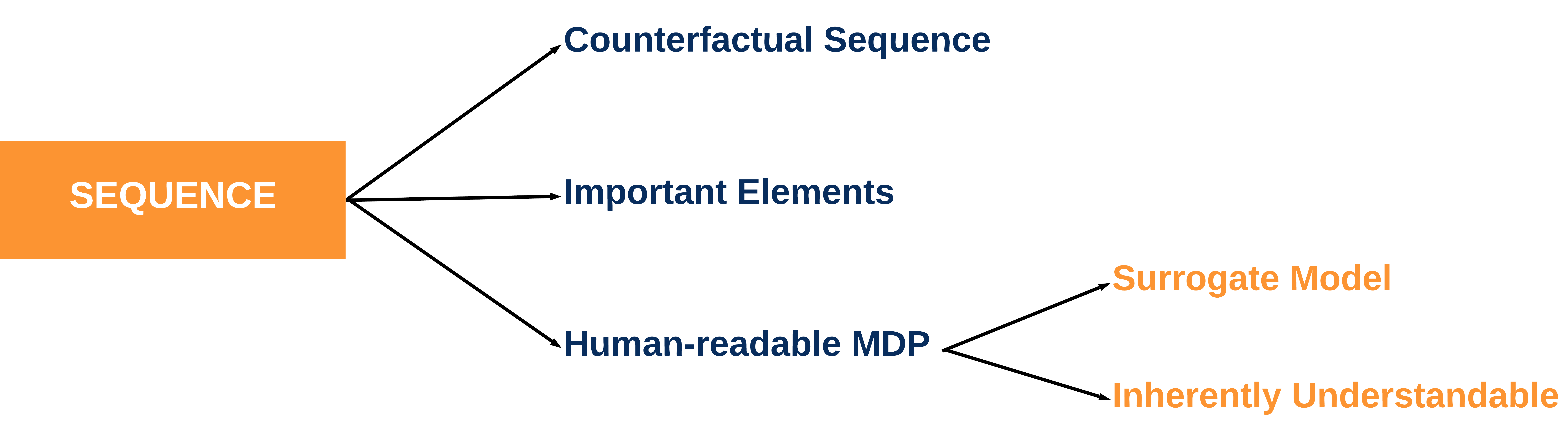}
        \caption[XRL taxonomy: sequence-level methods]{Detailed taxonomy for sequence-level methods.}
        \label{fig_ch1:Taxonomy_sequence}
    \end{figure}
    
The methods described in this section explain sequences of interaction between the agent and the environment. As shown in Figure \ref{fig_ch1:Taxonomy_sequence}, sequences are explained in three ways. 
\emph{Counterfactual Sequences} allow a comparison of the agent's behavior with a sequence that uses an alternative policy, and thus determine the agent's strengths and weaknesses. 
The \emph{Important Elements} of a sequence are those that have the greatest impact on the agent's ability to achieve a given objective. 
The use of \emph{Human-readable MDP} makes it possible to understand, in the context of a sequence, the information available to the agent, the actions and the dynamics of the environment. 
As with the policy explanation category, \emph{Human-readable MDP} are divided into two groups (cf. Figure \ref{fig_ch1:Taxonomy_sequence}): the use of surrogate models to make MDP component(s) interpretable, and the use of MDP component(s) interpretable by design.

    \begin{table*}[h]
        \centering
        \setlength\tabcolsep{5pt}
        \caption{Sequence-level works.}
        \vspace{2mm}
        \begin{tabular}{ c | c | l }
   
         \multicolumn{2}{c|}{\textbf{Type}} & \textbf{Refs}\\ 
         \hline
         \multicolumn{2}{c|}{\multirow{3}{*}{Counterfactual Sequence {\scriptsize (6)}}} & Alabdulkarim and Riedl \citeyear{DBLP:journals/corr/abs-2210-04723}; Waa et al. \citeyear{DBLP:journals/corr/abs-1807-08706} \\
         \multicolumn{2}{c|}{} & Gajcin and Dusparic \citeyear{DBLP:journals/corr/abs-2402-06503}; Tsirtsis, De, et al. \citeyear{DBLP:conf/nips/TsirtsisDR21} \\
         \multicolumn{2}{c|}{} & Stefik et al. \citeyear{stefik2021explaining}; Tsirtsis and Rodriguez \citeyear{DBLP:conf/nips/TsirtsisR23}
         \rule{0pt}{2.6ex} \\
         \hline
         \multicolumn{2}{c|}{\multirow{2}{*}{Important Elements {\scriptsize (4)}}} & W. Guo, X. Wu, et al. \citeyear{DBLP:conf/nips/GuoWKX21}; H. Liu et al. \citeyear{DBLP:conf/iccv/LiuZLWFGS23} \rule{0pt}{2.6ex} \\
         \multicolumn{2}{c|}{} & Saulières et al. \citeyear{DBLP:conf/pkdd/SaulieresCS23,DBLP:conf/ecai/SaulieresCS24}\\
         \hline
         \multirow{2}{*}{Human-readable MDP {\scriptsize(3)}} & Surrogate Model (2) & Soni \citeyear{soni2024towards}; Sreedharan, Soni, et al. \citeyear{DBLP:conf/iclr/SreedharanSVSK22} \rule{0pt}{2.6ex} \\ \rule{0pt}{3ex}
         & Inherently Understandable (1) & W. Li et al. \citeyear{DBLP:conf/aaai/LiFANZ20} \\
        \hline
        \end{tabular}
        \label{tab_ch1:sequence-level_works}
    \end{table*}
    
%------------------------------------------------
%------------------------------------------------
\subsection{Counterfactual Sequence}
\label{sec:SOTA_counterfactual_sequence}
%------------------------------------------------
%------------------------------------------------

A total of six works (cf. Table \ref{tab_ch1:sequence-level_works}) aims to compare the state-action sequence of an agent with a counterfactual sequence where one or more actions differ from the agent's policy (Alabdulkarim and Riedl \citeyear{DBLP:journals/corr/abs-2210-04723}; Gajcin and Dusparic \citeyear{DBLP:journals/corr/abs-2402-06503}; Stefik et al. \citeyear{stefik2021explaining}; Tsirtsis, De, et al. \citeyear{DBLP:conf/nips/TsirtsisDR21}; Tristsis and Rodriguez \citeyear{DBLP:conf/nips/TsirtsisR23}; Waa et al. \citeyear{DBLP:journals/corr/abs-1807-08706}).

This counterfactual sequence can be produced based on the user's question (Alabdulkarim and Riedl; Waa et al.), using an evolutionary algorithm (Gajcin and Dusparic) or an algorithm based on dynamic programming (Tsirtsis, De, et al.; Tristsis and Rodriguez). 
The explanation consists of comparing the agent's sequence with the counterfactual sequence according to the rewards obtained (Alabdulkarim and Riedl; Gajcin and Dusparic;  Tsirtsis, De, et al.; Tristsis and Rodriguez; Waa et al.), but also the occurrence of actions (Alabdulkarim and Riedl; Waa et al.), their cost (Alabdulkarim and Riedl; Waa et al.) or even the types of risks (Stefik et al.).

\begin{figure}[h]
        \centering
        \includegraphics[width=0.6\linewidth]{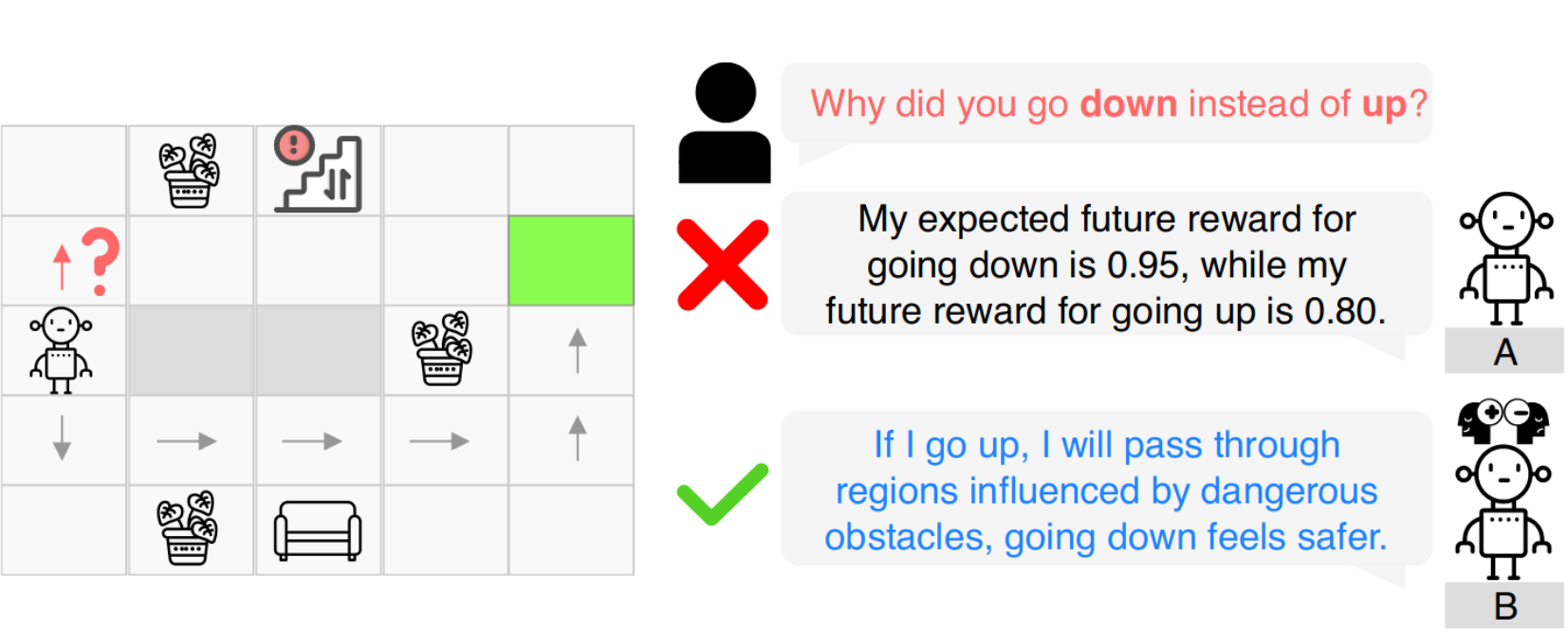}
        \caption[Example of explanation using influence predictors.]{Possible explanations for the sequence of actions that the agent (represented by a robot in cell (3,1)) performs in a 2D environment to reach the green cell (Alabdulkarim and Riedl \citeyear{DBLP:journals/corr/abs-2210-04723}). The user's question is written in red. Explanation A (in black and marked by a red cross) is based solely on the agent's policy, while explanation B (in blue marked with a correct tick) uses influence predictors. According to authors, explanation B is of better quality.}
        \label{fig_ch1:experential}
    \end{figure}

        % Experiential X Alabdulkarim (state and sequence)
To illustrate this line of work, in a context where the environment includes different reward classes, Alabdulkarim and Riedl train a set of influence predictor models for each reward type during agent training. 
The counterfactual sequence is generated by considering the action (or sequence of actions) proposed by the user, then using the agent's policy. 
The sequences are compared using influence predictors (i.e. a calculation of the average influence of each type of reward in each sequence), the cost of the actions and the final reward obtained.
An example showing the interest of this approach is presented in Figure \ref{fig_ch1:experential}.

For further details on the work of the \emph{counterfactual sequence} approach, we refer the reader to Appendix~\ref{sec:App_B_SOTA_counterfactual_sequence}.

    \begin{figure*}[]
        \centering
        \includegraphics[width=1.0\textwidth]{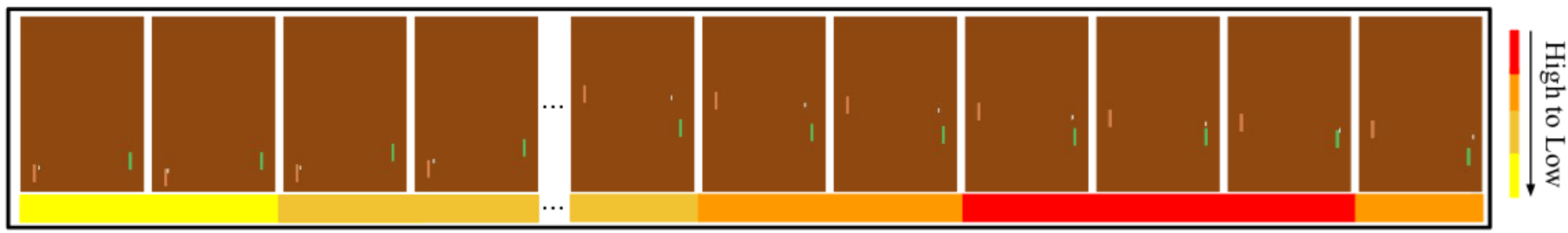}
        \caption[Explanation via EDGE of an episode from a game of Atari 2600 Pong.]{An explanation of an episode from a game of Atari 2600 Pong (W. Guo, X. Wu, et al. \citeyear{DBLP:conf/nips/GuoWKX21}). The objective is to determine the important time-steps for the defeat of the agent controlling the green pad.
        The colored bar below each state of the episode represents its level of importance (yellow represents a low level, orange a medium level and red a high level).}
        \label{fig_ch1:EDGE}
    \end{figure*}

%------------------------------------------------
%------------------------------------------------
\subsection{Important Elements}
\label{sec:SOTA_important_elements}
%------------------------------------------------
%------------------------------------------------

The aim of these methods is to determine, within a given sequence, the most important elements for achieving an objective.
Depending on the work, these important elements are time-steps (W. Guo, X. Wu, et al. \citeyear{DBLP:conf/nips/GuoWKX21}), states (H. Liu et al. \citeyear{DBLP:conf/iccv/LiuZLWFGS23}) or actions (Saulières et al. \citeyear{DBLP:conf/pkdd/SaulieresCS23,DBLP:conf/ecai/SaulieresCS24}). To provide these elements, (W. Guo, X. Wu, et al.; H. Liu et al.) learn a model upstream and (Saulières et al.; Saulières et al.) use the environment functions.

        % EDGE Guo
As an example of this line of work, the EDGE algorithm is a self-explainable model that predicts, for a given episode, the final agent’s reward (W. Guo, X. Wu, et al.). The provided explanation takes the form of a set of important time-steps within the agent’s interaction episode with the environment. 
An explanation of a game of Atari 2600 Pong (Bellemare et al. \citeyear{DBLP:journals/jair/BellemareNVB13}) is shown in Figure \ref{fig_ch1:EDGE}.

For more details, we refer the reader to Appendix~\ref{sec:App_B_SOTA_important_elements}.

%------------------------------------------------
%------------------------------------------------
\subsection{Human-readable MDP}
%------------------------------------------------
%------------------------------------------------

    %------------------------------------------------
    \subsubsection{Surrogate Model}
    %------------------------------------------------

(Soni \citeyear{soni2024towards}; Sreedharan, Soni, et al. \citeyear{DBLP:conf/iclr/SreedharanSVSK22}) propose to obtain, from a non-interpretable element of the MDP, in this case the action space, an interpretable surrogate model. The idea is to represent actions symbolically as a triplet comprising the preconditions, cost and effects of the action. In Soni, an additional task is to determine the type of user that the explainer is dealing with in order to provide a more appropriate explanation.
For more information on these two works, we refer the reader to Appendix~\ref{sec:App_B_SOTA_human-readable_MDP}.
%(i.e. their underlying mental model of the problem)
    
    %------------------------------------------------
    \subsubsection{Inherently Understandable}
    %------------------------------------------------

            % Actions interpretable by design (image reconstruction context) Li
In a context of sequential reconstruction of magnetic resonance images, Li et al. (\citeyear{DBLP:conf/aaai/LiFANZ20}) use RL to successively apply interpretable pixel-wise operations. 
In this environment, the action space is directly interpretable because each action corresponds to common Computer Vision filters, such as the Sobel filter or the Gaussian filter.

\vspace{0.5em}
    % Table description
Table~\ref{tab_ch1:sequence-level_works} summarizes the various works cited for \emph{sequence-level} methods. This target receives the least attention from the XRL community, with only $13$ works identified. Among these methods, the most common ways of explanation is \emph{Counterfactual Sequences}.
    
%%%%%%%%%%%%%%%%%%%%%%%%%%%%%%%%%%%%%%%%%%%%%%%%%%%%%%%%%%%%%%%%%%%%%%%%
\section{Action-level Methods}
\label{sec:action_SOTA}
%%%%%%%%%%%%%%%%%%%%%%%%%%%%%%%%%%%%%%%%%%%%%%%%%%%%%%%%%%%%%%%%%%%%%%%%

    \begin{figure*}[h]
        \centering
        \includegraphics[width=1.0\linewidth]{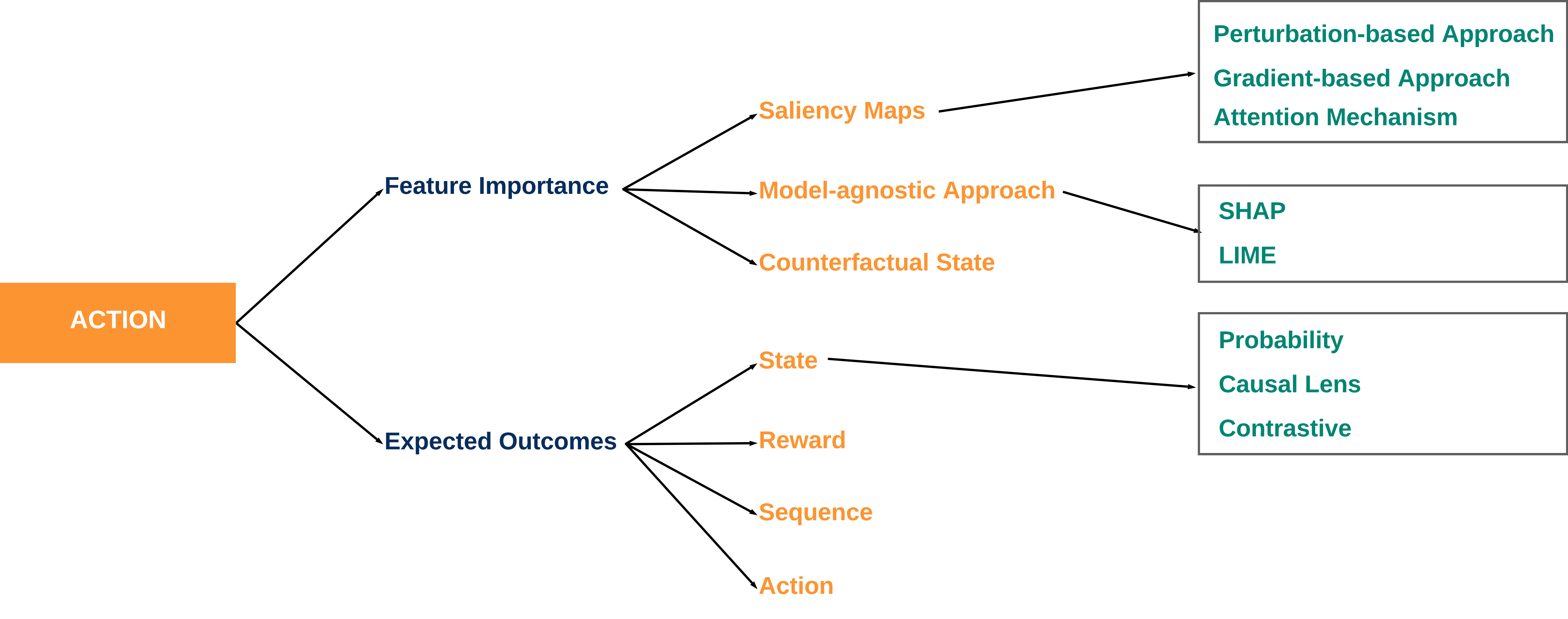}
        \caption[XRL taxonomy: action-level methods]{Detailed taxonomy for action-level methods.}
        \label{fig_ch1:Taxonomy_action}
    \end{figure*}
    
As described in Figure \ref{fig_ch1:Taxonomy_action}, the explanation of actions is carried out using two ways of explanation: \emph{Feature Importance} which determines the features of the agent's state that led it to choose an action, and \emph{Expected Outcomes} which justifies the agent's decision by providing its future potential impact.
\emph{Feature Importance} is divided into three groups of work: feature importance is calculated using saliency maps, with the use of model-agnostic approaches (SHAP (Lundberg and S. Lee \citeyear{DBLP:conf/nips/LundbergL17}) and LIME (Ribeiro et al. \citeyear{DBLP:conf/kdd/Ribeiro0G16})) or via the computation of a counterfactual state. Saliency maps are calculated using three distinct approaches: perturbation-based, gradient-based and attention mechanism.
The nature of \emph{Expected Outcomes} varies according to the work identified (cf. Figure \ref{fig_ch1:Taxonomy_action}): it can be a state, a reward, a state-action sequence or an action.

%------------------------------------------------
%------------------------------------------------
\subsection{Feature Importance}
%------------------------------------------------
%------------------------------------------------

To explain the choice of an action $a$ from a state $s$, the following works describe the importance of the features of $s$ in the agent's choice. 
Most of this work falls into three categories: saliency maps, model-agnostic approaches and counterfactual states.

    \begin{table*}[h]
        \centering
        \setlength\tabcolsep{5pt}
        \caption{Feature Importance works.}
        \vspace{2mm}
        \begin{tabular}{ c | c | l }
         \multicolumn{2}{c|}{\textbf{Type}} & \textbf{Refs}\\ 
         \hline
         \multirow{10}{*}{Saliency Maps {\scriptsize(32)}} 
         & \multirow{6}{*}{Attention Mechanism {\scriptsize(15)}} & {\small Itaya et al. \citeyear{DBLP:conf/ijcnn/ItayaHYFS21}; Mott et al. \citeyear{DBLP:conf/nips/MottZCWR19}; Nikulin et al. \citeyear{DBLP:conf/iccvw/NikulinIAN19}} \rule{0pt}{2.6ex} \\
         & & {\small Y. Liu et al. \citeyear{DBLP:conf/ksem/Liu0CJ22}; Ou et al. \citeyear{DBLP:journals/tfs/OuCWL24}; Shi et al. \citeyear{DBLP:journals/pami/ShiHSWLW22}} \\
         & & {\small Bao et al. \citeyear{DBLP:conf/iccv/Bao0K21}; Josef and Degani \citeyear{DBLP:journals/ral/JosefD20}; Y. Tang et al. \citeyear{DBLP:conf/gecco/TangNH20}} \\
         & & {\small X. Wang, Yuan, et al. \citeyear{DBLP:conf/ro-man/WangYZLS19}; Q. Yang et al. \citeyear{DBLP:journals/tsmc/YangWTSHS24}; Z. Yang et al. \citeyear{DBLP:journals/corr/abs-1812-11276}} \\
         & & {\small Huber, Schiller, et al. \citeyear{DBLP:conf/ki/HuberSA19}; Z. Wang, and S. Wang \citeyear{DBLP:conf/isc2/WangW22}} \\
         & & {\small Q. Zhang \citeyear{DBLP:journals/ral/ZhangMYLYLL21}} \\
         \rule{0pt}{3ex}
         
         & \multirow{4}{*}{Perturbation-based {\scriptsize(10)}} & {\small Douglas et al. \citeyear{DBLP:conf/tabletop/DouglasYKHMT19}; Greydanus et al. \citeyear{DBLP:conf/icml/GreydanusKDF18}; S. Guo et al. \citeyear{DBLP:conf/nips/GuoZLZBHS21}} \\
         && {\small Huber, Limmer, et al. \citeyear{DBLP:journals/frai/HuberLA22}; Puri et al. \citeyear{DBLP:conf/iclr/PuriVGKDK020}; Yan \citeyear{yan2023research}} \\
         & & {\small Iyer et al. \citeyear{DBLP:conf/aies/IyerLL0SS18}; Y. Li et al. \citeyear{DBLP:conf/gcai/LiSI17}} \\
         && {\small Anderson et al. \citeyear{DBLP:journals/tiis/AndersonDSJNICO20}; Persiani and Hellström \citeyear{DBLP:conf/atal/PersianiH22}} \\
         \rule{0pt}{3ex}
         
         & \multirow{3}{*}{Gradient-based {\scriptsize(7)}} & 
         {\small Dao, Huff, et al. \citeyear{DBLP:conf/ssci/DaoHL21}; Joo and K. Kim \citeyear{DBLP:conf/cig/JooK19}; Nie et al. \citeyear{DBLP:conf/ies2/NieHO19}} \\
         & & {\small Hilton et al. \citeyear{hilton2020understanding}; Zahavy et al. \citeyear{DBLP:conf/icml/ZahavyBM16}} \\
         & & {\small L. He et al. \citeyear{he2021explainable}; Weitkamp et al. \citeyear{DBLP:journals/corr/abs-1902-00566}} \\
         
         \hline
         
         \multirow{5}{*}{Model-agnostic Approach {\scriptsize(15)}} 
         & \multirow{4}{*}{SHAP {\scriptsize(13)}} & {\small Carbone \citeyear{carbone2020explainable}; Løver et al. \citeyear{lover2021explainable}; Rizzo et al. \citeyear{DBLP:conf/itsc/RizzoVC19}; Y. Wang et al.  \citeyear{DBLP:conf/aaaiss/WangME20}}  \\
         & & {\small Remman, Strümke, et al. \citeyear{DBLP:conf/amcc/RemmanSL22}; Theumer et al. \citeyear{theumer2022explainable}; K. Zhang et al. \citeyear{DBLP:journals/tcss/ZhangZXGG22}}  \\
         & & {\small Liessner et al. \citeyear{DBLP:conf/icaart/LiessnerDW21}; Remman and Lekkas \citeyear{DBLP:conf/eucc/RemmanL21}; Schreiber et al. \citeyear{schreiber2021towards}} \\
         & & {\small Beechey et al. \citeyear{DBLP:conf/icml/BeecheySS23}; Sequeira and Gervasio \citeyear{DBLP:conf/xai/SequeiraG23}; Xiong et al. \citeyear{DBLP:journals/corr/abs-2402-12685}} \\
         \rule{0pt}{3ex}
         
         & LIME {\scriptsize(3)} & {\small Dethise et al. \citeyear{DBLP:conf/sigcomm/DethiseCK19}; Kotevska et al. \citeyear{DBLP:conf/bigdataconf/KotevskaMKDASZ20}; H. Li et al. \citeyear{DBLP:journals/tdsc/LiWXLG22}} \\
         \hline
         
         \multicolumn{2}{c|}{\multirow{2}{*}{Counterfactual State {\scriptsize(5)}}} & {\small Huber, Demmler, et al. \citeyear{DBLP:conf/atal/HuberDMOA23}; Olson et al. \citeyear{DBLP:journals/ai/OlsonKNLW21}; Samadi et al.  \citeyear{DBLP:journals/corr/abs-2404-18326}} \\
         \multicolumn{2}{c|}{} & {\small Druce et al. \citeyear{DBLP:journals/corr/abs-2106-03775}; Gajcin and Dusparic \citeyear{DBLP:conf/atal/GajcinD24}} \rule{0pt}{2.6ex} \\
        \hline
        
        \multicolumn{2}{c|}{Others {\scriptsize(2)}}  & {\small Davoodi and Komeili \citeyear{DBLP:conf/smc/DavoodiK21}; Elizalde et al. \citeyear{DBLP:conf/micai/ElizaldeSNR09}} \rule{0pt}{2.6ex} \\
        \hline
        \end{tabular}
        \label{tab_ch1:feature-importance_works}
    \end{table*}

    %------------------------------------------------
    \subsubsection{Saliency Maps}
    \label{sec:saliency_maps}
    %------------------------------------------------

            % Saliency map (see paper Atrey for description)
Saliency maps are used when the state of an agent is represented by an image. This method makes it possible to show pixels, or groups of pixels, that are essential in the agent's decision-making. 
In other words, the explanation takes the form of an image (the agent's state) overlaid with a heat map, or saliency maps, representing the (groups of) pixels in the image that are most important for the agent's decision.
The major different ways of calculating them are: gradient-based, perturbation-based and attention mechanisms approaches. 

As this type of approach is fashionable, we do not intend to provide an exhaustive description of the work related to saliency maps in the XRL context. For a detailed analysis of the use of saliency maps for RL agents, see Atrey et al. (\citeyear{DBLP:conf/iclr/AtreyCJ20}). 
Most of the experiments on saliency maps were carried out on Atari2600 games using the Arcade Learning Environment (Bellemare et al. \citeyear{DBLP:journals/jair/BellemareNVB13}). Note that saliency maps can also be referred to as attention maps or attention masks in this section. 

                \paragraph{Attention Mechanism}
This section focuses on methods providing saliency maps based on attention mechanisms.
Attention mechanisms are modules used to make the agent focus on certain parts of the image.
The use of these mechanisms to produce explanations has been identified in 15 works (see Table~\ref{tab_ch1:feature-importance_works}). 

Unlike the vast majority of methods that construct the attention mask using an attention block positioned at the beginning of the NN, (Huber, Schiller, et al. \citeyear{DBLP:conf/ki/HuberSA19}; Y. Liu et al. \citeyear{DBLP:conf/ksem/Liu0CJ22}) propose constructing it using the entire NN.
To illustrate, the Action Region Scoring (ARS) module proposed in Y. Liu et al. is used both to explain the choice of action and to improve learning. In the CNN, at the output of each convolutional layer to which a ReLU activation has been applied, an ARS module is used to identify the important regions and then combined with the current image representation. This helps learning by indicating the regions to focus on. For the explanation part, the saliency map is produced by retrieving the outputs of the ARS modules and combining them.

(Y. Tang et al. \citeyear{DBLP:conf/gecco/TangNH20}; Z. Yang et al. \citeyear{DBLP:journals/corr/abs-1812-11276}) stand out from other works by identifying important regions (groups of pixels) rather than individual pixels, and (Ou et al. \citeyear{DBLP:journals/tfs/OuCWL24}; X. Wang, Yuan, et al. \citeyear{DBLP:conf/ro-man/WangYZLS19}) by adding a complementary explanation to the visual explanation (i.e. the saliency map).
As an example, the approach proposed by X. Wang, Yuan, et al. produces a natural language explanation in addition to an attention map. 
The architecture of the algorithm presented breaks down into three parts: an encoder, which encodes the image into a set of features, an attention mechanism, which returns the salient features and the decoder, which generates a verbal explanation from the salient features. 
%A total of four attention mechanisms have been proposed and compared with the BLEU metric \cite{DBLP:conf/acl/PapineniRWZ02} for the textual explanation. The best of these is called \emph{adaptive attention}, which consists of assigning a weight to the features dynamically using the last word generated by the verbal explanation. 

For a more detailed summary of other work using attention mechanisms to provide saliency maps, refer to Appendix~\ref{par:App_C_attention_mechanism}.

                \paragraph{Perturbation-based Approach}
Works in this section uses a perturbation-based approach to generate saliency maps.
Perturbation-based approaches perturb the input image and evaluate its impact on the agent's policy. 

\begin{figure*}[]
        \centering
        \includegraphics[width=1.0\linewidth]{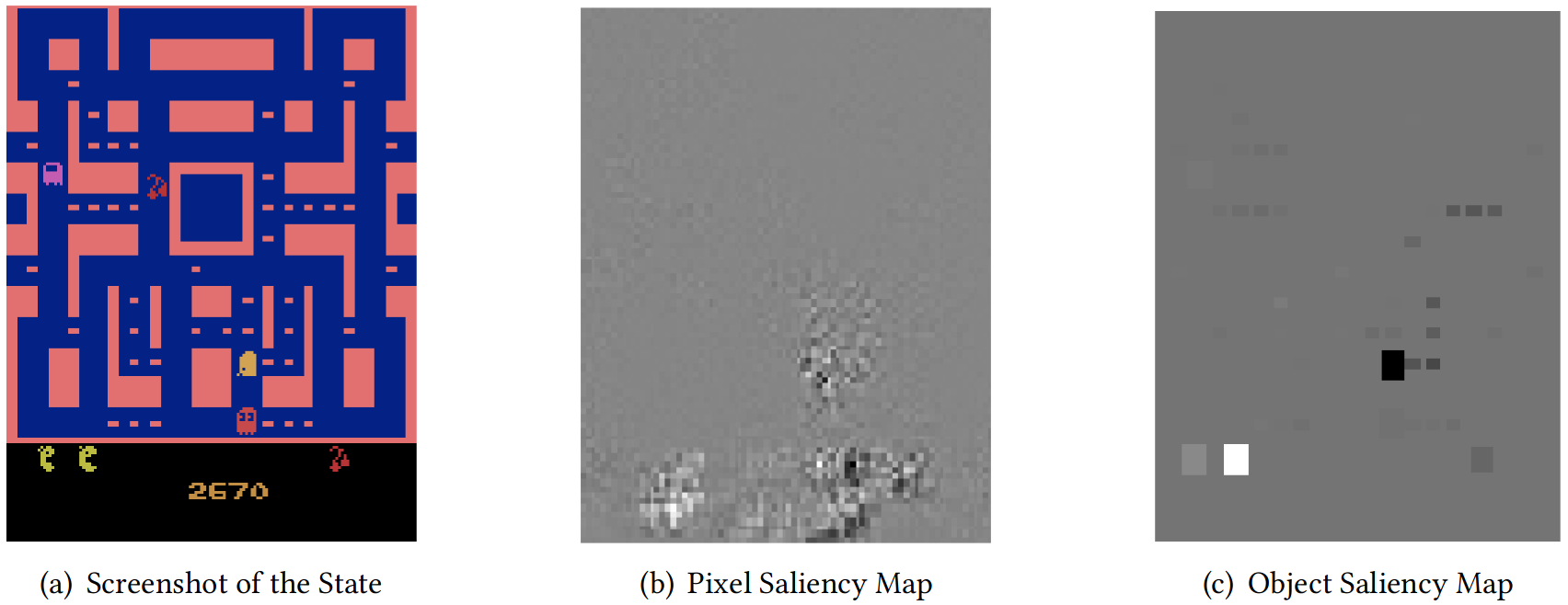}
        \caption[Two saliency maps for a state-action pair from the Ms PacMan environment.]{Two saliency maps for a state-action pair from the Ms PacMan environment (Iyer et al.  \citeyear{DBLP:conf/aies/IyerLL0SS18}). The action is \emph{`move right'} and the sub-figure `a' is the state from which a pixel saliency map (sub-figure `b') and an object saliency map (sub-figure `c') are generated. The saliency of an element is represented using a grey scale, where the darker the element, the more salient it is. In sub-figure `c', we can see that the agent has focused on Ms PacMan and the pills in the right path to choose the \emph{`move right’} action. Conversely, it did not focus on the part of the image that symbolises 1 life for the player, as shown by the white square in the object saliency map.}
        \label{fig_ch1:object_saliency_map}
    \end{figure*}

Some works (Anderson et al. \citeyear{DBLP:journals/tiis/AndersonDSJNICO20}; Iyer et al. \citeyear{DBLP:conf/aies/IyerLL0SS18}; Y. Li et al. \citeyear{DBLP:conf/gcai/LiSI17}) use this approach on groups of pixels identified as objects in the image, as shown in Figure \ref{fig_ch1:object_saliency_map}. 

Various methods are used to perturb images, such as Gaussian blur around a pixel (Douglas et al. \citeyear{DBLP:conf/tabletop/DouglasYKHMT19}; Greydanus et al. \citeyear{DBLP:conf/icml/GreydanusKDF18}; S. Guo et al. \citeyear{DBLP:conf/nips/GuoZLZBHS21}; Persiani and Hellström \citeyear{DBLP:conf/atal/PersianiH22}; Yan \citeyear{yan2023research}) or object masking (Anderson et al.; Iyer et al.; Y. Li et al.).
To evaluate the salience of a pixel or object, methods use the value function and policy (Douglas et al.; Greydanus et al.; S. Guo et al.; Persiani and Hellström), the Q function (Anderson et al.; Iyer et al.; Y. Li et al.; Puri et al. \citeyear{DBLP:conf/iclr/PuriVGKDK020}) or the advantage function (Yan).
As an example of this line of work, the method in Greydanus et al. involves adding spatial uncertainty around the pixel at coordinates $(i,j)$. This is done by interpolating the original image with a Gaussian blur applied around $(i,j)$. 
In practice, instead of applying this perturbation for each pixel, it was applied per group of 5 pixels, which reduces the computational cost and produces good saliency maps. 
Using the A3C learning algorithm, saliency maps are generated and displayed jointly for the agent's policy and for its value function $V$ (i.e. this captures the regions that are important for the agent's choice of action and evaluation of the image).
Three works based on this method (Douglas et al.; S. Guo et al.; Persiani and Hellström).
We refer the reader to Appendix~\ref{par:App_C_perturbation} for more information on the other methods of this line of work.

                \paragraph{Gradient-based approach}
This section focuses on methods providing saliency maps using a gradient-based approach. 
A total of seven works were identified in this line of work (cf. Table~\ref{tab_ch1:feature-importance_works}).
Based on the image, gradient-based approaches use the gradient of a class (i.e. action) in the last layer of the neural network and recover the information with back-propagation.

An example of a well-known method used in certain works (e.g. Joo and K. Kim \citeyear{DBLP:conf/cig/JooK19}; Nie et al. \citeyear{DBLP:conf/ies2/NieHO19}) is Grad-CAM (Selvaraju et al. \citeyear{DBLP:conf/iccv/SelvarajuCDVPB17}). This method can be used on CNN's without having to modify the architecture or re-train the model, in order to create saliency maps. 
To find out the influence of pixels on a class c for a given image, a forward pass is performed, the gradient of class c is calculated, then the signal is propagated with a backward pass. 
Importance weights of the feature maps (where a feature map is the output of a convolutional layer of a CNN) are calculated and then the saliency map is generated by applying a ReLU operation on the weighted linear combination of the feature maps. This method has been used in several works to explain RL agents (Dao, Huff, et al. \citeyear{DBLP:conf/ssci/DaoHL21}, Joo and K. Kim; Nie et al.; Weitkamp et al.  \citeyear{DBLP:journals/corr/abs-1902-00566}).
To illustrate, Joo and K. Kim propose an architecture combining A3C (Mnih, Badia, et al. \citeyear{DBLP:conf/icml/MnihBMGLHSK16}) with Grad-CAM in order to visualise the salient parts of the image for the agent's choice of action.

For a more detailed summary of the work using Grad-CAM (Dao, Huff, et al., Joo and K. Kim; Nie et al.; Weitkamp et al.) and others (L. He et al. \citeyear{he2021explainable}; Hilton et al. \citeyear{hilton2020understanding}; Zahavy et al. \citeyear{DBLP:conf/icml/ZahavyBM16}), refer to Appendix~\ref{par:App_C_gradient}.

    %------------------------------------------------
    \subsubsection{Model-agnostic Approach}
    \label{sec:SOTA_model-agnostic_approach}
    %------------------------------------------------
This section focuses on the use of SHAP (Lundberg and S. Lee \citeyear{DBLP:conf/nips/LundbergL17}) and LIME (Ribeiro et al. \citeyear{DBLP:conf/kdd/Ribeiro0G16}) and their variants to determine the importance of state features in the agent's choice of action.
As a reminder, SHAP is an approximation of shapley values (Shapley et al. \citeyear{shapley1953value})  performed using either Kernel SHAP or DeepSHAP.
            % LIME
LIME estimates the influence that each feature of input $x$ has on the model's output. In our context, the input is a state of the agent, the model the agent's policy and the output an action. 
To calculate the influence of features, a transparent surrogate model is learned locally around $x$. This surrogate model is a linear model. To train it, a set of points is generated by perturbing $x$ and weighted as a function of distance from $x$. 
After training, the weights of the linear model are used to explain the influence of each feature of $x$ on the output of the model.

            % Model Agnostic approaches
            \paragraph{SHAP}
 A total of 13 works were identified to provide explanations for the action performed by the agent via SHAP (cf. Table~\ref{tab_ch1:feature-importance_works}), using either Kernel SHAP (Carbone \citeyear{carbone2020explainable}; Løver et al. \citeyear{lover2021explainable}; Remman, Strümke, et al. \citeyear{DBLP:conf/amcc/RemmanSL22}; Rizzo et al. \citeyear{DBLP:conf/itsc/RizzoVC19}; Y. Wang et al. \citeyear{DBLP:conf/aaaiss/WangME20}) or DeepSHAP (Liessner et al. \citeyear{DBLP:conf/icaart/LiessnerDW21}; Remman and Lekkas \citeyear{DBLP:conf/eucc/RemmanL21}; Schreiber et al. \citeyear{schreiber2021towards}; Theumer et al. \citeyear{theumer2022explainable}; K. Zhang et al.  \citeyear{DBLP:journals/tcss/ZhangZXGG22}).
 For example, Rizzo et al. use Kernel SHAP for a traffic light control problem and K. Zhang et al. use DeepSHAP for a power system emergency control problem.
 
 Other variants of SHAP have been used in (Remman, Strümke, et al.; Xiong et al. \citeyear{DBLP:journals/corr/abs-2402-12685})  to provide explanations, namely Causal SHAP (Heskes et al. \citeyear{DBLP:conf/nips/HeskesSBC20}), TreeSHAP (Lundberg, Erion, et al. \citeyear{DBLP:journals/corr/abs-1802-03888}), and TabularSHAP (Xiong et al.).
 For a detailed description of the work based on SHAP, refer to Appendix~\ref{par:App_C_SHAP}.

                \paragraph{LIME} 
Only three works have been identified that use LIME to provide action explanations (Dethise et al. \citeyear{DBLP:conf/sigcomm/DethiseCK19}; Kotevska et al. \citeyear{DBLP:conf/bigdataconf/KotevskaMKDASZ20}; H. Li et al. \citeyear{DBLP:journals/tdsc/LiWXLG22}).
                
Dethise et al. uses LIME to explain an agent’s decision, or the agent’s global behavior. LIME is also used in H. Li et al. to add an explanatory layer to an advanced persistent threats defense mechanism and in Kotevska et al. to analyse the behavior of an agent on a heating ventilation and air-conditioning control task.

We refer the reader to Appendix~\ref{par:App_C_LIME} for further details.

    %------------------------------------------------
    \subsubsection{Counterfactual State}
    %------------------------------------------------
A body of work aims to provide an answer to the question \emph{`Why does the agent perform action $a$ from state $s$ rather than action $a'$?'} by proposing a counterfactual state $s'$, close to $s$, in which the agent would have chosen $a'$.

Among the five works reviewed (cf. Table ~\ref{tab_ch1:feature-importance_works}), three of them (Huber, Demmler, et al. \citeyear{DBLP:conf/atal/HuberDMOA23}; Olson et al. \citeyear{DBLP:journals/ai/OlsonKNLW21}; Samadi et al. \citeyear{DBLP:journals/corr/abs-2404-18326}) use GANs (Goodfellow et al.  \citeyear{DBLP:journals/corr/GoodfellowPMXWOCB14}) to generate $s'$. 
This architecture consists of jointly learning a generative model $G$ that creates data and a discriminative model $D$ that determines whether the input data is generated or comes from the training dataset. After learning, only the generative model is used.
As an example of work, in order to provide a counterfactual explanation where the state is an image, Olson et al. propose to learn, in addition to these two models, an encoder which encodes the state $s$ in order to omit information from $s$ related to the action choice. 
$D$ is trained to determine the action distribution based on the encoded state. 
$G$ is trained to reconstruct $s$ based on the action distribution and the encoded state. For training, it is necessary to get an initial dataset of state-action pairs using the agent to be explained. This method, tested on Atari2600 games, generates a state $s'$, close to $s$, in which the agent performs a different action.

In Druce et al. (\citeyear{DBLP:journals/corr/abs-2106-03775}), the counterfactual state $s'$ is obtained via predefined changes in $s$ and in Gajcin and Dusparic (\citeyear{DBLP:conf/atal/GajcinD24}) it is obtained via the computation of a sequence of actions leading to $s'$, providing recourse to the user.

For more details on the work addressing the question \emph{`Why does the agent perform action $a$ from state $s$ rather than action $a'$?'}, we refer the reader to Appendix~\ref{sec:App_C_counterfactual_state}.

    \begin{figure*}[]
        \centering
        \includegraphics[width=0.9\linewidth]{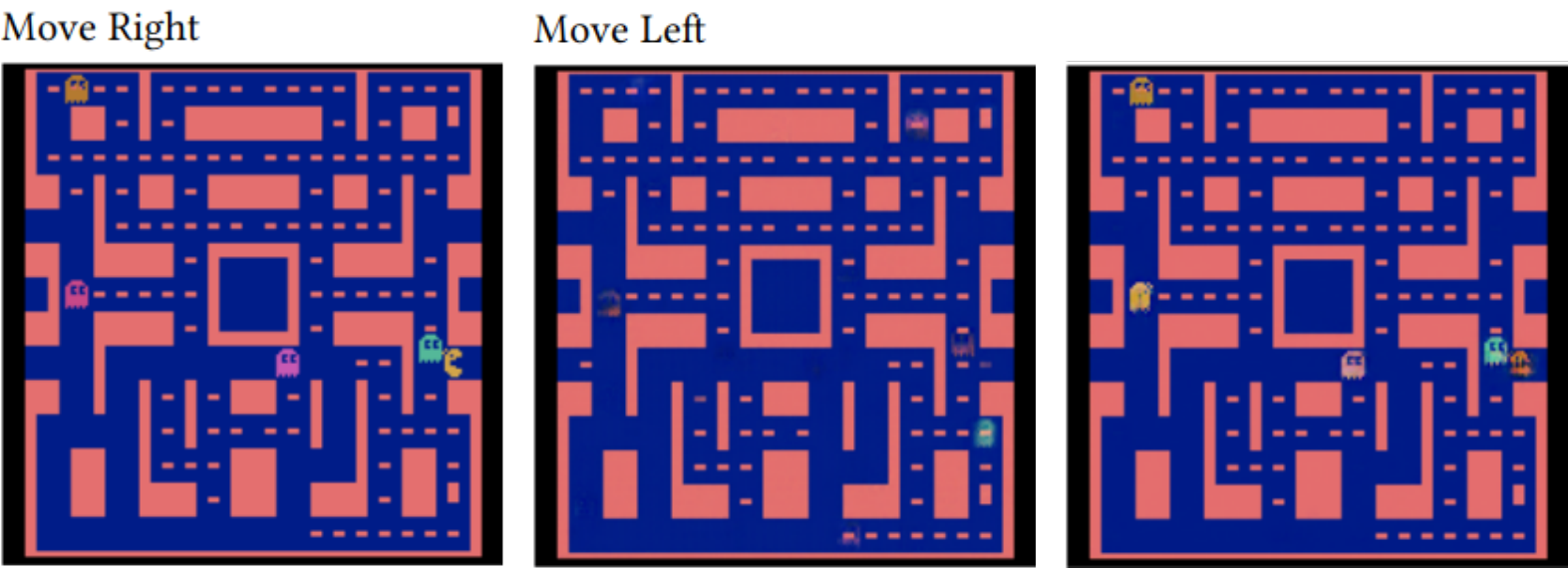}
        \caption[Comparison of counterfactual states in the Ms PacMan environment.]{Comparison of counterfactual states in the Atari 2600 Ms PacMan environment (Bellemare et al. \citeyear{DBLP:journals/jair/BellemareNVB13}) generated on the basis of a state-action pair (on the left) (Huber, Demmler, et al. \citeyear{DBLP:conf/atal/HuberDMOA23}). The agent's action is \emph{‘move right’} and the counterfactual one is \emph{‘move left’}. The state in the middle, where Ms PacMan is missing, is generated using (Olson et al. \citeyear{DBLP:journals/ai/OlsonKNLW21}) and the state on the right, where Ms PacMan is safe in the left part of the map, is generated using Huber, Demmler, et al. The first counterfactual state is not a convincing one as it shows that the agent chooses \emph{‘move left’} while Ms PacMan does not even appear, whereas in the second one, Ms PacMan is (partially) visible.}
        \label{fig_ch1:ganterfactual}
    \end{figure*}

    %------------------------------------------------
    \subsubsection{Others}
    %------------------------------------------------
Two works do not fall into the previously identified work groups (cf. Table~\ref{tab_ch1:feature-importance_works}).
Davoodi and Komeili (\citeyear{DBLP:conf/smc/DavoodiK21}) explain the impact of features in reaching risky states by learning a linear model that estimate feature importance and Elizalde et al. (\citeyear{DBLP:conf/micai/ElizaldeSNR09}) display to the user the most relevant feature according to the value function and provide an associated verbal explanation based on a knowledge-base dataset.

For further details, refer to Appendix~\ref{sec:App_C_others}.

\vspace{0.5em}

Table~\ref{tab_ch1:feature-importance_works} groups together \emph{Feature Importance} works. Of the 54 works, the two most commonly used ways of explaining actions are \emph{saliency maps}, followed by \emph{model-agnostic approaches}. Among these two ways of explanation, the majority of works uses attention mechanisms and SHAP to calculate feature importance. Only five methods propose counterfactual explanations via \emph{counterfactual states}.

%------------------------------------------------
%------------------------------------------------
\subsection{Expected Outcomes}
%------------------------------------------------
%------------------------------------------------

The work presented below focuses on explaining the agent's choice of action by giving and analysing the agent's expected outcomes. 
These works have been categorised according to the type of information provided to the user.
This information can be a (partial) state, a reward, a state-action sequence, or even an action.

\begin{table*}[h]
        \centering
        \setlength\tabcolsep{5pt}
        \caption{Expected Outcomes works.}
        \vspace{2mm}
        \begin{tabular}{ c | c | l }
         \multicolumn{2}{c|}{\textbf{Type}} & \textbf{Refs}\\ 
         \hline
         \multirow{6}{*}{State {\scriptsize(16)}} 

          & Causal Lens {\scriptsize(5)} & Madumal et al. \citeyear{DBLP:conf/aaai/Madumal0SV20,DBLP:journals/corr/abs-2001-10284}; Volodin  \citeyear{volodin2021causeoccam} 
          
          \rule{0pt}{2.6ex} \\
          
          &  & Herlau and Larsen \citeyear{DBLP:conf/aaai/HerlauL22}; Yu et al. \citeyear{DBLP:conf/ijcai/YuRX23}
          \rule{0pt}{2.6ex} \\
          \rule{0pt}{3ex} 
          
         & \multirow{2}{*}{Probability {\scriptsize(4)}} & Cruz, Dazeley, and Vamplew \citeyear{DBLP:conf/ausai/CruzD019}; Cruz, Dazeley, Vamplew, and Moreira \citeyear{DBLP:journals/nca/CruzDVM23} \\
         \rule{0pt}{3ex} 

         & & Dodson et al. \citeyear{DBLP:conf/aldt/DodsonMG11}; Khan et al. \citeyear{DBLP:conf/aips/KhanPB09} \\
         \rule{0pt}{3ex}
         
         & Contrastive {\scriptsize(4)} & Barros et al. \citeyear{DBLP:conf/icdl-epirob/BarrosT0S20}; Z. Lin et al. \citeyear{DBLP:conf/iclr/LinLF21}; Stein \citeyear{DBLP:conf/nips/Stein21}; Tsuchiya et al. \citeyear{DBLP:conf/aaaiss/TsuchiyaME23} \\
         \rule{0pt}{3ex} 
         
         & Others {\scriptsize(3)} & Fukuchi et al. \citeyear{DBLP:conf/iconip/FukuchiOYI17,DBLP:conf/hai/FukuchiOYI17}; Pan et al. \citeyear{DBLP:conf/icra/PanCCCY19} \\
         \hline
         %\multirow{3}{*}{Reward {\scriptsize(9)}} &  &
         \multicolumn{2}{c|}{\multirow{3}{*}{Reward {\scriptsize(10)}}} & Erwig et al. \citeyear{erwig2018explaining}; Iucci et al. \citeyear{DBLP:conf/icar/IucciHTIL21}; Juozapaitis et al. \citeyear{juozapaitis2019explainable}, Saldiran et al. 
         \citeyear{saldiran2024towards} \rule{0pt}{2.6ex} \\
        \multicolumn{2}{c|}{} & Anderson et al. \citeyear{DBLP:journals/tiis/AndersonDSJNICO20}; Feit et al. \citeyear{DBLP:conf/acsos/FeitMP22}; Rietz et al. \citeyear{DBLP:journals/nca/RietzMHSWS23} \\
         \multicolumn{2}{c|}{} & Amitai, Septon, et al. \citeyear{DBLP:conf/aaai/AmitaiSA24}; Kampik et al. \citeyear{DBLP:conf/atal/KampikNL19}; Towers et al. \citeyear{DBLP:conf/ecai/Towers0FN24} \\
         %\cline{2-3}
         \hline
         
        \multicolumn{2}{c|}{Sequence {\scriptsize(5)}}  & Amitai, Septon, et al. \citeyear{DBLP:conf/aaai/AmitaiSA24}; Rouveirol et al. \citeyear{DBLP:conf/rulemlrr/RouveirolASV23}; Yau et al. \citeyear{DBLP:conf/nips/Yau0H20} \rule{0pt}{2.6ex}  \\ 
        \multicolumn{2}{c|}{} & Luss et al. \citeyear{DBLP:conf/aaai/LussDL23}; Saulières et al. \citeyear{DBLP:conf/icaart/SaulieresCS23}
        \rule{0pt}{2.6ex}  \\
        \hline
         \multicolumn{2}{c|}{Action {\scriptsize(1)}}  & Dodson et al. \citeyear{DBLP:conf/aldt/DodsonMG11} \rule{0pt}{2.6ex}  \\
        \hline
        \end{tabular}
        \label{tab_ch1:expected-outcomes_works}
    \end{table*}

    %------------------------------------------------
    \subsubsection{State}
    \label{sec:SOTA_expected_outcomes_state}
    %------------------------------------------------

This line of work is interested in providing the future state(s), or features of the state(s), resulting from an action $a$ from a state $s$, as an explanation.
With 17 works identified, this line of research is the most substantial in terms of number of papers (cf.  Table~\ref{tab_ch1:expected-outcomes_works}). The works explain the choice of an action $a$ from a state $s$ in different ways, such as by calculating the causal link between the pair $(s,a)$ and the features of the next state, by providing the probability of reaching a certain state, and by comparing $a$ with a contrastive action (proposed by the user).
                    
                \paragraph{Causal Lens}
                
The five works (Herlau and Larsen \citeyear{DBLP:conf/aaai/HerlauL22}; Madumal et al. \citeyear{DBLP:conf/aaai/Madumal0SV20,DBLP:journals/corr/abs-2001-10284}; Volodin \citeyear{volodin2021causeoccam}; Yu et al. \citeyear{DBLP:conf/ijcai/YuRX23} propose using causal models to explain the impact of the agent's choice of action on, in particular, the features of the state reached.

\begin{figure}[]
        \centering
        \includegraphics[width=0.6\linewidth]{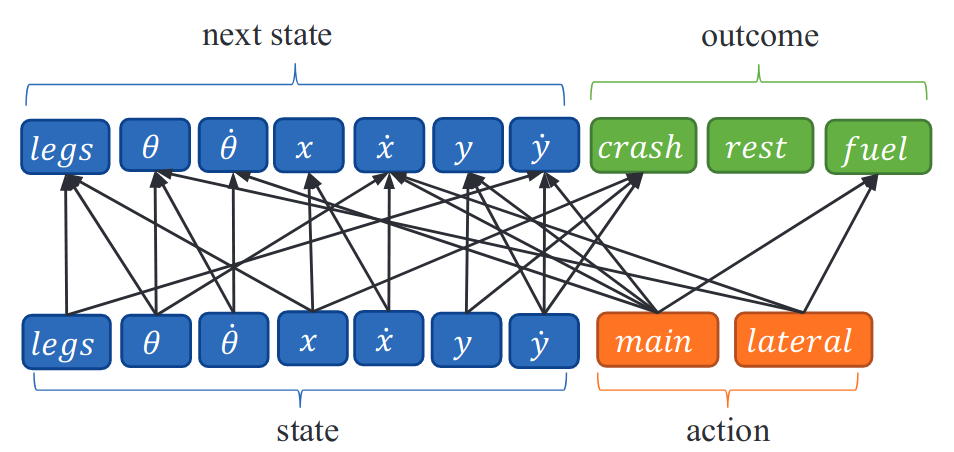}
        \caption[Causal model for the LunarLander environment.]{Causal model for the LunarLander environment (Yu et al. \citeyear{DBLP:conf/ijcai/YuRX23}). The features of a state are in blue, the actions in orange and the outcomes in green. An arrow between two elements describes a causal relation.}
        \label{fig_ch1:AIM}
    \end{figure}

Among these models, (Madumal et al. 2020a,b; Yu et al.) use an Action Influence Model (AIM) which is a structural causal model (Halpern and Pearl \citeyear{DBLP:conf/ijcai/HalpernP01}) with the
addition of action (introduced in Madumal et al. 2020b).
As an example, Yu et al. extend the use of AIM to problems with a continuous action space, without the need of prior knowledge of the environment causal structure. Indeed, to represent the dynamics of the environment, the causal model is learned based on the agent's interactions with the environment. 
An example of a learned causal model for the LunarLander environment (Brockman et al. \citeyear{brockman2016openai}) is shown in Figure \ref{fig_ch1:AIM}.

In this line of work, Madumal et al. 2020a,b use a manually defined causal model, whereas (Herlau and Larsen; Volodin; Yu et al.) learn a sparse model. 

For more details on this work, refer to Appendix~\ref{par:App_C_causal_lens}.

                \paragraph{Probability}
                
The four works (Cruz, Dazeley, and Vamplew \citeyear{DBLP:conf/ausai/CruzD019}; Cruz, Dazeley, Vamplew, and Moreira \citeyear{DBLP:journals/nca/CruzDVM23}; Dodson et al. \citeyear{DBLP:conf/aldt/DodsonMG11}; Khan et al. \citeyear{DBLP:conf/aips/KhanPB09}) (cf. Table~\ref{tab_ch1:expected-outcomes_works}) in this group explain the agent's choice of action with the probability of reaching a certain state or type of states.

To illustrate this line of work, the Memory-based eXplainable Reinforcement Learning approach (Cruz, Dazeley, and Vamplew) provides the probability of success (i.e. reaching a goal state) and the number of transitions to reach the goal state. When the agent is trained, the success probabilities $P$ and the number $T$ of transitions leading to a goal state are calculated and updated for each state.
The agent can then provide answers to the questions \emph{`Why does the agent choose action a?'}, \emph{`Why does the agent choose action a and not b?'} or \emph{`What is the probability of reaching a goal state in 8 steps from s?'}.

(Dodson et al.; Khan et al.) compute the probability of reaching certain types of states and provide explanations in natural language, using predefined templates.

For further details on (Cruz, Dazeley, and Vamplew; Cruz, Dazeley, Vamplew, and Moreira; Dodson et al.; Khan et al.), refer to Appendix~\ref{par:App_C_probability}.

                \paragraph{Contrastive}

This line of work proposes comparing the action performed by the agent with a contrasting action that the agent could have performed. This comparison is made based on the cost of the actions and the probability of reaching a terminal state (Stein \citeyear{DBLP:conf/nips/Stein21}), the confidence of reaching it (Barros et al. \citeyear{DBLP:conf/icdl-epirob/BarrosT0S20}), the type of state that the agent can reach (Tsuchiya et al. \citeyear{DBLP:conf/aaaiss/TsuchiyaME23}), or the features impacted by the actions (Z. Lin et al. \citeyear{DBLP:conf/iclr/LinLF21}).

As an example, Tsuchiya et al. present a method similar to reward decomposition (which will be described in the next section), but focused on states. Additional neural networks are learned to estimate the Q-value of each state type.
The different state types are provided in advance by the user. 
A contrastive explanation between the agent's choice of action and that of the user is performed by displaying and then comparing the different Q-values decomposed by state type.

We refer the reader to Appendix~\ref{par:App_C_contrastive} for more details.

                \paragraph{Others}
Three works do not fall into the previously identified work groups (cf. Table~\ref{tab_ch1:expected-outcomes_works}).
In Fukuchi et al. (\citeyear{DBLP:conf/iconip/FukuchiOYI17,DBLP:conf/hai/FukuchiOYI17}), the explanation is the impact of the action on the value of the state features $n$ time steps after performing it, which is associated with an appropriate expert instruction previously used for training the agent.
Image semantic segmentation consists of labelling each pixel of an image. The agent learns to predict an image semantic segmentation of the future state, which acts as an explanation, in Pan et al. (\citeyear{DBLP:conf/icra/PanCCCY19}).

For further details, refer to Appendix~\ref{par:App_C_others_state}.

    %------------------------------------------------
    \subsubsection{Reward}
    \label{sec:reward_decomp}
    %------------------------------------------------

The agent's choice of action can be explained in terms of the reward that the agent is seeking to obtain. 8 out of 10 works (cf. Table~\ref{tab_ch1:expected-outcomes_works}) uses the principle of reward decomposition (Amitai, Septon, et al. \citeyear{DBLP:conf/aaai/AmitaiSA24}; Anderson et al. \citeyear{DBLP:journals/tiis/AndersonDSJNICO20}; Erwig et al. \citeyear{erwig2018explaining}; Feit et al. \citeyear{DBLP:conf/acsos/FeitMP22}; Iucci et al. \citeyear{DBLP:conf/icar/IucciHTIL21}; Juozapaitis et al. \citeyear{juozapaitis2019explainable}; Rietz et al. \citeyear{DBLP:journals/nca/RietzMHSWS23}; Saldiran et al. \citeyear{saldiran2024towards}).

\begin{figure}[]
        \centering
        \includegraphics[width=0.4\textwidth]{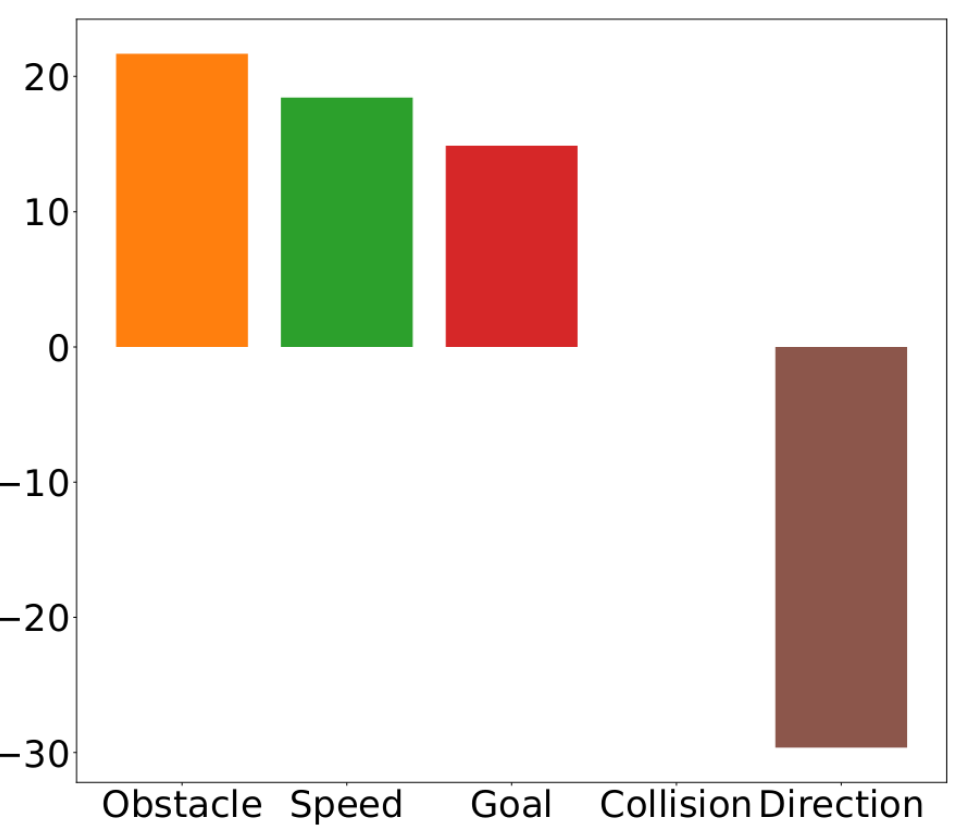}
        \caption[RDX for a Human-Robot collaboration task.]{RDX for a Human-Robot collaboration task (Iucci et al. \citeyear{DBLP:conf/icar/IucciHTIL21}). Two actions (related to robot speed) are compared across four different reward types (on the x-axis of the chart). The y-axis shows the loss and gain for each type of reward when performing the agent's action rather than the other one.}
        \label{fig_ch1:RDX}
    \end{figure}

Reward decomposition is used to represent the reward function in an interpretable way: it is decomposed into a set of comprehensible reward types. Thus, to understand the agent's choice of action, it is sufficient to look at the type of reward it seeks to maximise.
Based on this method, (Erwig et al.; Juozapaitis et al.) propose Reward Difference eXplanations (RDX) and Minimal Sufficient eXplanation (MSX). RDX simply compares 2 actions from a state by comparing their Q-values for each type of reward, while MSX provides a more compact explanation based on RDX. An example of RDX, from Iucci et al., is shown in Figure \ref{fig_ch1:RDX}.

Several works (Amitai, Septon, et al.; Anderson et al.; Feit et al.; Iucci et al.; Rietz et al.; Saldiran et al.) combine reward decomposition (by simply displaying the different Q-values or by using RDX, MSX) with other methods, e.g. autonomous policy explanation (Iucci et al.).

Towers et al. (\citeyear{DBLP:conf/ecai/Towers0FN24}) defines a new estimator allowing the agent to predict the $n$ future expected rewards, which is then used to provide explanations, and Kampik et al. (\citeyear{DBLP:conf/atal/KampikNL19}) proposes several ways to explain actions in a human-agent cooperation context, which could be transposed to a cooperative MARL context.

We refer the reader to Appendix~\ref{sec:App_C_reward} for further details on this line of work.

    %------------------------------------------------
    \subsubsection{Sequence}
    \label{sec:SOTA_expected_outcomes_sequence}
    %------------------------------------------------

The expected outcome of an action here takes the form of a sequence, or a summary of sequences, which the agent can reach by performing an action $a$ from a state $s$.

Three of the five works (cf. Table~\ref{tab_ch1:expected-outcomes_works}) propose summarising future sequences (Luss et al. \citeyear{DBLP:conf/aaai/LussDL23}; Rouveirol et al. \citeyear{DBLP:conf/rulemlrr/RouveirolASV23}; Saulières et al. \citeyear{DBLP:conf/icaart/SaulieresCS23}): Rouveirol et al. groups them using a rule-based classifier that allows each group to be represented by a logical formula, Saulières et al. summarises them into three sequences, which are the best-case, worst-case and most-probable case scenarios according to the reward and transition function of the environment, and Luss et al. groups future states into clusters, or sub-goals that the agent seeks to achieve.

\begin{figure*}[]
        \centering
        \includegraphics[width=1.0\textwidth]{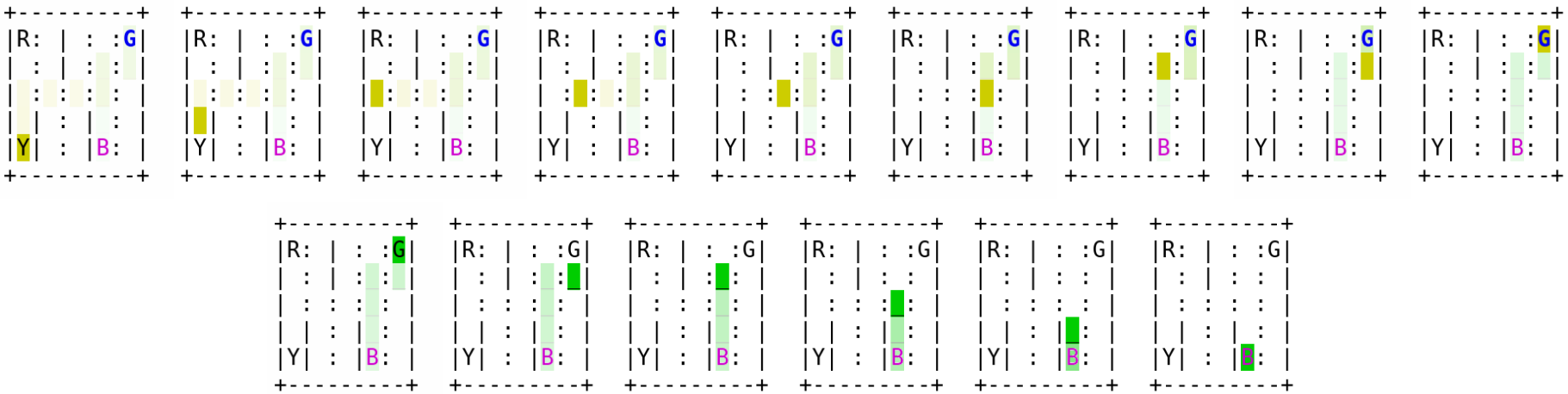}
        \caption[Explanation of each action in an episode in the Taxi environment.]{Explanation of each action in an episode in the Taxi environment (Yau et al. \citeyear{DBLP:conf/nips/Yau0H20}). The episode is described on two lines, starting on the left of the first line. In each state, the agent's future positions according to the belief map are displayed by colored cells. The color intensity reflects the agent's confidence in accessing the position (the more intense the colour, the more confident the agent). In this episode, the Taxi picks up the passenger at the position marked with a `G' and drops him off at the position marked with a `B'.}
        \label{fig_ch1:belief_map}
    \end{figure*}

In Yau et al. (\citeyear{DBLP:conf/nips/Yau0H20}), a belief map is learned, to show the user the states that the agent aims to reach by performing action $a$ from state $s$, as shown in Figure \ref{fig_ch1:belief_map} in the Taxi environment. In Amitai, Septon, et al.  (\citeyear{DBLP:conf/aaai/AmitaiSA24}), the proposed approach shows the user a counterfactual sequence where the agent would have performed $a'$ from $s$, to compare it with the actual sequence.

For more details, refer to Appendix~\ref{sec:App_C_sequence}.

    %------------------------------------------------
    \subsubsection{Action}
    %------------------------------------------------

Only one work explains the choice of an action by indicating future actions that could be taken.% The work \cite{DBLP:conf/aldt/DodsonMG11} is closely related to the problem studied and so future work could draw on it to explain RL agents (provided that the MDP is known or at least approximated). 

                % natural language: action factored differential values + statistics explanation based on database Dodson (MDP context)
Dodson et al. (\citeyear{DBLP:conf/aldt/DodsonMG11}) provide a natural language explanation for advising students on their choice of courses. This problem, represented by a MDP, is explained using two modules, one of which is called Model-Based Explanation.
This module extracts information from the MDP about the usefulness of performing the recommended action in terms of future actions. 
As an example, if the student takes course $x$, this will enable him to take courses $y,z$ in the future.
    
\vspace{0.5em}
% Table description
Table~\ref{tab_ch1:expected-outcomes_works} groups together \emph{Expected Outcomes} works. With 31 works, this ways of explaining action is the least common. Among these works, the majority focuses on expected outcomes taking the form of state (16 works), followed by reward (10 works), particularly obtained via reward decomposition. 

\vspace{0.5em}

In addition to the various works mentioned for action explanation, we have identified four works that use both types of explanation (i.e. \emph{Feature Importance} and \emph{Expected Outcomes}).
To illustrate, an approach is presented in N. Wang et al. (\citeyear{DBLP:conf/atal/WangPH16}) for directly extracting explanations based on the POMDP. The idea is to retrieve the various types of information from the POMDP model and translate them into natural language sentences using predefined templates. For example, the agent can use its expected reward outcome to justify its choice of action, the transition function to describe the probability of achieving different outcomes, or its observation to clarify it to the user..
Among other works, Neerincx et al. (\citeyear{DBLP:conf/hci/NeerincxWKD18}) proposes a general framework for providing perceptual and cognitive explanations based on the agent's confidence, goals, and beliefs, and (Ehsan, Harrison, et al. \citeyear{DBLP:conf/aies/EhsanHCR18}; Ehsan, Tambwekar, et al. \citeyear{DBLP:conf/iui/EhsanTCHR19}) first build a dataset in which each state-action pair is associated with an explanation provided by a user, and then train a model on this dataset to provide an explanation. Refer to Appendix~\ref{par:App_C_mix_explanation} for further details.

%%%%%%%%%%%%%%%%%%%%%%%%%%%%%%%%%%%%%%%%%%%%%%%%%%%%%%%%%%%%%%%%%%%%%%%%
\section{Related Domains}
\label{sec:domains_SOTA}
%%%%%%%%%%%%%%%%%%%%%%%%%%%%%%%%%%%%%%%%%%%%%%%%%%%%%%%%%%%%%%%%%%%%%%%%

    % Take a look at X Planning (close related field)
\paragraph{Explainable Planning}
Planning proposes a set of methods for providing a plan as a solution to a problem given the initial configuration, the goal configuration and a set of actions (or operators) defined by preconditions and effects. 
Planning and RL being relatively close domains, explainability methods for planning, grouped under the name XAIP, are of particular interest. We believe that this area should be further explored by XRL researchers.  
As a first step in this domain, we will briefly describe some XAIP surveys and methods.

        % Survey X Planning 
        % Fox 
A survey and roadmap for XAIP is described by Fox et al. (\citeyear{DBLP:journals/corr/abs-1709-10256}) in order to obtain more efficient methods. Different questions are posed to guide the search, such as \emph{`Why can't you do that?’} or \emph{`Why do I need to replan at this point?’} in the respective cases where a plan is not found by the algorithm and a failure of the current plan occurs.
            % Chakraborti 
Chakraborti et al. (\citeyear{DBLP:conf/ijcai/ChakrabortiSK20}) categorise the different XAIP works according to the target of the explanation and the properties of the explanation. 
The method may explain the planning algorithm, the problem model or the plan. As examples of properties, XAIP methods can use abstractions and/or contrastive approaches.
            % Hoffman 
A brief overview is given in Hoffmann and Magazzeni (\citeyear{DBLP:conf/rweb/HoffmannM19}), with a focus on contrastive explanations. For these explanations, several types of question are interesting to study, such as \emph{`Why action a instead of action b?’} or \emph{`Why does the current plan satisfy property p rather than q?'}.
            % Cashmore (framework contrastive X + roadmap)
Cashmore et al. (\citeyear{DBLP:journals/corr/abs-1908-05059}) describe a framework for answering a set of contrastive questions and the different challenges for effective XAIP methods.

        % Debug plans (Planning) Steinmetz
Plan debugging is carried out by finding states $s$ called bugs where there is a difference between the evaluation of the value of $s$ based on the current plan and the optimal plan (Steinmetz et al. \citeyear{DBLP:conf/aips/SteinmetzFEFGHH22}). This comparison can be performed according to two distinct criteria: a quantitative criterion that considers the cost of the plan, and a qualitative criterion that considers the resolution of the problem.
        % Explain unsolvability through hierarchical abstractions Sreedharan
The unsolvability of a planning problem is explained by a hierarchical abstraction of sub-goals (Sreedharan, Srivastava, Smith, et al. \citeyear{DBLP:conf/ijcai/SreedharanSSK19}). These sub-goals are necessary to solve the problem but are for the most part unreachable. 
The proposed method consecutively identifies the appropriate level of abstraction for the explanation, the sequence of sub-goals to be achieved and the first unreachable one in the sequence.
        
        % X actions (local/global) (Planning) Lomas
The Explaining Robot Action (Lomas et al. \citeyear{DBLP:conf/hri/LomasCCGHK12}) answers a user question in the form of a sentence. To do this, the method determines the information required, selects the template for the answer and sends requests to the world model and the planner to fill in the template.

        % Iterative planning with plan space explanation (Eifler)
In a problem where not all goals are feasible, Eifler et al. (\citeyear{DBLP:conf/aaai/EiflerC0MS20}) propose two types of explanations based on plan-property entailments. 
In this work, a plan property $p$, expressed as a Boolean function, entails $q$, which means that all plans satisfied by $p$ are satisfied by $q$. 
A local explanation for a \emph{`Why not p?’} question is to provide a set of undesirable plan properties that would be satisfied by satisfying $p$. A global explanation is a graph that returns all plan-property entailments. 
This method is used in Eifler and Hoffmann (\citeyear{DBLP:journals/corr/abs-2011-09705}) in an iterative planning context where the user can, via a user interface, ask questions about the different plan-properties. 
        
        % explain necessity of plan step or ordering of plan steps using a constructed axiomatic system Seegebarth
Seegebarth et al. (\citeyear{DBLP:conf/aips/SeegebarthMSB12}) construct an axiomatic system composed of first-order logic formulae and then use it to explain the plan steps and the ordering constraints between plan steps. 
The explanations take the form of a sequence of applications of the axioms allowing the question about the plan to be answered. 

        % User mental model reconciliation using minimally complete explanations (Sreedharan)
In a problem where the user's mental model differs from that of the agent, the explanation consists of a set of model changes so that the plan is optimal for both the agent and the user (Sreedharan, Chakraborti, et al. \citeyear{DBLP:conf/aips/SreedharanCK18}). 
The aim is to provide the user with the minimum number of model changes. 
This approach provides explanations for an uncertain user model or a several-users model. 

        % Interactive interface (model reconciliation problem) contrastive explanation Karthik
RADAR-X (Valmeekam et al. \citeyear{DBLP:conf/aaai/ValmeekamSSK21}) is a user interface that allows the user to ask contrastive questions about the plan in an interactive way. The user describes a plan, and the explanation provides a reduced set of information, based on Chakraborti, Sreedharan, Y. Zhang, et al. (\citeyear{DBLP:conf/ijcai/ChakrabortiSZK17}), to explain the interest of the plan compared to the one proposed.
        
        % Compare two set of plans / generate contrastive X using logical specifications (LTL) Kim
Given a set of positive and negative plans, Kim et al. (\citeyear{DBLP:conf/ijcai/KimMSAS19}) generate a set of LTL formulas as an explanation with a Bayesian inference-based method that is robust to noise. In this context, noise corresponds to swapping plans from the two sets. 
The explanations describe the positive plans and not the negative ones.

        % (pas garder si focus XRL) Multi-objective - contrastive explanations - alternative objectives (context MDP planning) Sukkerd Sukkerd
In the context of multi-objective probabilistic planning, Sukkerd et al. (\citeyear{DBLP:conf/ro-man/SukkerdSG20,DBLP:conf/icse/SukkerdSG18}) propose to compare an optimal policy with alternative policies based on the different objectives. 
This contrastive approach takes the form of a verbal explanation where the values of the policies for the different objectives are compared. 
An algorithm is presented to generate alternative policies that are Pareto optimal according to a certain objective. 

         % (Robot plan explanation Edmonds)
Two explanations are described in Edmonds et al. (\citeyear{DBLP:journals/scirobotics/EdmondsGLXQRZWL19}) for a robotic task. This task is solved with a symbolic action planner and a haptic prediction model. 
The explanation extracted from the symbolic action planner is the symbolic action sequence performed by the robot and the one extracted from the haptic prediction model is a visualisation of the effects of the previous action.

\paragraph{Model Checking}
    % Take a look at Model checking for performing X? Survey  
This domain brings together a set of approaches that analyse a program according to correctness properties (for a survey of the domain, see Jhala and Majumdar \citeyear{DBLP:journals/csur/JhalaM09}). 
Probabilistic (or statistical) model checking is a domain that could be of greater interest to XRL (for a survey, see Agha and Palmskog \citeyear{DBLP:journals/tomacs/AghaP18}). It covers a range of methods for checking properties expressed in stochastic temporal logic. These approaches are based on sampling, and return confidence scores associated with the respect of the properties studied.
In the following, we briefly describe a few works that are more or less close to XRL and that use methods from these domains to provide explanations.

        % XRL context MA Temporal Queries Boggess (probab model checking)
In a MARL context, Boggess et al. (\citeyear{DBLP:conf/ijcai/BoggessK023}) propose an algorithm that verifies the user request expressed in a probabilistic Computational Tree Logic formula (Aziz et al. \citeyear{DBLP:conf/cav/AzizSB95}) using probabilistic model checking. The user request corresponds to a (partial) plan proposed by the user. 
The verification is performed on an abstraction of the multi-agent policy, which is updated if the request is not verified. If, despite this, the request is still not verified, the algorithm generates an explanation as to why it has not been respected.

        % viz tool for Deep Statistical Model Checking (MDP context) 
TraceVis (Gros, Gro{\ss}, et al. \citeyear{DBLP:conf/isola/GrosGGHKS20}) makes it possible to combine visualization methods with model checking, carried out with Deep Statistical Model Checking (Gros, Hermanns, et al. \citeyear{DBLP:conf/forte/GrosH0KS20}). This approach makes it possible to analyse a policy represented by a NN. 

        % probab MC in human-in-the-loop context Li
Li et al. (\citeyear{DBLP:conf/icse/LiAKG20}) use a probabilistic model checker to provide a synthesised explanation to the user when necessary in a human-on-the-loop approach. In this work, an explanation is a content, effect and cost triplet.

        % Utility of MC and PMC
Model checking and statistical model checking would make it possible to study the agent's policy concerning respecting a set of properties in order to analyse in more detail what the agent has learned. 
In addition to properties linked to the agent's performance, it would be interesting to consider more varied properties based on, for example, the use of a certain action (or strategy) that is deemed dangerous or costly by the user, or access to certain regions of the state space. 
Such diversity would help the user decide whether or not to use an agent's policy. %[refVIPER] is an example of work that verifies properties (e.g. ...) based on a policy in the form of a DT. 

\paragraph{Algorithmic Recourse}
    % Take a look at Algorithmic Recourse (close to counterfactual explanations)
This domain is generally applied to classification or regression models and aims to propose a set of approaches explaining a prediction in a contrastive way. The aim is to answer the question \emph{`Why a rather than b from instance x?’} and to provide recommendations for obtaining $b$. 
These recommendations can be seen as a set of actions to be performed from instance $x$ to reach an instance $x'$ where the output of the model would be $b$. 
        % Survey Karimi 
For an overview of the different methods in this domain, we recommend Karimi, Barthe, et al. ( \citeyear{DBLP:journals/corr/abs-2010-04050}).

        % From counterfactual to intervention Karimi
As an example, the work of Karimi et al. (\citeyear{DBLP:conf/fat/KarimiSV21}) presents a formulation of the problem that consists of generating counterfactual instances by minimizing not the change in features of the instance to be explained, but the cost of the actions that lead to a counterfactual instance. 
This reformulation makes it possible to avoid obtaining counterfactuals whose recommendations are sub-optimal or even infeasible.
        % Kind of algorithmic recourse
More recent work in XAI (Pfeifer et al. \citeyear{pfeifer2025explaining}; Rozajac et al. \citeyear{rozajac2025explaining}) proposes approaches that are close to algorithmic recourse. Rozajac et al. propose a generative framework for producing counterfactual explanations in 3D semantic segmentation tasks. Pfeifer et al. introduce a model-agnostic method that generates global explanations of feature importance through counterfactual paths in knowledge-graph-based domains.

        % Utility of AC
Algorithmic recourse seems to be an interesting area to investigate in order to provide XRL methods. Indeed, this approach could be used to explain the choice of one action over another (or the choice of one sequence of actions over another), and provide recommendations for reaching a state (resp. set of states), allowing the contrastive action (resp. sequence of actions) to be performed. This explanation could simply target the understanding of actions or outcomes (e.g. state, reward, respected predicate).
Several works come close to this domain by providing \emph{Counterfactual States} (Druce et al. \citeyear{DBLP:journals/corr/abs-2106-03775}; Gajcin and Dusparic \citeyear{DBLP:conf/atal/GajcinD24}; Huber, Demmler, et al. \citeyear{DBLP:conf/atal/HuberDMOA23}; Olson et al. \citeyear{DBLP:journals/ai/OlsonKNLW21}; Samadi et al. \citeyear{DBLP:journals/corr/abs-2404-18326}) or \emph{Counterfactual Sequences} (Alabdulkarim and Riedl \citeyear{DBLP:journals/corr/abs-2210-04723}; Gajcin and Dusparic \citeyear{DBLP:journals/corr/abs-2402-06503}; Stefik et al. \citeyear{stefik2021explaining}; Tsirtsis, De, et al. \citeyear{DBLP:conf/nips/TsirtsisDR21}; Tsirtsis and Rodriguez \citeyear{DBLP:conf/nips/TsirtsisR23}; Waa et al. \citeyear{DBLP:journals/corr/abs-1807-08706}) but do not provide recommendations. Properties (e.g. recourse, sparsity) and ideas for counterfactual explanations specific to RL are proposed in Gajcin and Dusparic (\citeyear{DBLP:journals/csur/GajcinD24}). One work has been identified as being directly inspired by the domain: Gajcin and Dusparic (\citeyear{DBLP:conf/atal/GajcinD24}) describes a method for providing a counterfactual state reachable by agent actions from the base state.

\paragraph{RL Sub-domains}
    % Reminder HRL / RRL / R-MDP
HRL (cf. Section \ref{par:HRL}), RRL and relational MDP (cf. Section \ref{par:RRL_RMDP}) have been the subject of brief method overviews in this state of the art. These areas bring interpretability by default, hence are interesting to look at in more detail.

    % Take a look at Causal RL : recent Survey Causal RL (Zeng)
Similarly, Causal RL brings together a body of work that combines causality with RL. The aim of this combination is to improve the data efficiency of RL problems while providing interpretability on the causal relations between states, actions, etc. Although some of the work in this survey falls into this category (Druce et al. \citeyear{DBLP:journals/corr/abs-2106-03775}; Herlau and Larsen \citeyear{DBLP:conf/aaai/HerlauL22}; Madumal et al. \citeyear{DBLP:conf/aaai/Madumal0SV20,DBLP:journals/corr/abs-2001-10284}; Volodin \citeyear{volodin2021causeoccam}; Yu et al. \citeyear{DBLP:conf/ijcai/YuRX23}), we recommend the survey Zeng et al. (\citeyear{DBLP:journals/corr/abs-2302-05209}) which presents a general overview of the advances in this domain by categorising the work into two categories: approaches which must learn the causal information of an environment before the agent is trained, and those which already have this information.

\paragraph{RL as an Explanation Framework}

The following section mentions different works found during our research that provide explanations for AI models using RL. 
To clarify, the idea behind this framework is to model (in part) the problem of finding explanations as an RL problem. Different AI models can be explained using this framework, such as classification or regression models (Z. Chen et al. \citeyear{DBLP:conf/cikm/ChenSW0AT22}; Kohler, Akrour, et al. \citeyear{DBLP:journals/corr/abs-2304-05839}; Lash \citeyear{DBLP:journals/corr/abs-2206-01343}; Nguyen et al. \citeyear{DBLP:journals/corr/abs-2103-12983}; Samoilescu \citeyear{DBLP:journals/corr/abs-2106-02597}), recommendation systems (X. Wang, Y. Chen, et al. \citeyear{DBLP:conf/icdm/WangCYWW018}) or policies learned via RL (Saulières et al. \citeyear{DBLP:conf/icaart/SaulieresCS23}).

As an example of work, to answer in a counterfactual fashion the question \emph{`Why predict a rather than b from instance x?'}, Lash proposes to describe the problem of generating a counterfactual instance of a classification or regression model as a decision-making problem, then solve it using a DRL algorithm.

For more details on the other works mentioned, we refer the reader to Appendix~\ref{par:App_C_RL_explanation_framework}.

\vspace{0.5em}

    % Transition
We believe that the domains discussed are of real interest in the future design of XRL methods. We encourage researchers to read the various surveys highlighted: Agha and Palmskog (\citeyear{DBLP:journals/tomacs/AghaP18}); Cashmore et al. (\citeyear{DBLP:journals/corr/abs-1908-05059}); Chakraborti, Sreedharan, and Kambhampati (\citeyear{DBLP:conf/ijcai/ChakrabortiSK20}); Fox et al. (\citeyear{DBLP:journals/corr/abs-1709-10256}); Hoffmann and Magazzeni (\citeyear{DBLP:conf/rweb/HoffmannM19}); Jhala and Majumdar (\citeyear{DBLP:journals/csur/JhalaM09}); Karimi, Barthe, et al. (\citeyear{DBLP:journals/corr/abs-2010-04050}); Otterlo (\citeyear{DBLP:books/sp/12/Otterlo12}); Pateria et al. (\citeyear{DBLP:journals/csur/PateriaSTQ21}); Zeng et al. (\citeyear{DBLP:journals/corr/abs-2302-05209}).
The explainability of agents that have learned by reinforcement is a flourishing domain that requires particular attention on several points, or needs. We will list these needs below, which could also be considered for the XAI domain in general.

%%%%%%%%%%%%%%%%%%%%%%%%%%%%%%%%%%%%%%%%%%%%%%%%%%%%%%%%%%%%%%%%%%%%%%%%
\section{Needs for XRL}
\label{sec:needs_SOTA}
%%%%%%%%%%%%%%%%%%%%%%%%%%%%%%%%%%%%%%%%%%%%%%%%%%%%%%%%%%%%%%%%%%%%%%%%

\paragraph{Compare Methods}
    % Need for comparison of XRL methods + create specific benchmarks
Of the various works cited in this survey, a relatively small proportion compare their approach with other XRL methods.
Indeed, of the $238$ papers presented in this paper that fall directly within the field of XRL, only $55$ works offer a comparison in terms of explainability with other XRL methods.
        % Milani survey
As stated in the surveys (Milani, Topin, et al. \citeyear{DBLP:journals/csur/MilaniTVF24}; Vouros \citeyear{DBLP:journals/csur/Vouros23}), it is necessary to compare XRL methods. 
In this way, for a given target (e.g. action) and for the same type of method (e.g. feature importance), developers could determine which method is best suited to the problem considered. %Comparing two methods that do not have the same type may be more difficult. 
To give a few examples, the works (Barceló et al. \citeyear{DBLP:conf/nips/BarceloM0S20}; Pierson et al. \citeyear{DBLP:journals/nca/PiersonAMT24}; Pocius et al. \citeyear{DBLP:conf/aaai/PociusNF19}; Xiong et al. \citeyear{DBLP:journals/corr/abs-2402-12685}) tend to compare explainable methods in different ways. 

For classifiers, Barceló et al. compare methods by studying of the computational complexity of generating explanations.
The evaluation of XRL methods can either compare several types of explanation (Pierson et al.; Pocius et al.) or focus on a specific group of methods (Atrey et al. \citeyear{DBLP:conf/iclr/AtreyCJ20}; Xiong et al.). 

For instance, Pierson et al. compare multiple XRL methods for explaining an agent’s policy through a user study. Theses methods include HIGHLIGHTS (D. Amir and O. Amir \citeyear{DBLP:conf/atal/AmirA18}), a combination of HIGHLIGHTS with saliency maps, graphs summarizing the agent’s behavior, and a hybrid approach combining saliency maps and graphs. 
Pocius et al. propose three evaluation tasks based on the StarCraft II environment (Vinyals et al. \citeyear{DBLP:journals/corr/abs-1708-04782}) to assess XRL methods.

Conversely, (Atrey et al.; Xiong et al.)  focus on evaluating a specific group of XRL methods. 
Xiong et al. introduce a benchmark dedicated to \emph{Feature Importance} approaches, including a set of environments, implemented XRL methods, and evaluation metrics such as fidelity, stability, and computation time. 
More specifically, Atrey et al. develop a dedicated methodology for evaluating saliency maps based on image interventions, allowing them to analyze the reliability of such explanations.

\paragraph{Provide Metrics}
    % Need unified way to compare XRL methods through specific metrics
One of the shortcomings of XRL is the lack of a unified way of comparing methods, whatever their type or target. For a high level overview of metrics used in XRL, we refer the reader to Milani, Topin, et al. (\citeyear{DBLP:journals/csur/MilaniTVF24}). %are proposed, we have not found any metrics that are not specific to a type of method. 
We believe that a unified assessment of methods is important, but also a more diverse set of metrics for each type of method in order to measure the quality of the explanations. In the following, we present different ways of measuring the quality of explanations and categorizing the evaluation methods in the XAI domain.

    % methods and metrics XAI Zhou
Zhou et al. (\citeyear{zhou2021evaluating}) present a survey of methods and metrics for evaluating XAI methods. A set of metrics are categorised according to the type of explanation evaluated and the properties of explanations taken into account by the metric (e.g. clarity, parsimony, soundness). 

        % Doshi-Velez three main levels for explanation evaluation
Doshi-Velez and Kim (\citeyear{doshi2017towards}) divide the methods for evaluating the interpretability of a model into three parts. 
Application-grounded evaluation groups together methods that rely on end-task experts to determine its interpretability. For human-grounded evaluations, this is determined using lay persons. 
Functionally-grounded evaluations are a set of methods that use functions defined by certain properties that act as interpretability metrics, and therefore do not require human experiments.

        % Evaluate Interpretable ML without ground truth Yang LO
An overview of evaluation methods for interpretable machine learning models is given in F. Yang, Du, et al. (\citeyear{DBLP:journals/corr/abs-1907-06831}). Three properties are considered in the evaluation methods: generalizability, which determines the extent to which the explanations are specific to the instance, fidelity, which determines the extent to which the explainer's output matches the model's decision-making process, and persuasibility, which determines user satisfaction. 

        % Metrics for XAI
            % Hoffman
A set of measures is presented in Hoffman et al. (\citeyear{DBLP:journals/corr/abs-1812-04608}) for XAI. These are grouped into four sets: methods that evaluate the quality of explanations, user satisfaction, user understanding of the model and performance according to their mental model.

            % Rosenfeld 
4 metrics are set out in the blue-sky paper (Rosenfeld \citeyear{DBLP:conf/atal/Rosenfeld21}). $D$ quantifies the difference in performance between the agent's opaque model and its surrogate model. $R$ measures the simplicity of the proposed model, $F$ the number of elements useful for generating the explanation and $S$ the stability of the explainer.

            % Axioms Amgoud
Amgoud and Ben-Naim (\citeyear{DBLP:conf/ijcai/AmgoudB22}) present a total of 10 axioms that explainers should satisfy. These axioms are limited to explainers whose explanation is a subset of the features of an instance. Among the axioms, irreducibility defines that an explainer should only contain useful features and feasibility defines that an explainer is a subset of at least one instance of the classification problem.

            % Ground truth Data Amiri 
Depending on the dataset, Amiri et al. (\citeyear{DBLP:journals/corr/abs-2011-09892}) propose to generate a ground truth of explanations, which allows comparison with the explanations produced by LIME (Ribeiro et al. \citeyear{DBLP:conf/kdd/Ribeiro0G16}).

        % Evaluate XAI methods 
            % Waldchen
In Wäldchen et al. (\citeyear{DBLP:conf/icml/WaldchenPH22}), several methods based on the notion of feature importance are compared using an auxiliary task. This task consists of guiding an agent playing Connect4 to win. In this context, the agent has only a partial observation of the board, which is given by a feature importance method. 
The idea is to measure, over a set of games, whether the agent manages to win based solely on the cells on the board indicated as important by an explainer.

            % Assessment Sokol
Sokol and Flach (\citeyear{DBLP:conf/fat/SokolF20}) present a set of elements to be considered when evaluating XAI methods, which are grouped according to $5$ dimensions: functional, operational, usability, safety and validation.

\paragraph{Perform User Studies}
    % Need more user study
Of the various works cited in this survey, a relatively small proportion validate their approach with a user study (which is also noted in Milani, Topin, et al. \citeyear{DBLP:journals/csur/MilaniTVF24}; Wells and Bednarz \citeyear{DBLP:journals/frai/WellsB21}). 
Indeed, of the $238$ papers presented in this paper that fall directly within the field of XRL, only $55$ works conducted a user study to evaluate the proposed XRL method. 
As consumers of explanations of agents' behavior, it makes sense that any XRL method should at least be evaluated by a user study. 
We believe that taking end-users into consideration in the development process of an explainable method is essential, whether the end-users are domain experts, developers or lay persons. Furthermore, with sufficiently in-depth user studies, it would be possible to identify the strengths and weaknesses of the approach, and then use this feedback to improve the approach.
        % Saliency + reward bars Anderson
        % Special XRL user study Dodge 
        % Compare XRL methods (Pierson)
        % Examples of work with user studies sequeira / gajcin (RACCER)  
%Here is a non-exhaustive list of works described in this state of the art that validate their approach through user studies \cite{DBLP:journals/ai/SequeiraG20,DBLP:conf/atal/GajcinD24,DBLP:conf/atal/AmirA18} or simply compare two already existing XRL methods with each other \cite{DBLP:journals/tiis/AndersonDSJNICO20,dodge2021no,DBLP:journals/nca/PiersonAMT24}.

\paragraph{Develop User Interface}
    % Develop IHMs
In line with the above need, we believe that it is important to propose, in addition to an XRL method, a dedicated user interface. The construction of a toolbox is suggested as a future direction in Vouros (\citeyear{DBLP:journals/csur/Vouros23}). An interface should be intuitive and ergonomic so that the user gets the most out of the explainability method. 
In user studies, a more or less basic interface is proposed. The various works explaining the agent's policy through \emph{Visual Toolkit} (W. He et al. \citeyear{DBLP:conf/apvis/HeLBWS20}; Jaunet et al. \citeyear{DBLP:journals/cgf/JaunetVW20}; McGregor et al. \citeyear{DBLP:journals/vlc/McGregorBDHMM17}; A. Mishra et al. \citeyear{DBLP:journals/corr/abs-2104-02818}; J. Wang et al. \citeyear{DBLP:journals/tvcg/WangGSY19}) is a good example of a way of thinking about explainability. 
In our opinion, it is as important to provide a good XRL method as it is to provide a user interface that improves its usefulness from a user's point of view. 

%%%%%%%%%%%%%%%%%%%%%%%%%%%%%%%%%%%%%%%%%%%%%%%%%%%%%%%%%%%%%%%%%%%%%%%%
\section{Conclusion}
\label{sec:conclusion_SOTA}
%%%%%%%%%%%%%%%%%%%%%%%%%%%%%%%%%%%%%%%%%%%%%%%%%%%%%%%%%%%%%%%%%%%%%%%%

This paper has categorised recent work in the field of XRL using two questions: \emph{`What?’} and \emph{`How?’}. 
3 targets for explanations have been identified, namely the agent's policy, the state-action sequence resulting from the agent's interaction within the environment and the agent's choice of action. 
%Several ways of providing these explanations have been proposed, which can be summarised in three clear ideas. 
%The first idea is to make behavior interpretable by directly influencing its knowledge representation. The second idea is to represent the environment in which the agent evolves in a comprehensible way. The last idea is to integrate methods from outside the RL paradigm to explain the agent's decisions.

    % Majority / Minority type of explanation 
This state of the art allows us to highlight the low interest of researchers in explaining sequences of agent interactions: $176$ works use methods which are categorised as \emph{Policy-level methods}, $90$ as \emph{action-level methods} and only $13$ as \emph{sequence-level methods}. We assume that this is due to the fact that XAI for classifiers influences XRL. Indeed, the vast majority of classifier explanation methods simply explain a decision locally or the model globally. 
%We encourage researchers to explain the agent's action choice sequences, to provide better heterogeneity of XRL methods.
On the other hand, many methods have been proposed to explain a simple agent decision or policy. Among the different approaches identified, the majority of methods that explain an agent's decision by \emph{Feature Importance} do so using saliency maps, although this approach is limited to agents whose state is an image. For methods based on expected outcomes, the majority return states or features of states as explanations. 
For policy, the majority of papers fall into the \emph{Interpretable Policy} category, with a predominance of methods proposing directly interpretable policies.

    % Transition
%A small proportion of the works presented do not aim at explainability, but rather at performance or policy generalisation. However, we thought it would be interesting to include these works, as the methods described make it possible to obtain an interpretable agent or to take an intermediate step between opacity and clarity of the agent's behavior. 

    % Conclusion of Conclusion
%As described at the beginning of the conclusion, 
This state of the art proposes an intuitive taxonomy that allows us to quickly identify a set of works related to the target we want to explain and the way we want to do it. 
%The different targets are the policy, a sequence of actions and an agent action. %Few works are devoted to the explanation of sequences in comparison with the other two targets. 
This paper has also enabled us to highlight the types of work done to explain or make interpretable one of the targets, to show the use of methods initially intended for the explanation of classifiers in the context of XRL. Moreover, it briefly describes a set of domains that we consider relevant to explore in order to propose new XRL methods and lists several needs for this topic.

%%%%%%%%%%%%%%%%%%%%%%%%%%%%%%%%%%%%%%%%%%%%%%%%%%%%%%%%%%%%%%%%%%%%%%%%
\begin{acks}
%%%%%%%%%%%%%%%%%%%%%%%%%%%%%%%%%%%%%%%%%%%%%%%%%%%%%%%%%%%%%%%%%%%%%%%%

We thank the reviewer for their valuable feedback and suggestions that helped improve this survey.

\end{acks}

%%
%% The next line prints the references.
\printbibliography

\appendix
\section{Summaries of Policy-level Methods}
\label{sec:App_A}

%    \begin{figure*}[h]
%        \centering
%        \includegraphics[width=1.0\textwidth]{images/Taxonomy_policy.png}
%        \caption[XRL taxonomy: policy-level methods]{Detailed taxonomy for policy-level methods.}
%        \label{fig_ch1:Taxonomy_policy}
%    \end{figure*}
This appendix provides a summary of each policy-level method mentioned in the paper.
These methods explain the agent’s policy.
As a reminder, a total of four ways of explaining the policy have been identified. 
Constructing \emph{Interpretable Policies} makes the agent’s reasoning transparent. 
\emph{Summarising Policies} provides an overview of the agent’s capabilities. 
The use of \emph{Human-readable MDP} allows the user to understand the information available to the agent, its actions and the dynamics of the environment. 
\emph{Visual analysis} helps the user to study the agent’s policy visually, using various tools and metrics. 

%------------------------------------------------
%------------------------------------------------
\subsection{Interpretable Policy}
\label{sec:App_A_interpretable_policy}
%------------------------------------------------
%------------------------------------------------

    %------------------------------------------------
    \subsubsection{Inherently Understandable}
    \label{sec:App_A_policy_inherently_understandable}
    %------------------------------------------------

                \paragraph{Hierarchical Policy.}
                \label{par:App_A_IU_Hierarchical}
                % Hierarchical RL
As a reminder, Hierarchical RL (HRL) is a sub-domain of RL consisting in learning a policy which is structured in a hierarchical manner, generally comprising 2 levels, so as to break down the task to be performed into a set of sub-tasks. 
For each sub-task, a \emph{low-level} policy is learned. A \emph{high-level} policy is learned using the various \emph{low-level} policies as an action.

%\todoinflo{les paragraphes suivants forment une liste à la prévert on ne voit pas de structure en quoi ils se ressemblent en quoi ils différent, peut-on en regrouper certains?}
                    % XAI papers
                        
                        % Shu (capture temporal transitions between tasks) (multi-task RL)
The low-level policies are represented using a human language description in the following works. For Minecraft games, agents learn to determine when it is necessary to learn a new policy (Shu et al. \citeyear{DBLP:conf/iclr/ShuXS18}). If not, the agent chooses one of the policies it has already learned. These policies are interpretable  due to their description%they are represented by a human language description
, such as \emph{`Get object o'}. 
To learn the dependencies between different tasks/policies (for example, it is necessary to use \emph{`Get object o'} before \emph{`Put object o'}), a stochastic temporal grammar model is learned.
                        % low level policies correspond to language instructions Jiang (human interpretable  actions)
In the same vein, Jiang et al. (\citeyear{DBLP:conf/nips/JiangGMF19}) present Hierarchical Abstraction with Language using high-level actions, sub-goals, represented by a human language description.
The high-level policy and the low-level policy are trained separately, where the low-level policy is learned based on a reward function that depends on an (assumed known) function determining whether a sub-goal is respected or not. 
The results show that this framework produces interpretable policies that perform well on long-horizon problems with sparse reward. 
                        
                        % Xu et Fekri (learn symbolic state transitions using ILP)
Model-based RL methods use the dynamics of the environment (i.e. the transition function and the reward function) to obtain an agent's policy. Note that the dynamics are not necessarily known beforehand.
In a model-based HRL context, Xu and Fekri (\citeyear{DBLP:journals/corr/abs-2106-11417}) propose to learn a transition function for high-level symbolic states (i.e. sub-goals) by modifying DILP (Evans and Grefenstette \citeyear{DBLP:journals/jair/EvansG18}) so as not to generate meaningless clauses.
The transitions generated are logic rules expressing the preconditions and effects of the various sub-goals. 
                        % Wu (imperfect world models used to decompose a complex task into simpler ones / simultaneously learn subpolicies and controller)
In the same context, the Model Primitives HRL framework (B. Wu et al. \citeyear{DBLP:conf/atal/WuGK19}) allows a set of modular interpretable policies to be learned and combined based on a set of imperfect models of the environment.
The idea is that approximate models of the environment are specialised in different regions of the environment. In addition to the policies, a gating controller is learned to compose the policies and solve the basic task.

                        % Symbolic planning High level + RL low levels (similar to hierarchical?) Lyu
Three works propose to mix Planning and HRL. The Symbolic DRL framework (Lyu et al.  \citeyear{DBLP:conf/aaai/LyuYLG19}) combines a high-level planner that returns a plan composed of symbolic actions with sub-policies learned by DRL for each symbolic action. The framework is made up of three elements: a planner that builds a plan of symbolic actions, a controller that learns a sub-policy and a meta-controller that obtains the quality of the proposed plan and thus helps the planner to propose a better plan.
An intrinsic and environment reward are respectively used to learn the sub-goals, i.e. symbolic actions, and to measure the plan quality.              
                        % Wang  sub-task are learnt via RL
%\leo{Planning}In the same idea, a
The combination of Planning and HRL is also proposed in Chang et al. (\citeyear{chang2020coactive}) for a human-UAVs cooperation problem. The plan represented by a tree and the sequential order of the actions to be performed is displayed to the user in such a way as to make the plan interpretable.
                        % Mix Planing RL Jin
%\leo{Planning}With the aim of combining Planning and HRL, 
Jin et al. (\citeyear{DBLP:conf/aaai/JinMJZCY22}) use the option framework (Sutton, Precup, et al. \citeyear{DBLP:journals/ai/SuttonPS99}). %\leo{with the aim of combining Planning and HRL}.
An option consists of pre-conditions, a policy and effects. Option policies are learned in RL and a planner is used to generate a plan. 
To use this approach, the user requires access to a function defined upstream which extracts a symbolic state from an image state.

                        % Beyret
For a robotic manipulation task, the Dot-to-Dot method (Beyret et al. \citeyear{DBLP:conf/iros/BeyretSF19}) uses HRL to decompose the problem into simpler tasks, assimilated to sub-goals, which the low-level agent learns to perform. These sub-goals are iteratively generated by the high-level agent to guide the low-level agent.
The low-level agent's input is its observation and the sub-goal to be achieved. This approach makes it possible to interpret the impact of the sub-goals on the agent's learned Q-values, and to observe that the high-level agent has learned a notion of distance for the given problem.
    
                        % Decompose task into interpretable sub-tasks (not labelled as HRL)
                        % van Rossum  
In Rossum et al. (\citeyear{DBLP:journals/corr/abs-2104-06630}), a reward function is based on the achievement of a sub-goal for learning. In addition, a curiosity reward is generated using a Generative Adversarial Network (GAN) (Goodfellow et al. \citeyear{DBLP:journals/corr/GoodfellowPMXWOCB14}) that generates the state resulting from an action $a$ from a state $s$. The sub-goal is generated by a neural network that selects a feature of the agent's current state as the sub-goal to be achieved. This method is specific to a problem where the state represents a grid in which the agent can move. The sub-goal is transcribed manually using a natural language description. In this work, a single policy is learned and is not technically an HRL method, although the task is decomposed into sub-objectives.

                        % Rietz (mix with other XRL)
Rietz et al. (\citeyear{DBLP:journals/nca/RietzMHSWS23}) combine HRL with reward decomposition to provide explanations of the agent's decisions. A context is associated with the reward decomposition, providing the user with the sub-goal that the agent is currently aiming to achieve.
This method explains the policy with the different sub-goals but also the agent's local decision by providing a reward decomposition.

                        % Andreas (learn from handcrafted policy sketches) (multi-task RL)
Using policy sketches, the agent learns to solve multi-task RL problems (Andreas et al. \citeyear{DBLP:conf/icml/AndreasKL17}). These policy sketches are sequences of symbolic labels. Each label is a comprehensible task that is learned by a sub-policy in the form of a neural network.
%A \emph{stop} action is artificially added to allow the next sub-policy (in the policy sketch) to learn.
    
                        % Tasse (sub-goals are boolean formulas)
\emph{Composition} (Todorov \citeyear{DBLP:conf/nips/Todorov09}), a slightly different approach to HRL, has also been used in Tasse et al. (\citeyear{DBLP:conf/nips/TasseJR20}). It consists of learning a new skill based on old skills that have already been learned.
The interpretable aspect presented in this paper is that the new skill is composed without new learning, simply by using a boolean algebra on the tasks (modelled by MDP's) and on the value functions already learned.

                    % Not XAI papers:
In this paragraph, we give a brief presentation of some HRL methods or methods close to the domain which are not focused on explainability, but which are nonetheless worth mentioning. 
                    % Leonetti (ASP)
                    % Sridharan (use of ASP program to obtain a plan of abstract actions)
Two works (Leonetti et al. \citeyear{DBLP:journals/ai/LeonettiIS16}; Sridharan et al. \citeyear{DBLP:journals/jair/SridharanGZW19}) use Answer Set Programming (Lifschitz \citeyear{DBLP:conf/aaai/Lifschitz08}) to constrain the agent's policy (Leonetti et al.) or define a high-level plan made up of abstract actions (Sridharan et al.).
                        % Furelos-Blanco (Finite-state Automaton high level using ILP + RL subgoals expressed prop logic formulas)
Furelos-Blanco et al. (\citeyear{DBLP:journals/jair/Furelos-BlancoL21}) mix Inductive Logic Programming (ILP) and RL by learning an automaton composed of edges which represent a sub-goal by a logic formula.
                        % Symbolic planning + HRL Yang (not XAI)
The framework proposed in F. Yang, Lyu, et al. (\citeyear{DBLP:conf/ijcai/YangLLG18}) combines planning and RL to provide robust symbolic plans.
                        % Attributes space planning + HRL Zhang (not XAI)
In A. Zhang et al. (\citeyear{DBLP:conf/icml/ZhangSLSF18}), the agent learns a graph, the environment model based on a high-level state representation and uses it to learn a policy.
                        % State-graph search for planning + HRL Eysenbach (not XAI)
In the same vein, Eysenbach et al. (\citeyear{DBLP:conf/nips/EysenbachSL19}) build a graph from the sub-goals obtained using the replay buffer and then use Djikstra's algorithm (Djikstra \citeyear{DBLP:journals/nm/Dijkstra59}) to plan.
                        % Hierachical + RRL (FOL Q-DT) Driessens paper: 'Learning Digger using Hierarchical Reinforcement Learning for Concurrent Goals'
HRL is combined with the RRL in Driessens and Blockeel (\citeyear{driessens2001learning}) to learn an efficient policy where each sub-policy is learned by a Q-tree.
                        % Hierarchical DQN Kulkarni
The hierarchical DQN (Kulkarni et al. \citeyear{DBLP:conf/nips/KulkarniNST16}) represents a set of Q-functions in a hierarchical manner, so that the low-level ones learn a sub-policy for a given objective and the high-level ones solve the base problem.

                \paragraph{Rules}
                \label{par:App_A_IU_Rules}

                    % Fuzzy rules (if ... then ...)
                        % Fuzzy RL policy (NN world models) Hein
A model-based approach is presented in Hein, Hentschel, et al. (\citeyear{DBLP:journals/eaai/HeinHRU17}), entitled Fuzzy Particle Swarm RL.
The aim is to learn a set of fuzzy rules offline with particle swarm optimization (Kennedy and Eberhart \citeyear{DBLP:conf/icnn/KennedyE95}), using NN's to represent world models. The produced rules take the form: \emph{`if s is m then o'} where $s$ is a state, $m$ a Gaussian membership function and $o$ a real number.
The number of rules to be produced must be provided prior to learning. Note that only 2 rules are needed to learn a good policy for the Mountain Car and CartPole problem. 
The rules are visualised by presenting the membership functions and an example state. The learning of world models has an impact on the production of efficient policies: models that do not approximate the real dynamics correctly can cause the agent to learn the wrong policy. 

                        % Fuzzy sets Huang
In the same vein, Huang et al. (\citeyear{DBLP:journals/eaai/HuangAY20}) propose the interpretable fuzzy RL method to produce an interpretable policy in the form of fuzzy rules. This method is model-free and uses an actor-critic structure to learn the rules. To use this approach, the state-action space is discretised.
                       
                        % Akrour
As a preliminary work (Akrour et al. \citeyear{akrour2019towards}), the interpretable policy takes the form of a succession of blocks \emph{`if close(s, center[k]) do action[k]'} where the state space is split into clusters with associated actions.
The \emph{close} function, which is assumed to be known, determines whether a state $s$ belongs to a cluster. The number of clusters is limited upstream and the cluster centres are not optimised. Learning such a policy is based on the approximate iteration framework (Bertsekas \citeyear{bertsekas2011approximate}).
                        
                        % Ou + attention maps
With a state represented by an image, the architecture described in Ou et al. (\citeyear{DBLP:journals/tfs/OuCWL24}) breaks down into three parts: a Compact Convolutional Transformer (CCT) (Hassani et al. \citeyear{DBLP:journals/corr/abs-2104-05704}) extractor which uses an attention mechanism to extract relevant features, a fuzzy-based decision network which learns fuzzy if-then rules from the output of the CCT, and a Convolutional NN (CNN) latent decoder which reconstructs the agent's state.
The policy is analysed by visualising the attention map extracted from the CCT and the most influential rules in the agent's choice of action.

                    % Logic rules
In the following works, the rules are expressed in first order logic, each taking the form of a clause $a \leftarrow a_1, a_2, \dots a_n$ where a is the head atom (i.e. the action to be performed) and $a_1, a_2, \dots a_n$ the body atoms (i.e. the preconditions to be met). Most of the works (Hazra and Raedt \citeyear{DBLP:conf/pkdd/HazraR23}; Z. Jiang and Luo \citeyear{DBLP:conf/icml/JiangL19}; Payani and Fekri \citeyear{DBLP:journals/corr/abs-2003-10386}; L. Zhang et al. \citeyear{DBLP:conf/pkdd/ZhangLWT21}) is based on DILP (Evans and Grefenstette \citeyear{DBLP:journals/jair/EvansG18}) to produce interpretable policies. This neuro-symbolic approach makes it possible to learn a set of symbolic rules weighted by a confidence score, while still having the advantages of using NN's.
                    
                        % importance: weighted rules Jiang
Z. Jiang and Luo present Neural Logic RL (NLRL) for learning a policy in the form of first-order logic rules. For this purpose, a DILP architecture is proposed as well as an MDP with logic interpretation.
For the experiments, the authors provided only minimal atoms as input to describe the agent's states and background. The NLRL policies are interpretable and achieve near-optimal performance on the problems tested.

                        % Differentiable Logic RL: wieghted rules Zhang
Also based on DILP, L. Zhang et al. propose an off-policy algorithm for learning a set of rules.
This method is designed to use approximate inference to reduce the number of rules and has the same or better performance than (Jiang and Luo).
                        
                        % DERRL Hazra
Deep Explainable Relational RL (Hazra and Raedt) represents an interpretable policy by a set of if-then rules.
The objective is to find a policy such that given a state and background knowledge, an action is predicted. An NN is learned to produce rules and semantic constraints are proposed to avoid redundancy between rules. 
The approach outperforms NLRL (Jiang and Luo) in terms of generalisation and computation time.

                        % logic state representation + logic rules learned Payani
In Payani and Fekri, the approach is a joint use of RRL (Dzeroski et al. \citeyear{DBLP:journals/ml/DzeroskiRD01}) and DILP. RRL is used to transpose the state of an agent in image form into a set of auxiliary predicates.
To do this, a CNN is used to take a low-level representation as input and convert it, using auxiliary predicates, into a high-level representation. The dNL-ILP engine (Payani and Fekri  \citeyear{DBLP:journals/corr/abs-1906-03523}) is then used to express the policy in the form of rules.

                        % Neuro-symbolic Ma
In comparison with DILP-based methods, Neural Symbolic RL (Ma et al. \citeyear{DBLP:journals/corr/abs-2103-08228}) avoids storing all the rules by using an attention module. This framework is divided into three parts: the attention module, the reasoning module and the policy module.
This approach provides near-optimal policies composed of chain-like rules. However, the rules generated are not always true and must be selected by an expert. 
                        
                        % Dong (logic program instead of weighted combination of logic formulas)
Neural Logic Machines (NLM) (Dong et al. \citeyear{DBLP:journals/corr/abs-1904-11694}) is a neuro-symbolic architecture for learning the underlying logical rules of a problem.
To do this, starting from a set of basic predicates applied to objects, a succession of first-order rules are applied using Multi-Layer Perceptron's (MLP), to obtain a set of conclusions about the objects. 
NLM is able to solve simple reinforcement learning tasks. The problem with this approach is that the rules learned are not interpretable.

                        % FOL program Zimmer / Song
Based on the NLM architecture (Dong et al.), Zimmer et al. (\citeyear{DBLP:journals/corr/abs-2102-11529}) introduce the Differentiable Logic Machine (DLM), a high-performance neuro-logic architecture which outperforms dNL-ILP (Payani and Fekri \citeyear{DBLP:journals/corr/abs-1906-03523}) and NLM in terms of computation time and memory used.
Two operations on predicates are added compared to NLM, namely negation and preservation, and MLP's are replaced by logic modules, which correspond to fuzzy AND and OR operations on predicates, to improve the interpretability of the model. A post-process approach is used to extract logical formulas from the model, making its reasoning interpretable.
In Song et al. (\citeyear{song2022interpretable}), this competitive method is tested on a set of autonomous driving scenarios represented in ILP problems.
DLM achieves good performance (similar to or better than NLM) while extracting an interpretable set of rules. 
                        
                \paragraph{Decision Tree}
                \label{par:App_A_IU_DT}

                 % DTs:
                    % Conservative Q Improvement Roth
Conservative Q improvement (Roth et al. \citeyear{DBLP:journals/corr/abs-1907-01180}) is an algorithm that manages the trade-off between interpretability and performance during learning.
The policy takes the form of a DT where the internal nodes are conditions based on features of a state and the leaves contain the Q-values of possible actions. Using a dynamic threshold, the tree is extended only if this extension results in a sufficient expected discounted future reward.
                    
                    % differentiable DT (policy gradient) discretized + tranformed into decision list tree Silva
Differentiable DT (DDT) is used in Silva et al. (\citeyear{silva2019optimization}). The DDT can be used for Q-learning, where each leaf contains the Q-values of the actions, or for a gradient-based approach, where each leaf contains the distribution of the actions.
The tree is then discretised to obtain an interpretable DT (an example for the CartPole environment (Brockman et al. \citeyear{brockman2016openai}) is shown in Figure \ref{fig_ch1:DDT}).
In addition, a decision list can be extracted. 
The results show that these trees have a good performance compared to a MLP, while being more interpretable, according to the user study carried out. 

%    \begin{figure}[]
%        \centering
%        \includegraphics[width=0.5\linewidth]{images/DDT.png}
%        \caption[Example of discrete decision tree]{A discrete decision tree for the CartPole environment \cite{silva2019optimization}. Each internal node of the tree (in blue) represents a condition on a feature of the agent's state, and each leaf of the tree (in green) corresponds to an action.}
%        \label{fig_ch1:DDT}
%    \end{figure}
                    
                    % Iterative bounding mdps Topin
Topin et al. (\citeyear{DBLP:conf/aaai/TopinMFV21}) propose a specific representation of an MDP, called Iterative Bounding MDP (IBMDP), to provide a policy in the form of DT's for the original MDP. This policy is learned using modified versions of either policy-gradient or Q-learning algorithms. The aim is to create a policy that also works for the base MDP. The CUSTARD algorithm presented for solving IBMDPs obtains better results than VIPER (Bastani, Pu, et al. \citeyear{DBLP:conf/nips/BastaniPS18}) concerning tree size and policy performance.
                    
                    % Evolutionnary approach Custode / Custode
An evolutionary approach coupled with Q-learning is presented in Custode and Iacca (\citeyear{DBLP:journals/access/CustodeI23}). The evolutionary approach, inspired by the Grammatical Evolution algorithm (Ryan et al. \citeyear{DBLP:conf/eurogp/RyanCO98}), is used to find the tree structure and Q-learning to determine the action to be taken for each leaf of the tree.
Once the policy has been obtained, a test phase is carried out on 100 episodes to compress the tree. Nodes that are not visited are deleted, as well as internal nodes whose leaves lead to the same action. 

                    % Interp. Continuous Control trees Paleja (not really a DT?)
Paleja et al. (\citeyear{DBLP:conf/rss/PalejaNSRCG22}) present the Interpretable Continuous Control Trees (ICCTs) which represent the policy in a sparse way. %In this way, fuzzy internal nodes and fuzzy leaf nodes are converted.
The interpretable representation consists of internal nodes whose condition is set only by one feature and leaves which are sparse linear controllers (i.e. based on few features). On a set of control tasks tested, ICCTs achieved a good compromise between performance and interpretability.

                    % binary DT (leaf: parametric interpretable policy) Das Gupta
The Policy Tree algorithm (Gupta et al. \citeyear{DBLP:conf/aaai/GuptaTB15}) represents the policy as a tree. This method is applied with agent states described by binary features and the policy is learned by gradient-based methods.
The tree returned by the algorithm is a binary DT where each internal node corresponds to a feature and a leaf to a distribution of actions. 
Starting from any policy and a binary DT containing only one node, two steps are repeated: Parameter Optimisation, which optimises the leaves, and Tree Growth, which extends the tree starting from a leaf.
                    
                    % parameterized (interpretable rules provided) DTs (rule-based policy) Likmeta
For a set of autonomous driving scenarios, Likmeta et al. (\citeyear{DBLP:journals/ras/LikmetaMTGRR20}) propose parametric rule-based policies, visualised in the form of trees.
Such a policy is a parametric set of rules which is learned using policy gradients with parameter-based
exploration (Sehnke et al. \citeyear{DBLP:conf/icann/SehnkeORGPS08}).
               
Although the following works do not aim at interpretability, we think it is interesting to mention them, because of their representation of the Q-table as a tree. Only a few works are presented, to give an overview.
                % Q tree
                    % Q tree (not XAI) Ernst
Ersnt et al. (\citeyear{DBLP:journals/jmlr/ErnstGW05}) present two methods: extra trees and totally randomized trees. The RL batch mode consists of approximating the Q-function based on a set of agent's interactions within the environment. In this context, the two methods are compared with a set of classical supervised learning methods.
                    % Relational RL (not XAI)
                        % Q-learning with relational regression tree Q-DT Dzeroski / Das
In Relational MDP, states, actions and background knowledge are represented by a set of predicates and constants forming ground atoms.
Das et al. (\citeyear{DBLP:journals/corr/abs-2006-05595}) describe a sample efficient approach to approximate the Q-function in the form of a regression tree in a Relational MDP. %This approach is more sample efficient than previous work in the same field.
                        %  Higher OL Q-DT Cole
By providing a set of knowledge about the problem, the Q function and policy are approximated in Cole et al. (\citeyear{cole2003symbolic}). The ALKEMY algorithm (K. S. Ng \citeyear{ng2004alkemy}) is used to learn a set of DT's for each agent task. In addition, a P function is deduced from the Q function, allowing a compact Boolean representation of the policy.

                \paragraph{Graph}
                \label{par:App_A_IU_Graph}
Two works use a neural network based on the structure of a graph.

                % GNN policy: task structure Lin               
Using a total of 20 expert demonstrations, a single Graph Neural Network (GNN) is learned via IL to solve robot manipulation tasks (Y. Lin et al. \citeyear{DBLP:journals/ral/LinWUR22}). The GNN is used to model the underlying structure of the task.
Based on GNNExplainer (Ying et al. \citeyear{DBLP:conf/nips/YingBYZL19}), the explanation consists of determining the neighbouring nodes and edges of the graph that contributed most to the agent's choice.

                % Graph neural network (not XAI) pas sûr de garder Wang
In T. Wang et al. (\citeyear{DBLP:conf/iclr/WangLBF18}), the NerveNet model is proposed to solve continuous control problems. It has been tested on a set of MuJoCo (Todorov et al.  \citeyear{DBLP:conf/iros/TodorovET12}) environments.
The graph structure used for a given robot is based on its skeleton. Thus, two types of nodes are part of the graph: bodies and joints that connect two bodies. 
Learned representations are interpreted using different plots, making it possible to analyse the learned policy.  

The following three works use abstract machines (or automata), assimilated to graphs, to represent the agent's policy.

                % Synthezise programs into state machine policy  Inala
The approach proposed by Inala et al. (\citeyear{DBLP:conf/iclr/InalaBTS20}) consists of learning, in a teacher-student setting, a policy represented in the form of a state machine. This policy can be easily interpreted and modified.
The proposed approach is called adaptive teaching. Alternatively, the student learns to imitate the teacher's behavior, and the teacher learns to provide behavior similar to that of the student. 
The student's policy takes the form of a state machine, which consists of a set of modes (an internal memory state) to which is associated an action function (which is akin to a sub-policy). Transitions between two modes are modelled by switching conditions, which reflect the probability of an agent observation switching from one mode to another. Action functions and switching conditions are represented by programs.

                % Close to HRL synthesized MDP form traces (+ transitions are sub-task) Hasanbeig Produce automata
DeepSynth (Hasanbeig et al. \citeyear{DBLP:conf/aaai/HasanbeigJAMK21}) synthesises a set of sequences of objects detected from agent states into an automaton. Using an image segmentation method, a set of objects is extracted from a state.
With the sequences collected, a deterministic finite automaton (DFA) is produced. This DFA is used to guide the agent in learning its policy. Using both the DFA and the MDP, a Product MDP is generated, where each transition corresponds to a sub-task of the agent.
                
                % Tsetlin machine for interpretable RL (not a success) Varma
In Varma (\citeyear{varma2021interpretable}), a Tsetlin Machine (Granmo \citeyear{DBLP:journals/corr/abs-1804-01508}) is used to learn a policy which is an alternative to the classical NN which uses Tsetlin Automaton to form clauses. Thus, the Regression Tsetlin Machine method (Abeyrathna et al. \citeyear{DBLP:conf/epia/AbeyrathnaGJG19}) is combined with Q-learning. The results show that too many clauses are needed to have a good policy, making the method uninterpretable.

                \paragraph{Equation}
                \label{par:App_A_IU_Equation}

                % Algebraic expression
                    % Genetic prog Hein / Hein
The work of Hein et al. (\citeyear{DBLP:conf/gecco/0001UR19}) proposes to use a genetic programming (GP) approach to find interpretable policies in a model-based RL setting. To do this, a dataset of interactions is collected (using any policy) to learn a world model ĝ. This is used to learn a policy via GP.
This method is compared with two baselines: a NN model-based policy $\pi_m$ using ĝ, and a policy learned via GP based on a dataset of interactions generated by $\pi_m$. 
To measure the complexity of a GP-generated policy represented by a function tree, it is simply necessary to sum the different nodes by weighting them according to pre-defined weights (e.g. an addition costs 1 and the function tanh 4). 
The policies returned by the algorithm have a similar/better performance to the baselines, while being less complex and therefore more interpretable.
                    
                    % auto regressive RNN Landajuela
Deep Symbolic Policy learns interpretable policies for environments where the action space is continuous (Landajuela et al. \citeyear{DBLP:conf/icml/LarmaPKSGMPF21}).
The framework is split into two parts: the Policy Generator, which uses an RNN to iteratively generate an expression tree representing the policy, and the Policy Evaluator, which evaluates the policy and returns the average episode reward. This is used to train the Policy Generator. 
Experiments show that the policies of this approach outperform the 7 DRL baselines, such as \emph{Deep Deterministic Policy Gradient} (Lillicrap et al. \citeyear{DBLP:journals/corr/LillicrapHPHETS15}) and \emph{Trust Region Policy Optimization} (Schulman et al.  \citeyear{DBLP:conf/icml/SchulmanLAJM15}). To handle the multiple action dimensions of certain environments, the authors propose to use a pre-trained policy to `anchor' the symbolic policy during training.
                    
                    % Multi-arm bandit setting to find closed-form formula 
Finding an efficient and interpretable policy, represented by a small formula, is modelled by a Multi-arm bandit problem in Maes et al. (\citeyear{DBLP:conf/dis/MaesFWE12}). The space of policies represented by a formula $F$, such that its size does not exceed a certain threshold (i.e. $|F| \leq K$), is approximated by a clustering method.
A formula of minimum size is extracted from each cluster, resulting in $N$ potential policies, each of them represented via an arm in the Multi-arm bandit setting. The best performing policy is then extracted by solving this problem. 
Compared to previous work described in this section, this approach does not take into account a continuous action space.

                % polynomial function (of state variables) Ault
For a traffic signal control problem, a regulatable precedence function is learned jointly with the DQN (Ault et al. \citeyear{DBLP:conf/atal/AultHS20}).
A precedence function is regulatable if it is monotonic on state features. In this work, such a function is assumed to be interpretable. 
3 variants of DQN are proposed to approximate such a function in addition to learning Q-values.

                \paragraph{Program}
                \label{par:App_A_IU_Program}

                % Synthesize programs Trivedi 
The Learning Embeddings for lAtent Program Synthetis (LEAPS) framework (Trivedi et al.  \citeyear{DBLP:conf/nips/TrivediZSL21}) synthesises a policy in the form of a program based on a domain-specific language. This framework consists of two stages: firstly, LEAPS learns an encoder-decoder that allows a program to be represented in a latent space, and secondly, an iterative search for a latent program to solve a given task.
The latent space is learned in an unsupervised fashion, using randomly generated programs. The aim is to ensure that two programs with similar policies are close in the latent space. 
Finding a good latent program according to the reward obtained is performed based on the Cross Entropy Method (CEM) (Rubinstein \citeyear{rubinstein1997optimization}). To demonstrate the interpretability of the programs generated, a user study shows that users have succeeded in analysing and modifying the program so as to obtain a better reward.

                % Cartesian Genetic prog for Atari Games Wilson (DAG / not XRL)
In Wilson et al. (\citeyear{DBLP:conf/gecco/WilsonCLM18}), simple programs are proposed for playing games in the arcade learning environment (Bellemare et al. \citeyear{DBLP:journals/jair/BellemareNVB13}).
Cartesian Genetic Programming (Miller and P. Thomson \citeyear{DBLP:conf/eurogp/MillerT00}) represents the programs by a directed graph indexed by Cartesian coordinates. The input is an image, and the output is a set of scalars used to select the action to be performed.
This evolutionary approach relies on a set of functions to build an interpretable policy where each internal node corresponds to a function from this set.

                % Programmatic policy (structure searched using MCTS) Gu
For a traffic signal control task, Gu et al. (\citeyear{DBLP:conf/aaai/GuZLGLZ24}) use Monte Carlo tree search to find a program, and Bayesian optimisation (Snoek et al. \citeyear{DBLP:conf/nips/SnoekLA12}) to fine tune its parameters.
Beforehand, a domain-specific language is defined including control flows, conditions and instructions. In addition, four transformation rules are defined to iteratively build a program. 
The best programs found by the algorithm are small in terms of number of instructions and are easily interpretable.

                %\paragraph{Others}                
                
                % Smooth actions interpretable by domain expert Jia context: heating,ventilation and air conditioning control
%For a heating, ventilation and air conditioning control problem, Jia \emph{et al.} \cite{jia2019advanced} propose to learn control policies where learning is stabilized by adding a penalty on abrupt changes of actions. 
%The resulting policy provides smooth actions, which are interpretable by a domain expert.

    %------------------------------------------------
    \subsubsection{Surrogate Model}
    \label{sec:App_A_surrogate_model}
    %------------------------------------------------

                \paragraph{Decision Tree}
                \label{par:App_A_SM_DT}

                % DTs
                    % Oblique DT Dai
A Deep RL (DRL) method is a method that combines RL and Neural Networks (NN) to obtain an agent's policy.
A DRL agent is one that has learned using a DRL method.
To extract an interpretable policy from a DRL agent, Dai et al. (\citeyear{dai2022enhanced}) propose an \emph{information gain rate weighted oblique DT} (IGR-WODT). Based on three defined metrics, IGR-WODT provides a policy that requires a much smaller number of parameters (only 1.1\% of the number of DRL model parameters).
An \emph{oblique DT} is composed of internal nodes that separate data based on linear combinations of features (i.e. of the form $w_1f_1 + w_2f_2 + ... + w_nf_n > 0$ where $w_i$ are weights, $f_i$ are state features and $n$ is the number of features). %\todoflo{tu veux dire feature values?}. 
The WODT (B. Yang et al. \citeyear{DBLP:conf/aaai/YangSG19}), on which IGR-WODT is based, uses a logistic regression model to determine the probability of an instance belonging to the left or right child node. The extension provided by IGR-WODT consists of using a different objective function inspired by the information gain (Quinlan \citeyear{DBLP:journals/ml/Quinlan86}).
                    
                    % Soft DT Coppens

In the same vein, a \emph{soft DT} (SDT) is used to approximate the policy of a DRL agent trained on the Mario AI benchmark (Karakovskiy and Togelius \citeyear{DBLP:journals/tciaig/KarakovskiyT12}) in Coppens et al. (\citeyear{coppens2019distilling}).
An SDT (Frosst and G. E. Hinton \citeyear{DBLP:conf/aiia/FrosstH17}) has a pre-defined depth, where each internal node corresponds to a weighted vector to which a bias is added (i.e. perceptron form) and the leaves correspond to a softmax distribution (which in this case represents the actions distribution).
The advantage of this approach is that the weights learned from each perceptron can be displayed as heatmaps of the agent's state (which takes the form of an image). In this way, the user can successively visualise the pixels that do or do not have an impact on the agent's decision-making. %\leo{An example of agent's choice of action is described in Figure \ref{fig_ch1:SDT.}
A compromise between interpretability and performance is observed: as the depth of the tree increases, the SDT performance gets closer to the 
performance of the DRL agent, at the expense of interpretability. 

    % \begin{figure}[]
    %     \centering
    %     \includegraphics[width=1.0\textwidth]{images/SDT.png}
    %     \caption[Example of use of Soft Decision Tree for Mario AI benchmark]{\leo{Example of use of Soft Decision Tree for Mario AI benchmark \cite{coppens2019distilling}. The letters `R' and `L' between the nodes correspond respectively to access to the right child node or left child node. On the right, the SDT and PPO \cite{DBLP:journals/corr/SchulmanWDRK17} distribution action is shown in blue and orange respectively.}}
    %     \label{fig_ch1:SDT
    % \end{figure}

                % SDT Guo
The same approach is proposed in W. Guo and Wei (\citeyear{guo2022explainable}) for an aircraft separation problem. A visualisation module allows the user to see both the tree plot, where each node is represented by a heatmap, and the trajectory plot, which provides the features that have the greatest impact on the agent's choice of action. %\todoinflo{ici aussi}

                % Ding
In order to outperform the SDT-based approaches, Ding et al. (\citeyear{DBLP:journals/corr/abs-2011-07553}) present a \emph{cascading DT}, an architecture comprising two kinds of trees: a forest of %\todoflo{sachant qu'une forêt particulière ne contient qu'un seul arbre}
learning trees F used to provide a compact representation of features, and a decision-making tree D which uses the features learned by F to learn a policy. This architecture reduces the number of parameters and improves performance compared with other SDT-based methods, e.g. Coppens et al.

                    % non linear DT Dhebar
Another type of DT, the \emph{non linear DT} (NLDT), initially proposed for classification problems (Dhebar and Deb \citeyear{DBLP:journals/tcyb/DhebarD21}), allows a DRL agent to be represented by a more interpretable model (Dhebar, Deb, et al. \citeyear{DBLP:journals/corr/abs-2009-09521}). Such a tree has internal nodes that take the form of conditions on non-linear functions.
Based on a dataset of DRL agent interactions within the environment, the NLDT is learned using an evolutionary approach. 
This tree is then pruned to give an interpretable NLDT and re-optimised for performance.

                    % VIPER imitation learning Bastani / multi-agent extension: MAVIPER Milani
With a concern for reliability, the VIPER algorithm is used to learn a ``robust'' DT that has the same performance as the DRL agent (Bastani, Pu, et al. \citeyear{DBLP:conf/nips/BastaniPS18}). Given a set of state-action situations, imitation learning (IL) (Halbert \citeyear{halbert1984programming}) aims to learn a policy that mimics the given decisions and generalizes over unseen situations. Based on IL and model compression methods, VIPER produces a DT used to verify the policy according to certain properties such as \emph{stability}, \emph{robustness} and \emph{correctness}.
Stability seeks to determine whether the agent asymptotically achieves its objective. The robustness degree of a policy is defined by comparing the action returned for states close to a state $s$ with the one returned for $s$. Correctness varies according to the environment: for the Atari Pong game, the objective is to prove that the policy never loses, and for CartPole, the objective is to prove that the pole never falls below a certain height.
An extension of this work focuses on multi-agent settings. IVIPER (Milani, Z. Zhang, et al.  \citeyear{DBLP:conf/pkdd/MilaniZTSKPF22}) produces a good quality DT for each agent, but does not take into account the coordinated behavior of the agents. To overcome this, MAVIPER (Milani, Z. Zhang, et al.) was introduced. Experiments have shown that MAVIPER generates better and more robust policies than IVIPER and the other baselines tested.

                    % Mixture of expert trees Vasic
Also, in a sort of multi-agent view, the MoËT method (Vasic et al. \citeyear{DBLP:journals/nn/VasicPWNSK22}) is based on \emph{mixture of experts} (Jacobs et al. \citeyear{DBLP:journals/neco/JacobsJNH91}). It consists of a set of expert DT's and a gating function that uses weights to define the extent to which an expert's decision is referred to in the agent's choice of action. The DT's are learned by imitating a DRL agent. The softmax function is used as MoËT's gating function.
A variant called MoËT$_h$ consists of selecting only one DT using the gating function. This variant is used to transform the learned policy into an Satisfiability Modulo Theories (SMT) formula and thus verify properties using the SMT solver Z3 (Moura and Bjørner  \citeyear{DBLP:conf/tacas/MouraB08}). MoËT$_h$ outperforms VIPER (Bastani, Pu, et al.)  in terms of policy fidelity and reward obtained.

                    % pixels -> symbolic features Sieusahai
The method proposed by Sieusahai et al. (\citeyear{DBLP:conf/aiide/SieusahaiG21}) involves learning a DT based not on the image but on a set of \emph{sprites} associated with their position in the image. A sprite is an image representing a character or object within a game. To learn the DT, a function transformation is used to extract a set of sprites with their coordinates from the image representing the state of the DRL agent.
The DT learns to approach the DRL agent based on this more compact representation of the agent's state, which makes the internal nodes of the tree interpretable.
                                
                    % Imitation Learning: DTs (random forest) Cichosz
In Cichosz and Pawelczak (\citeyear{DBLP:journals/amcs/CichoszP14}), DT's and {Random Forests (RF) are learned in The Open Racing Car Simulator (TORCS)  environment\footnote{\url{https://sourceforge.net/projects/torcs/}} based on IL.
The continuous actions are vectors composed of the steering, acceleration, brake and gear parameters in TORCS.
RF generalise better and perform better, at the expense of interpretability, than DT's. 
In order to learn such models, a number of modifications had to be made to the problem: the decomposition of the multidimensional actions problem into a set of one-dimensional actions and the discretization of the continuous actions.        
                    
                    % Tree interpretation from object representation Liu
A more user friendly approach uses a tree of objects (G. Liu, Sun, et al. \citeyear{DBLP:conf/nips/LiuSSP21}). The RAMi framework  is used to learn a tree that mimics a DRL agent based on a representation of states by objects, where an object is a sub-part of an image. This representation is learned using the Identifiable Multi-Object Network and then, the tree is learned using a Monte Carlo Regression Tree Search algorithm.
It has a good compromise between fidelity to the agent and simplicity in terms of the number of nodes in it. %An internal node of the tree represents a threshold of a variable for an object present in the image (i.e. state) of the agent. 
In addition, a method is proposed for calculating the importance of features and obtaining causal relationships from the mimic tree. 

                    % Extract strategy labels and conditions (conjunction of relevant features) from trajectories Acharya
In the similar aim to produce interpretable nodes, Acharya et al. (\citeyear{DBLP:journals/corr/abs-2011-09004}) propose an algorithm to generate a set of strategy labels and interpretable experiential features representations, i.e. predicates, from a set of sequences of interactions of the DRL agent within the environment.
Using these elements, a DT is constructed where an internal node uses a generated predicate. The tree is used to represent the agent's different strategies. 
The condition for following a certain strategy is thus represented by a branch (which is a conjunction of interpretable features).

                    % Decisions rules displayed via flowchart Skirzynski 
The algorithm based on IL and program induction proposed in Skirzynski et al.  (\citeyear{DBLP:journals/ml/SkirzynskiBL21}) is used to cluster agent interaction data into a set of rules expressed by a set of predicates.
The output of the algorithm returns a set of DT's, which are displayed to the user in the form of flowcharts. 
These flowcharts represent the predicates of the internal nodes of the DT's in the form of natural language questions.
                    
            % Linear model trees
                    % Carbone (from DDPG policy + comparison with SHAP)
                    % Lover (from PPO + comparison with SHAP/LIME)
Linear Model Trees (LMT) are used to approximate the policy directly. The leaves are linear models representing an action to be taken. LMT's have been compared in (Carbone \citeyear{carbone2020explainable}; Løver et al. \citeyear{lover2021explainable}) with the model-agnostic methods SHAP (Lundberg and S. Lee \citeyear{DBLP:conf/nips/LundbergL17}) and LIME (Ribeiro et al. \citeyear{DBLP:conf/kdd/Ribeiro0G16}).
The main advantage of LMT's is the lower computational cost, as the model is learned in advance. To explain an action choice, LMT's allow the influence of features %\leo{(on the choice of action)} 
to be extracted directly, whereas other approaches require the influences to be calculated. 
Also, representing the policy by an LMT makes it directly interpretable, to the extent of the size of the tree. As with most other work, a trade-off between simplicity and fidelity is necessary.

                    % Linear Model U-Trees Liu (extension of CUTS Uther)
For a state $s$, the value function represents the expected return starting from $s$.
For a state-action pair $(s,a)$, the Q function represents the quality, or expected return, of executing $a$ from $s$. Q-learning aims to learn the optimal Q-function by iteratively updating the Q-values based on the Bellman equation. Deep-Q-Network (DQN) (Mnih, Kavukcuoglu, et al. \citeyear{DBLP:journals/nature/MnihKSRVBGRFOPB15}) is a method that learns to approximate Q-values using NNs.
G. Liu, Schulte, et al. (\citeyear{DBLP:conf/pkdd/LiuSZL18}) introduce Linear Model U-trees (LMUTs) to approximate the Q-values of a DQN agent. This method differs from the other approaches previously presented in this section, where the aim is not to mimic the policy directly.
LMUTs is an extension of CUTs (Uther and M. M. Veloso \citeyear{DBLP:conf/aaai/UtherV98}), which are regression trees for value functions. Compared to CUTS, LMUTs has leaf nodes that contain linear models of state features.
The model is interpreted in three ways: by identifying the importance of the splitting features in the LMUT, by extracting the rule that led to the Q-values obtained and by displaying the pixels that have an above-average influence on the Q-values.

                    % Q-value based trees Jhunjhunwala
With the same idea of approximating Q-values, Jhunjhunwala et al. (\citeyear{DBLP:conf/ijcnn/JhunjhunwalaLSA20}) proposes a new structure, called Q-BSP Tree, which models the Q-value in an interpretable way and an Ordered Sequential Monte Carlo algorithm which learns to approximate the Q-value of the DQN.
For a problem comprising $n$ actions, the Q-BSP forest is composed of $n$ Q-BSP trees, each learning the Q-value of one action. This approach outperforms the baselines in terms of performance and fidelity with respect to the DQN agent policy, including LMUT (G. Liu, Schulte, et al.).

                    %\todoinflo{faire un tableau récapitulatif pour les DT: nom de la méthode, structure utilisée}

                \paragraph{Graph}
                \label{par:App_A_SM_Graph}

                % Abstract policy
                    % Abstracted policy graph Topin / Vila (questions similar to Hayes and Shah)
Topin and Veloso (\citeyear{DBLP:conf/aaai/TopinV19}) present \emph{abstracted policy graphs} (APG), which are Markov chains of abstract states. 
In an APG, edges are transitions labelled by their probability of occurring, nodes are abstract states labelled by the agent action performed from that state. An abstract state represents a grouping of similar states that is constructed with respect to the actions and feature importance of the states, calculated via FIRM (Zien et al. \citeyear{DBLP:conf/pkdd/ZienKSR09}).
The transition probability from node $S$ to node $S'$ is an average of the transition probabilities on the states represented by $S$ and $S'$. To build the APG, a policy, a value function and a set of data on the agent's interactions within the environment are required. An example of APG is described in Figure \ref{fig_ch1:APG}.

%    \begin{figure}[]
%        \centering
%        \includegraphics[width=0.8\linewidth]{images/APG.png}
%        \caption[Example of Abstract Policy Graph]{Example of Abstract Policy Graph \cite{DBLP:conf/aaai/TopinV19}. Each vertex is an abstract state with an associated action, and the edges between two vertices represent transitions (weighted by their probability of occurring).}
%        \label{fig_ch1:APG}
%    \end{figure}
                    
                    % CAPS McCalmon directed graph
In the same vein, McCalmon et al. (\citeyear{DBLP:conf/atal/McCalmonLA022}) propose CAPS, a method that constructs a directed graph composed of abstract states.
The state space is abstracted using a decision tree-based clustering algorithm called CLTree (B. Liu et al. \citeyear{liu2005clustering}). This algorithm returns a hierarchy of cluster configurations, the best of which is extracted by taking into account the interpretability of the clusters and the accuracy of the state transitions.
To improve user comprehension, the authors use an approach to highlight the most important abstract states according to the agent's stochastic policy (S. H. Huang, Bhatia, et al.  \citeyear{DBLP:conf/iros/HuangBAD18}) and a method to describe each abstract state by a natural language sentence.

                    % Vila
Based on a set of agent's interactions within the environment, the Policy Graph proposed in Domenech i Vila et al. (\citeyear{domenech2024explaining}) is a directed graph that represents the policy with states expressed via predicates (a state is a formula in propositional logic). These handcrafted predicates are defined upstream according to the environment.
In the graph, the nodes represent the states, the edges represent the transitions with their probability and the agent's action. 
With this graph, the authors propose to answer three questions (inspired by Hayes and Shah  \citeyear{DBLP:conf/hri/HayesS17}) using custom algorithms: \emph{`What will you do when you are in state region X?'}, \emph{`When do you perform action a?'} and \emph{`Why did you not perform action a in state s?'} This approach was tested on a single-agent and a multi-agent environment.
                    
                    % MARLeME (MARL context): Value-based Abstract argumentation (represented as directed graph) Kazhdan
MARLeME (Kazhdan et al. \citeyear{DBLP:conf/ijcnn/KazhdanSL20}) is a library that fits into a Multi-Agent RL (MARL) setting to create different directed graphs from agent policies, each representing a Value-based Argumentation Framework (VAF). A VAF is used to model a set of action arguments and the attack relations between them.
These arguments are valued in such a way as to order the arguments according to their usefulness. MARLeME can be used to represent an agent, or a group of agents, in the form of a VAF. To use this library, it is necessary to enter the arguments, their valuations and the attack relations. %\todoinflo{mettre un schéma issu de l'article précédent}
                    
                    % Verification (build predicate abstraction graph) Vinzent
In order to verify safety properties of a DRL agent, Vinzent et al. (\citeyear{DBLP:conf/aips/VinzentS022}) construct an abstract graph based on predicates. Abstraction is only performed for observable states, taking as input the agent's policy and the predicates. %\flo{\st{are an input to the method}}.
Abstract states are constructed on the basis of satisfying the set of predicates (so that states satisfying the same predicates belong to the same abstract state). An SMT solver, Z3 (Moura and Bjørner \citeyear{DBLP:conf/tacas/MouraB08}), is used to define the transitions in the graph. %\flo{\st{To avoid numerous costly calls, the authors describe a set of algorithmic enhancements.}}\todoflo{n'apporte rien}

                % Compact Moore Machines
The following work focuses on transforming a policy that takes the form of a Recurrent NN (RNN) into a Moore Machine (MM) (Moore \citeyear{moore1956gedanken}), which is a finite state machine whose output depends only on the current state and where each state is labelled by an output value (an action in this context). %\flo{(a Moore Machine (MM) being ...)}\todoflo{à définir + une image?}.
                    % RNN -> Compact Moore Machines ('Switching and finite automata theory' Kohavi)
                    % Koul
To do this, Koul et al. (\citeyear{DBLP:conf/iclr/KoulFG19}) first propose an auto-encoder called Quantized Bottleneck Network (QBN), which learns to quantize the agent's observations and the RNN's memory states for a MM representation.
Next, the QBN is combined with the RNN to create a Moore Machine Network (MMN). 
Finally, a MM is extracted from the MMN and minimised to be interpretable. 
In the MM, the nodes represent the states of the memory, and the edges represent the agent's observations. Each node is labelled with the action to be performed from a memory state. In some environments, it has been observed that the number of memory states and observations to represent the agent's policy is small.  
                    % Danesh
An extension of this work, Danesh et al. (\citeyear{DBLP:conf/icml/DaneshKFK21}), criticises the interpretability of the MM provided and proposes four ways of reducing the size of the MM in an interpretable way. Indeed, in Koul et al. the minimisation techniques used do not take into account the user's comprehension, for example certain states judged to be different for a user can be merged into a single state.
The proposed minimisation methods improve the visualisation of the MM, while preserving the key decision points of the agent's behavior. A tool based on the saliency map principle (presented later in this survey) is proposed as a complement to understand the differences in the agent's choice of action between 2 observations. 

                    % Imitation learning agent's policy + belief update process Huyuk (decision boundaries)
Based on a set of sequences of interactions of the agent within the environment, Hüyük et al. (\citeyear{DBLP:journals/corr/abs-2310-19831}) present a Bayesian algorithm for learning decision dynamics and decision boundaries.
Decision dynamics aggregate sequences as beliefs, which correspond to the probability that a state $s$ exists at a time $t$. Decision boundaries represent policy by partitioning decision dynamics into regions where the same action is performed.

                    % Complementary agent based on set of rules with simple inner mechanisms Lee
The quasi-symbolic (QS) agent (J. H. Lee \citeyear{DBLP:journals/corr/abs-1901-00188}) is composed of two interpretable networks made up of a single layer: the matching network, which memorises transitions, and the value network, which evaluates them.
This agent uses the RL agent and the environment function to predict a plan of actions that leads to one of the most valuable transitions (in terms of reward). The structure of the QS agent makes its behavior interpretable.

                \paragraph{Program}
                \label{par:App_A_SM_Program}

                % Programmatic policy   
                    
                    % framework PIRL Verma / find a policy that fit a syntactic sketch algo NDPS
In order to design interpretable and verifiable policies, the \emph{programmatically interpretable reinforcement learning} (PIRL) framework is proposed in Verma, Murali, et al.  (\citeyear{DBLP:conf/icml/VermaMSKC18}). Policies are generated using a environment-specific programming language.
To compensate for the size of the policy search-space, the specification of the policy form is used in PIRL through a sketch, which is a restriction of the grammar proposed by the language.
The Neurally Directed Program Sketch (NDPS) algorithm is used to locally search for a policy that takes the form of a program based on the NN agent policy, a Partially Observable MDP (POMDP) and a sketch. POMDP (Åström \citeyear{aastrom1965optimal}) is a generalisation of MDP where the agent does not have full visibility of the state it is in.
NDPS returns a policy close to that of the DRL agent and which obtains the highest expected aggregated reward among the programs evaluated during the search.

                    % Zhu close to NN + respect a desired logical property
In the same vein, Zhu et al. (\citeyear{DBLP:conf/pldi/ZhuXMJ19}) synthesise a program close to the DRL agent policy which avoids reaching unsafe states of the environment. To this end, the \emph{counterexample-guided inductive synthesis} (CEGIS) framework is presented to search for a program parameterised by a sketch (i.e. a high-level description of the sought program). %\todoflo{préciser ce qu'est ce sketch}.
In this approach, a counterfactual example is an initial state in which the program under study has not yet been proven safe. 
In addition to the interpretability aspect, the program produced is used as a safety shield: it takes the relay when the agent chooses an action that may lead to an unsafe region.

                    % Delfosse
Based on a DRL agent policy, INTERPRETER (Kohler, Delfosse, et al. \citeyear{DBLP:journals/corr/abs-2405-14956}) is used to obtain an interpretable and editable python program. To do this, a regularized oblique tree (Murthy et al. \citeyear{DBLP:journals/jair/MurthyKS94}) is learned from the DRL agent policy via IL, then converted into a python program, notably by translating the nodes of the tree using \emph{if-else} statements.
                    
                    % Burke context: robot manipulation tasks (from robot demonstrations)
For robot manipulation tasks, a program is extracted from a set of demonstrations in Burke et al. (\citeyear{DBLP:conf/rss/BurkePR19}). To do this, a probabilistic generative task model is used to infer from the demonstrations a sequence expressed by interpretable low-level motion primitives.
Program induction is then used on this sequence to simplify it into a program.

                    % PROPELPROG Verma other algorithm analogue to NDPS % PROPELTREE Verma
The PROPEL approach (Verma, H. M. Le, et al. \citeyear{DBLP:conf/nips/Verma0YC19}) makes it possible to discover policies that are interpretable, verifiable and generalisable. The idea is to iteratively search for a policy represented by a NN using policy-gradient methods and to project the policy into the space of programmatic policies (i.e. policies representable by programs).
Two PROPELs searching for policies in different programmatic policy classes have been proposed: PROPELPROG, which uses a environment-specific language, like NDPS (Verma, Murali, et al.) and PROPELTREE, which uses tree regression like VIPER (Bastani, Pu, et al.  \citeyear{DBLP:conf/nips/BastaniPS18}).

                    % Program synthesis Bastani Overview
For a detailed overview of program synthesis for interpreting DRL agents, we recommend Bastani, Inala, et al. (\citeyear{DBLP:conf/icml/BastaniIS20}) to the interested reader.
%\todoinflo{petit tableau récapitulatif des 4 méthodes proposées}

                \paragraph{Rules}
                \label{par:App_A_SM_Rules}

                % Rule-based 
                    % Fuzzy Rules Nageshrao
In Nageshrao et al. (\citeyear{DBLP:conf/icmla/NageshraoCF19}), a set of fuzzy rules is learned to approximate an agent's policy. Based on a dataset representing the agent's interactions with the environment, the rules are learned using the Evolving Takagi-Sugeno method (Angelov and Filev \citeyear{DBLP:journals/tsmc/AngelovF04}). To obtain these rules, it is necessary to learn the rule antecedents, which decompose the state space into relevant regions, and to estimate the feature coefficients of the rule consequents, which are linear model of each rule used for determining the action to be taken.
This approach allows the policy to be approximated by a set of conditioned interpretable linear models.

                    % Fuzzy rules vizualisation Soares
Soares et al. (\citeyear{DBLP:journals/tfs/SoaresACCNF21}) propose prototype-based fuzzy rules which are then visualised by displaying either features or prototypes. To reduce the number of fuzzy rules generated, the authors propose a hierarchical mechanism to group adjacent prototypes in the data space that have the same consequent (i.e. action).

                \paragraph{Equation}%\todoinflo{à regrouper avec rules? "Rules or Equations"}
                \label{par:App_A_SM_Equation}
The substitute policy takes the form of equations in the following two works. It is in a context where the agent's action is determined as a function of continuous features of the agent's state. Genetic algorithms are used to find such policies.

                    % Genetic programming tree-based (using evolutionary feature synthesis) (algebraic expression) Zhang
Zhang et al. (\citeyear{zhang2020interpretable}) use the evolution feature synthesis (Arnaldo  et al. \citeyear{DBLP:conf/gecco/ArnaldoOV15}) to generate control policies that mimic the DRL agent policy. A generated policy is a combination of operations (defined on a set of operators O) using reals and features of a state.
In order to make the policy interpretable, a method based on the complexity-performance trade-off is proposed.
                    
                    % Symbolic regression based on gen prog to get equation Kubalik
In the same line, the method presented in Kubalík et al. (\citeyear{kubalik2017optimal}) is a variant of Single Node Genetic Programming (Jackson \citeyear{DBLP:conf/eurogp/Jackson12}), which represents the individuals in a population by a node in a graph. For a robotic task, this approach approximates a policy that produces smooth control by an equation, based on a set of interactions of the agent within the environment.%\todoinflo{on ne voit pas le rapport avec des équations + single node genetic programming n'a pas été introduit}

%------------------------------------------------
%------------------------------------------------
\subsection{Policy Summary}
\label{sec:App_A_policy_summary}
%------------------------------------------------
%------------------------------------------------

    %------------------------------------------------
    \subsubsection{Sequences}
    \label{sec:App_A_summary_sequences}
    %------------------------------------------------
            % Trajectories   

                % HIGHLIGHTS Amir 
The HIGHLIGHTS algorithm (D. Amir and O. Amir \citeyear{DBLP:conf/atal/AmirA18}) provides a set of sequences that summarise the agent's behavior. To select sequences, Amir and Amir focus on important states using the notion of \emph{state importance} (Clouse  \citeyear{clouse1996integrating}) (which uses the agent's Q-values). From the important states, sequences are created to provide the user with more context. Thus, a sequence includes states that precede the important state, the state itself, and states that follow it. In the same work, the HIGHLIGHTS-DIV algorithm is proposed to provide more varied sequences as a summary.
Several studies have used HIGHLIGHTS to provide explanations: Huber, Weitz, et al. \citeyear{DBLP:journals/ai/HuberWAA21}; Septon et al \citeyear{DBLP:conf/paams/SeptonHAA23}.
                % HIGHLIGHTS + Saliency maps Huber
Huber et al. use HIGHLIGHTS-DIV coupled with saliency maps to provide additional information.
A saliency map makes it possible to determine which part of the state (which takes the form of an image) is responsible for the agent's choice of action.
                % Contrastive HIGHLIGHTS + Reward decomp Septon (additional element)
In the same vein, Septon et al. combine HIGHLIGHTS-DIV with the reward decomposition approach.
This method decomposes the reward function into several interpretable objectives, making it possible to determine which goal(s) the agent is seeking to maximise by performing a certain action. 
Saliency maps and reward decomposition will be described in more detail in Section \ref{sec:saliency_maps} and Section \ref{sec:reward_decomp}  respectively.
                % Counterfactual action outcomes Amitai
Amitai et al. (\citeyear{DBLP:conf/aaai/AmitaiSA24}) propose the CoVIZ algorithm to compare the outcomes between the action $a$ taken by the agent from a state $s$ and a counterfactual action $a'$.
A variant of this algorithm is described for extracting a summary of sequences. In this case, the agent's sequences are proposed at the same time as the counterfactuals. These counterfactual sequences start with a counterfactual action $a'$, then follow the agent's policy. 
The pairs of sequences that contrast the most in terms of outcomes (based on the value function) are displayed to the user.  

                % Interestingness elements Sequeira / Sequeira with behavior clustering and local/global explanations using SHAP
Sequeira and Gervasio (\citeyear{DBLP:journals/ai/SequeiraG20}) propose a different approach to providing a summary. During the agent's interaction with the environment, a set of data is collected to extract information about, for example, the frequency of situations encountered or the agent's confidence in these decisions. 
These Interestingness Elements are used to provide the user with a visual summary.
                % Use of interestingness elements (+reward decomp) Feit
In Feit et al. (\citeyear{DBLP:conf/acsos/FeitMP22}), this approach was used in conjunction with the reward decomposition method.
                % Interestingness elements Sequeira / Sequeira with behavior clustering and local/global explanations using SHAP
Interestingness Elements has been extended by the IxDRL framework (Sequeira and Gervasio  \citeyear{DBLP:conf/xai/SequeiraG23}) which contains new interestingness elements to analyse, a method for clustering trajectories based on interestingness elements and the use of SHAP (Lundberg and S. Lee \citeyear{DBLP:conf/nips/LundbergL17}). This model-agnostic method is used to determine the impact of each feature of a state on the agent's decision. SHAP will be described in more detail in Section \ref{sec:policy_SHAP}.

                % (Robust mental model) policy Summarization Lage / Lage
Another approach (Lage et al. \citeyear{DBLP:conf/ijcai/LageLDA19,DBLP:conf/atal/LageLDA19}) shows the impact of the user's mental model and the model used to create the summary on the user's reconstruction of the agent's policy. Lage et al. compare IL with Inverse RL (IRL) (A. Y. Ng and Russell \citeyear{DBLP:conf/icml/NgR00}).
As a reminder, given a set of state-actions situations, IL aims to learn a policy that mimics the given decisions and generalizes over unseen situations. 
IRL consists of extracting a reward function based on a set of sequences that matches the observed behavior of the agent.
                % Illustrative examples using human inference models Huang
In order to provide informative sequences to communicate the agent's goal, S. H. Huang, Held, et al. (\citeyear{DBLP:journals/arobots/HuangHAD19}) use an algorithmic teaching approach to model users' mental models.

                % Study impact of env. shift Frost
Frost et al. (\citeyear{DBLP:journals/corr/abs-2201-12462}) use an exploratory policy to reach a set of various states not seen during the training phase. Sequences are created from each of these states using the agent's policy.
The aim is to provide an informative summary of the agent's behavior in the case of a state distribution shift that may occur during the test phase.
                % generate diverse populations of demonstratations using evolutionary algorithm Altmann
In the same idea, Altmann et al. (\citeyear{DBLP:journals/corr/abs-2404-03359}) use an evolutionary algorithm to perturb the initial state and obtain a varied set of sequences.

                % Interpret / Correct Off policy evaluation by identifying highly influential datapoint to remove/modify Gottesman
In order to learn a policy based on a dataset, Gottesman et al. (\citeyear{DBLP:conf/icml/GottesmanF0PCBD20}) propose an efficient approach to identify the transitions in the dataset that are most important in the agent's learning.
The idea is that removing an influential transition has a big impact on learning. 
To determine the influence of a transition $t$, a disparity of predictions is performed between a critical function that has learned with the entire dataset and a critical function that has learned without $t$.

    %------------------------------------------------
    \subsubsection{Critical States}
    \label{sec:App_A_critical_states}
    %------------------------------------------------

An agent's policy can be summarised simply as a set of so-called critical states in the agent's interactions with the environment.

                % Appropriate trust via crit. states Huang
In this sense, Huang et al. (\citeyear{DBLP:conf/iros/HuangBAD18}) provide the user with a set of critical states of a policy as an explanation. This provides a restricted view of the agent's capacities on critical states, i.e. states where the Q-value varies greatly between different possible actions.

                % Find weaknesses (generated critical states) Rupprecht (what if ?)
Based on a generative model producing an agent's state and a user-defined objective function, the method proposed by Rupprecht et al. (\citeyear{DBLP:conf/iclr/0001IP20}) can be used to identify and visualise agent weaknesses. It has been tested on Atari games, where the state is an image. With this approach, it is possible, for example, to look at a given action $a$ by generating states in which the agent assigns a large Q-value to $a$, then seeing whether this action is coherent from the states generated.

                % TldR find landmarks (sub goals) and build graph Sreedharan
In the context of MDP, Sreedharan et al. (\citeyear{DBLP:conf/aips/SreedharanSK20}) propose a policy summary in the form of milestones, called landmarks, that the agent seeks to reach. The summary is a partial order of landmarks.
For the exploding blocks-world problem (Younes \citeyear{younesexploding}), an ordering of landmarks is described in Figure \ref{fig_ch1:TLdR}.
The identification of landmarks is done using compilation procedures and the ordering between these landmarks using methods for reasoning about task hierarchies in MDP's.

%    \begin{figure*}[]
%        \centering
%        \includegraphics[width=1.0\linewidth]{images/TldR.png}
%         \caption[Example of the use of TLdR]{Landmarks and their ordering obtained using TLdR \cite{DBLP:conf/aips/SreedharanSK20} for the exploding blocks-world problem \cite{younesexploding}. Vertices are landmarks and each blue edge defines an order between two landmarks. The green nodes are goals that the agent must reach.}
%        \label{fig_ch1:TLdR}
%    \end{figure*}
    
                % Identify key events via reward redistribution Dinu (no graph but sub goals
Reward redistribution is the process of creating a dense reward function for a given problem (Arjona-Medina et al. \citeyear{DBLP:conf/nips/Arjona-MedinaGW19}) (as opposed to a sparse reward function). Using this, Dinu et al. (\citeyear{DBLP:conf/icml/DinuHPDBBAH20}) present a method for extracting key events from an agent's behavior.
An event corresponds to a cluster of state differences. The agent's policy is then expressed as a sequence of key events.

                % 'Summarize' training by recording most important moments (state-action) Mishra
                % 'Summarize' training by recording most important moments (state-action) Dao + Extension using Grad CAM Dao ('Learning Sparse Evidence- Driven Interpretation to Understand Deep Reinforcement Learning Agents')
Other works are particularly interested in summarising the agent's training: Dao, Huff, et al. \citeyear{DBLP:conf/ssci/DaoHL21}; Dao, I. Mishra, et al. \citeyear{DBLP:conf/icmla/DaoML18}; I. Mishra, et al. \citeyear{DBLP:conf/ssci/MishraDL18}.
The idea is to save relevant tuples of snapshot images (i.e. states), actions, and weight distributions in memory to keep the agent's important learning experiences. The weight distribution associated with each snapshot-action pair corresponds to the impact that snapshot has on the development of the policy. 
Important snapshots can then be viewed to understand the agent's decisions. In Dao, Huff, et al., saliency maps are generated to reduce the number of important snapshots using Grad-CAM (Selvaraju et al. \citeyear{DBLP:conf/iccv/SelvarajuCDVPB17}).
                
                % Formal verification: generate counter state example (not really policy summary) Kazak
As a verification framework for DRL, Verily (Kazak et al. \citeyear{DBLP:conf/sigcomm/KazakBKS19}) allows the user to check whether a policy satisfies a certain requirement. This framework combines scalable model checking and formal NN verification methodologies. This verification is carried out at the level of the states of an agent. If the requirement is not met, a counter-example is proposed (in the form of a state).

    %------------------------------------------------
    \subsubsection{SHAP}
    \label{sec:App_A_policy_SHAP}
    %------------------------------------------------

       % SHAP (Kernel SHAP and DeepSHAP Lundberg) 
This work is based on the model agnostic method, called SHAP (Lundberg and S. Lee \citeyear{DBLP:conf/nips/LundbergL17}). It is interesting to note that this method is mainly used to explain classifiers.
As already mentioned, it consists of determining the impact of each feature of an input on the output of the model. In our context, the input is a state, the output an action, and the model the agent's policy. This method is an approximation to the computation of Shapley values (Shapley et al. \citeyear{shapley1953value}).
Originally from the field of game theory, Shapley values are used to determine the extent to which a member of a coalition contributes to the final value obtained. To determine Shapley values, it is necessary to calculate for each player $m$ of the game, the set of possible coalitions $c$ containing $m$, and to see the impact of its removal from $c$ on the final value.
SHAP proposes different approximations to Shapley values, which consist in limiting the number of coalition samples. 
Two approximations have been used in the work presented in this survey: Kernel SHAP, a model-agnostic estimation, and DeepSHAP, an estimation specific to neural networks. Kernel SHAP is based on Linear LIME (Ribeiro et al. \citeyear{DBLP:conf/kdd/Ribeiro0G16}) (this method will be described in Section \ref{sec:App_A_inspection}) and DeepSHAP on DeepLIFT (Shrikumar et al.   \citeyear{DBLP:journals/corr/ShrikumarGSK16}).
As this section is concerned with the explanation of policies, SHAP is used globally: it identifies the impact of each feature on the agent's policy.

            % DeepSHAP (+local explanation) Zhang
            % Kernel SHAP (+local explanation) Wang
There are two works that use the method as is: Zhang et al. (\citeyear{DBLP:journals/tcss/ZhangZXGG22})   use the DeepSHAP method in the context of power system emergency control and Wang et al. (\citeyear{DBLP:conf/aaaiss/WangME20}) use Kernel SHAP for a machine control use case (an example of SHAP is shown in Figure \ref{fig_ch1:SHAP}).

%    \begin{figure}[]
%        \centering
%        \includegraphics[width=0.6\linewidth]{images/Global_SHAP_Wang.png}
%        \caption[SHAP for a machine control use case]{SHAP for a set of states along an episode for a machine control use case \cite{DBLP:conf/aaaiss/WangME20}. %The base state used for the calculation is the start position, i.e. the initial state. 
%        The evolution of the importance score of each feature is represented by a colored curve in the chart, this importance varying over time.}
%        \label{fig_ch1:SHAP}
%    \end{figure}

            % XGBoost Sequeira interpretable features need to be set
In the ixDRL framework (Sequeira and Gervasio \citeyear{DBLP:conf/xai/SequeiraG23}), SHAP values are calculated using XGBoost machines (T. Chen and Guestrin \citeyear{DBLP:conf/kdd/ChenG16}) to simplify the analysis.
            % Beechey 
Beechey et al. (\citeyear{DBLP:conf/icml/BeecheySS23}) distinguish their approach from previous work by proposing a general framework for the use of shapley values in the context of RL. They also propose a method for providing global explanations of the agent's performance, i.e. the expected return.

            % Monte Carlo approx. Shapley v. for coop MARL Heuillet
In the context of cooperative MARL, Heuillet et al. (\citeyear{DBLP:journals/cim/HeuilletCR22}) compute SHAP values using Monte Carlo sampling. This use of SHAP is closer to the basic principle of Shapley values. In fact, this method makes it possible to determine the contribution of each agent.
            % Compare Shapley values and Myerson values using hierarchical knowledge graph as prior knowledge (Angelotti)
            % individual attribute and policy contribution in MARL context
In the same context, Angelotti and Diaz-Rodriguez (\citeyear{angelotti2023towards}) compare the calculation of Shapley values with that of Myerson values (Myerson  \citeyear{DBLP:journals/mor/Myerson77}), preceded by the construction of a Hierachical Knowledge Graph (HKG) representing the agents and their features. Myerson values are an equivalent of Shapley values specific to cooperative games constrained by a graph. They are therefore calculated on the basis of the HKG. In this work, the contribution of agents' policies and their features are studied jointly. The computation of Myerson values leads to a significant time saving compared to Shapley values.

    %------------------------------------------------ 
    \subsubsection{Policy Comparison}
    \label{sec:App_A_policy_comparison}
    %------------------------------------------------

        % Performance-based comparison Amitai
The DISAGREEMENTS algorithm (Amitai and O. Amir \citeyear{DBLP:conf/aaai/AmitaiA22}) shows a visual summary of the most important disagreements between the two agents. The agents' Q-values are used to determine their disagreement. The summary takes the form of a set of various trajectories.
        % Preference-based comparison Gajcin
This disagreement is measured in terms of ability whereas Gajcin, Nair, et al. (\citeyear{DBLP:journals/corr/abs-2112-09462}) focuses on a disagreement based on preferences. Indeed, in a problem where different winning strategies may exist, an agent may prefer one strategy over another. Based on a set of disagreements obtained between two policies, the method determines the type of states (in terms of features) that the different agents seek to achieve.

%------------------------------------------------
%------------------------------------------------
\subsection{Human-readable MDP}
\label{sec:App_A_human-readable_MDP}
%------------------------------------------------
%------------------------------------------------

   %------------------------------------------------
    \subsubsection{Inherently Understandable}
    \label{sec:App_A_mdp_inherently_understandable}
    %------------------------------------------------
            \paragraph{MDP Representation.}
            \label{par:App_A_MDP_Representation}

This section groups works that represents the MDP, especially the dynamics of the environment (i.e. the reward function and the transition function) in an interpretable way.

                % Reward
In the following work, interpretability is based solely on the reward function. 

                    % Global reward decomposition into each region of the state space + heatmaps of Q-values 
                    % Saldiran
In a close air combat context, the reward decomposition (Erwig et al.  \citeyear{erwig2018explaining}) is used globally to determine in each tactical region the type of reward that is dominant (Saldiran et al. \citeyear{saldiran2024towards}). A tactical region is defined beforehand and represents a set of states.
In addition, a visualisation method is used to analyse the impact of the different types of reward on the different tactical regions. This representation takes the form of heatmaps of normalised Q-values.
                    
                    % IRL context weighted components Bica Learn a reward function 'weighted sum over potential outcomes
The reward function learned in Bica et al. (\citeyear{DBLP:conf/iclr/BicaJHS21}) is a weighted sum over potential outcomes, i.e. changes in features of a state. The approach is in an IRL batch setting, where the objective is to learn the reward function based on a dataset of expert policy trajectories.
The weights of the different outcomes are used to determine their influence on the agent's choice of action. The main limitation of this approach is that it assumes that the reward function is a linear function over the features.

                    % Transform reward into simpler one to assess it. Context: evaluate a learned reward function? Jenner
In Jenner and Gleave (\citeyear{DBLP:journals/corr/abs-2203-13553}), the idea is to transform a reward function $r$ into a more interpretable function $r'$. To use this framework, it is necessary to instantiate an equivalence relation between two reward functions and a cost function measuring the interpretability of a function.
For example, the cost function can take into account the sparsity or smoothness of a reward function. 
In a gridworld-type environment, additional pre-processing is carried out to make the generated function more interpretable, which is then displayed in the form of a heatmap.
                    
                    % Create a sparse reward function Devidze
EXPRD (Devidze et al. \citeyear{DBLP:conf/nips/DevidzeRKS21}) designs interpretable reward functions by making a compromise between informativeness and sparseness. This function is learned from a dense MDP reward function.
The interpretable reward function must satisfy an invariance requirement, i.e. the optimal policy obtained with this function must belong to the set of optimal policies of the base MDP. 
To construct such functions in a large state space, a state abstraction method is proposed.  
                    
                    % Learn interpretable reward / form: Deep Neural DT Srinivasan
The Adversarial IRL algorithm (AIRL) (Fu et al. \citeyear{DBLP:journals/corr/abs-1710-11248})  is extended by Srinivasan and Doshi-Velez (\citeyear{srinivasan2020interpretable}) to build more interpretable reward functions. AIRL consists of using a discriminator to help learn a reward function from which the learnt policies are close to expert demonstrations.
The interpretability aspect of the reward function is added using the Deep Neural DT architecture (Y. Yang et al. \citeyear{DBLP:journals/corr/abs-1806-06988}), which represents in tree structure the different feature valuations (called binning layer in NN), before using a decision layer to provide a reward. This method obtains smooth and sparse reward functions.
                  
                    % Preference-based RL context: build a tree reward function Bewley
Preference-based RL (PbRL) consists of solving a problem modelled by an MDP where the reward function is not provided. A fitness function reflecting the user's preference on pairs of trajectories is used to learn a reward function upstream of the agent's policy. 
In Bewley and Lécué (\citeyear{DBLP:conf/atal/BewleyL22}), the proposed PbRL algorithm represents the interpretable reward function as a binary tree structure where each internal node is a test on a state-action pair and each leaf returns a reward. An example is shown in Figure \ref{fig_ch1:reward_function_tree}.

%    \begin{figure}[]
%        \centering
%        \includegraphics[width=0.5\linewidth]{images/reward_tree.png}
%        \caption[A reward function in the form of a binary tree structure]{A reward function as a binary tree structure for the original RoboCar environment consisting of a 4-wheeled vehicle whose goal is to reach a certain area while avoiding obstacles \cite{DBLP:conf/atal/BewleyL22}. In this reward function, each internal node represents a test on a feature of an agent state, and each leaf a reward. The color of the leaves reflects the amount of reward the agent can obtain (blue means a high amount, red a low one).}
%        \label{fig_ch1:reward_function_tree}
%    \end{figure}
                    
The following works do not speak of explainability or interpretability but are mentioned because of the non-opaque form of the reward functions presented. 
                    % Finite-state machine (reward machine) (not XAI) Icarte
Icarte et al. (\citeyear{DBLP:conf/nips/IcarteWKVCM19}) propose to express the reward function in the form of a Finite State Machine as well as an algorithm, Q-learning for Reward Machines, to learn policies efficiently.
                    % From LTL to automata-based representations Camacho (not XAI)
Based on specifications given in a formal language such as Linear Temporal Logic (LTL), Camacho et al. (\citeyear{DBLP:conf/ijcai/CamachoIKVM19}) construct a reward machine to represent the different goals and temporal properties.
                    % Learn a Reward Machine for POMDPs (not XAI Icarte / Xu / Gaon)
For partially observable environments (i.e. the POMDP setting),  Icarte, Klassen, et al. (\citeyear{DBLP:conf/icml/IcarteKVM18}) learns with the help of experiences a reward machine and Z. Xu et al. (\citeyear{DBLP:conf/aips/0005GAMNT020}) jointly learns high level knowledge modelled by a reward machine and the agent's policy.
In Gaon and Brafman (\citeyear{DBLP:conf/aaai/GaonB20}), a model comprising non-Markovian Rewards is transformed into a Markovian model by learning a reward deterministic finite automaton from a set of sequences.
                    % Generate reward directly from high level specification (interpretable reward + task spec) Li
                    % RL with TL rewards Li (not XAI)
The robustness degree (X. Li et al. \citeyear{DBLP:conf/iros/LiVB17}) is used to extract a reward function from a logical formula expressed in Truncated LTL (TLTL). Such a formula corresponds to a specification of a task to be performed.
                    % TL reward context: apprenticeship learning Kasenberg
Based on a set of demonstrations, Kasenberg and Scheutz (\citeyear{DBLP:conf/cdc/KasenbergS17}) represent the task specification in the form of an LTL formula, instead of inferring a reward function that cannot be interpreted using IRL techniques.

                % Transition
Only the transition function is interpretable in the three following works.

                    % Use of expert knowledge graphical model (tree like) Probab. model Kaiser
Using expert knowledge about the environment, Kaiser et al. (\citeyear{DBLP:conf/esann/KaiserORE19}) describe a model-based RL method that learns an interpretable transition function.
In the illustrative problem, the expert knowledge gives the information that the state $s_{t+1}$ obtained from the state $s_t$ by performing the action $a_t$ depends on three different sources. 
The transition function is learnt using a Bayesian approach, taking into account a structure including the three sources. The resulting probabilistic model shows the sources impact on the transition for each agent's state-action pair. 

                    % Goel (learns a specific representation of the state(could be classified differently)) not really salientcy method / detect objects
Motion-Oriented REinforcement Learning (MOREL) (Goel et al. \citeyear{DBLP:conf/nips/GoelWP18}) improves the interpretability and, depending on the problem, the sample complexity of the RL algorithm used, by tracking moving objects in states. This approach is tested for a set of 59 Atari Games, where the state of an agent takes the form of an image.
First, an unsupervised method is applied to learn to predict object and camera motions as well as object masks using a random policy to generate the data. 
Then, the agent uses and refines the previously learned model to learn its policy. Users can view the object masks to understand the agent's policy.

Although not presented as XAI papers, we find it interesting to mention the following works. 
                    % Weighted labeled multi-Graph rpz Probab. model Metzen (not XAI)
Metzen (\citeyear{DBLP:conf/pkdd/Metzen13}) proposes an efficient heuristic called FIGE to represent the transition function by a graph.
                % Transition + Reward
                    % DT learned  Degris (model-based RL) (not XAI) Probab. model
The SPITI algorithm (Degris et al. \citeyear{DBLP:conf/icml/DegrisSW06}) learns a decision tree for the reward function and one for the transition function using the decision tree induction algorithm ITI (Utgoff et al. \citeyear{DBLP:journals/ml/UtgoffBC97}).

        % contrastive MDP transforms Finkelstein / Finkelstein
Finkelstein et al. (\citeyear{DBLP:conf/aaai/FinkelsteinLKPR22}) set out an approach based on a partial contrastive policy on a subset of states described by the user. The explainer looks for a meaningful transformation sequence to modify the current MDP into a contrastive MDP in which the agent would act according to the user's expected policy.
A set of such user-interpretable transformations is provided as input to the problem. An example of such a transformation is the single-outcome determinization transform (Yoon et al.  \citeyear{DBLP:conf/aips/YoonFG07}), which makes the transition function deterministic by considering only the most likely transitions.
To find a sequence of transformations, a Djikstra-like search is performed.
        
        % Lazy-MDP: agent takes control only on critical states Jacq ('sparse' policy)
Lazy-MDP's (Jacq et al. \citeyear{DBLP:journals/corr/abs-2203-08542}) are MDP's that include a default policy. The agent must then learn to estimate when it is necessary to replace the default policy and what action to take in replacement. Such an approach makes the policy sparse, making it possible to understand the states in which it is critical to take over from the default policy. In this work, the modelling problem and the policy generated make the approach interpretable.
        
        % Human-robot intercation: identify/correct human missing piece in it's reward function (based on inverse RL) Tabrez
In a human-robot cooperation setting, the framework proposed by Tabrez and Hayes (\citeyear{DBLP:conf/hri/TabrezH19}) consists of detecting differences between the human and robot task models and proposing an interpretable modification for the human model. 
To do this, the human's reward function is inferred based on its behavior during cooperation using Hidden Markov Models.  
A particular POMDP is presented, so that the agent can choose between an action that informs the human or an action related to the task to be performed. This communicative action tells the human about a missing piece of information in its reward function.

        \paragraph{Relational RL and MDP}
        \label{par:App_A_RRL_RMDP}
        % Merge relational MDP and RL?
        
        % Relational RL (Dzeroski): (non-exhaustive set of papers) (not directly targeted for explanations) (link with inherently interpretable policies)
        % Relational MDP: Guestrin (paper:'Generalizing plans to new environments in relational MDPs')
The following works provide an interpretable representation of states and transitions and are based on works presenting relational MDPs (Guestrin et al. \citeyear{DBLP:conf/ijcai/GuestrinKGK03}) and Relational RL (Dzeroski et al. \citeyear{DBLP:journals/ml/DzeroskiRD01}).
As a reminder, the objective is to be able to learn a generalizable and efficient policy based on a representation of the environment by a set of objects and relations between them. In this section, we present only an overview of work in these domains.
        % Survey: van Otterlo (logic-based and relational MDP)
For more details, we recommend reading the state of the art Otterlo et al. (\citeyear{DBLP:books/sp/12/Otterlo12}) dating from 2012, which does not therefore include recent work.
       
The works presented in this paragraph are described briefly because they are not proposed as explainability methods, although the approaches do make the MDP interpretable. 
        % Object-Oriented MDP Det. Model Diuk (not XAI)
Diuk et al. (\citeyear{DBLP:conf/icml/DiukCL08}) introduce the Oriented-Object MDP (OO-MDP) to model the problem based on objects and their interaction and describe an algorithm for solving deterministic OO-MDPs.
        % Deictic OO-MDP Det. Model Marom (not XAI)
This algorithm is modified to solve Deictic OO-MDPs (Marom and Rosman \citeyear{DBLP:conf/nips/MaromR18}), which represent relationships between an object and an object class (instead of a simple object-object relationship), enabling better abstraction.
        % Physic based model OO-MDP Det. model  Scholz (not XAI)
Based on this idea, Scholz et al. (\citeyear{DBLP:conf/icml/ScholzLIW14}) focus on physical domains, and propose two approaches: an extension of OO-MDP and a more efficient approach using a physics engine.
        % Object-centric perception, prediction and planning NN model Veerapaneni
In a model-based RL context, Veerapaneni et al. (\citeyear{DBLP:conf/corl/VeerapaneniC0JF19}) approximate the observation model which represents an image (i.e. agent state) by a set of semantic masks representing objects and the dynamics model which describes the temporal evolution of the different objects for planning.
        % Schema networks learned Kansky (not firts purpose XAI) Proba. model
Schema networks (Kansky et al. \citeyear{DBLP:conf/icml/KanskySMELLDSPG17}) is a generative model of an MDP that learns the dynamics of the environment in a setting where the state is represented by a set of objects in an image.
        % Relational world model as a relational MDP Walker  (not XAI)
An interpretable MDP is built in Walker et al. (\citeyear{DBLP:conf/ilp/WalkerTSM07}) where abstract states are learned via ILP techniques and transitions and rewards are estimated. The policy is simply learned by value iteration in this MDP.
        % Relational action schema Walsh (not XAI)
The relational actions schemas class (Walsh \citeyear{walsh2010efficient}) groups together a set of languages that compactly represent the transition function using relational conditions and effects.

        % Martinez
An RRL algorithm is proposed in Martínez et al. (\citeyear{DBLP:journals/ai/MartinezAT17}) in the context of robotic tasks where the agent can request demonstrations from the expert in order to learn efficiently. This allows the agent to learn in a sample-efficient way and to request a demonstration when an action is unknown to it.
In this work, each action is defined by a set of preconditions and effects weighted by probabilities. 
To guide the expert in generating demonstrations, the excuse principle (Göbelbecker et al.  \citeyear{DBLP:conf/aips/GobelbeckerKEBN10}) is used to explain why the plan failed. An excuse returns the important predicates that caused the plan to fail.

        % Object-oriented predictor Zhu (not first goal XAI) NN Model (one object)
The framework described in G. Zhu, Z. Huang, et al. (\citeyear{DBLP:conf/nips/ZhuHZ18}) is used to extract objects from an image and learn the transition function. The objective of this framework is to generate the next image (i.e. state) using the information extracted from the image (its background and the various objects) and the action of the agent.
To do this, the background extractor learns to extract the background, the object detector learns to produce masks of different types of static or dynamic objects and the dynamics net learns to make the transition. 
The object masks show that the method correctly learns to distinguish objects in the game tested, even on different levels of the game (i.e. unseen environments).
        % Object-oriented predictor Zhu NN Model (multiple objects)
An extension of this work, G. Zhu, J. Wang, et al. (\citeyear{DBLP:conf/aaai/ZhuWRLZ20}) proposes a three-level learning architecture for learning model dynamics from the most to the least abstract. To aid learning, the output of an abstract level is provided to the level below.
In order from most to least abstract, motion detection detects regions containing dynamic objects from an image sequence, instance segmentation produces coarse masks of dynamic objects and dynamics learning learns the transition function based on agent action and object relationships (as in G. Zhu, Z. Huang, et al.).
        
        % Object-oriented representation Agnew NN model
The Object-Level RL (Agnew and Domingos \citeyear{agnew2018unsupervised}) is a framework that produces a high performance policy by learning a representation of states using relative and absolute distances, velocities, acceleration and contacts between the agent and objects in an image. This method was tested on a set of Atari games.
During sample-efficient learning, the agent's exploration is based on a prior knowledge, which states that interesting experiences occur when two objects come into contact.

        % Relational DRL Zambaldi (from one observation, build underlying graph leading to the goal)
Zambaldi et al. (\citeyear{DBLP:journals/corr/abs-1806-01830}) describe an architecture where the agent learns to extract objects from the image, determine relations between them and produce the policy and value for a state.
For extraction, a CNN is used and a self attention mechanism (Vaswani et al.  \citeyear{DBLP:conf/nips/VaswaniSPUJGKP17}) is used to compute relations between objects. These representations are then given as input to an RL algorithm for learning the policy and value function.
The visualization of attention weights on different objects allows to analyze the learned relations between objects and to understand the decision making of the agent.

        % state Representation (from state into relational representation to inform Q/policy network)
        % relational architecture with relational units (Q-learning) Adjodah
Symbolic Relation Network (Adjodah et al. \citeyear{adjodah2018symbolic}) takes as input a set of objects which describes the agent's state, represents them as a unary or binary relation and then concatenates them into a relational state used to learn Q-values with a DQN.
In the environment used, the learned relations are interpretable: they represent the type of object and the relative location of the objects with respect to the agent. 
In this work, the object representation of states is not learned.% an state is a set of objects are assumed to be already extracted from an upstream agent state.
    
% From Surrogate model Policy
% pixels -> symbolic features Sieusahai
% Cascading DT Ding
    %------------------------------------------------
    \subsubsection{Surrogate Model}
    \label{sec:App_A_surrogate_model}
    %------------------------------------------------

            \paragraph{States Clustering.}
            \label{par:App_A_States_Clustering}

                % Hyperstates + Policy graph Bekkemoen et Langseth
Bekkemoen and Langseth (\citeyear{DBLP:conf/acml/BekkemoenL23})  propose the ASAP method, which groups states into clusters representing abstract states. 
The explanation consists of two parts: a display of a representative state and its associated action for each abstract state identified and a modelling of a policy graph, which is a Markov chain showing the agent's behavior in the abstract state space. 
Abstract states are identified using attention maps (which can be likened to saliency maps).    
ASAP is illustrated with the Mountain Car environment (Brockman et al.  \citeyear{brockman2016openai}) in Figure \ref{fig_ch1:ASAP}.

%    \begin{figure*}[h]
%        \centering
%        \includegraphics[width=1.0\textwidth]{images/ASAP_hyperstates.png}
%        \caption[Explanation via ASAP of the agent's policy for the MountainCar environment]{Explanation via ASAP \cite{DBLP:conf/acml/BekkemoenL23} of the agent's policy for the Mountain Car environment. The first line shows a representative state for each hyperstate with its associated action. The second line is divided into two parts. On the left, a policy graph models the policy based on the hyperstates. Each vertex is a hyperstate and each edge represents a transition between two hyperstates (weighted by its probability of occurrence). On the right, the attention maps of each hyperstate are displayed (where an attention score is associated with each feature, determining the impact of the features on the agent's decision-making).}
%        \label{fig_ch1:ASAP}
%    \end{figure*}

In the next two paragraphs, the explanation describes a specific cluster of states in natural language that answers the user's question.

                % minimized DNF state regions description Hayes + use of technique Climent
Given a set of communicable predicates for describing states, Hayes and Shah (\citeyear{DBLP:conf/hri/HayesS17}) describe a framework for answering different questions about policy, such as \emph{`When do you do action $a$?'}. To do this, the user's question is first identified, then a region of states is extracted based on the question. 
Using the optimised Quine-McCluskey algorithm (Jain et al. \citeyear{DBLP:conf/icas/JainKM08}), this region represented by a logical formula is minimised. Since the predicates are communicable, the resulting compact formula is transposed into a human language description for the user.
                % Sort of clustering: knowledge compilation of predicates (When do you action?) (not RL) Booth (based on hayes and shah) generate control logic summaries (not RL)
                % similar work of Hayes Iucci (use of same method?!) paper: 'Explainable reinforcement learning for human-robot collaboration'
A complementary study of this approach is carried out in Climent et al. (\citeyear{DBLP:conf/ccia/ClimentGA21}) on the CartPole problem. This approach is combined with Reward Decomposition (Erwig et al. \citeyear{erwig2018explaining})  to provide both local and global explanations of agent behavior in Iucci et al. (\citeyear{DBLP:conf/icar/IucciHTIL21}). Based on this method, Booth et al. (\citeyear{DBLP:conf/ijcai/BoothMS19}) carried out a user study to analyse the interpretability of different propositional theories.
                
                % (d-DNNF form) knowledge compilation Wollenstein-Betech
Using knowledge compilation techniques, the framework of Wollenstein-Betech et al. (\citeyear{DBLP:conf/itsc/Wollenstein-Betech20a}) answers a set of questions relating to the choice of a given agent action. For example, this method can provide the probability of doing an action knowing that we are in a state where one of its features is true.
To do this, given a set of state-action tuples and a question, a deterministic Decomposable Negation Normal Form (d-DNNF) is constructed, model counting and probabilistic inference are performed on the d-DNNF and the result is translated into natural language.

                % Summarize training histories with tree representation of abstract states / abstract Markov models Bewley (not really critical states) salient points?
By collecting all the transitions during agent training, Bewley et al. (\citeyear{DBLP:journals/corr/abs-2201-07749}) propose a state abstraction as well as a temporal abstraction to analyse agent learning.
The state space is decomposed into regions, called abstract states. The characterisation of these states is provided in the form of a tree diagram. The learning phase is divided into several windows, where each window is represented by a transition graph using the abstract states. This reflects the agent's interactions with the environment over a learning period. 
In addition, two graphs are provided to represent the rate of visits to abstract states as the training progresses, as well as the abstract states in which an agent's episode ends. 

                % Interpretable rpz of state space Bewley and Lawry
TRIPLETREE (Bewley and Lawry \citeyear{DBLP:conf/aaai/BewleyL21}) is an algorithm that constructs a decision tree where the branches represent the action, the value and the state derivative estimate. This method allows the state space to be discretised according to three criteria: agent's action, value function and temporal dynamics. In addition, the transition probabilities between the different regions obtained are computed.
A set of visualisation tools is used to analyse the environment and the agent. In addition, with these different regions, a set of explanations can be provided on the different criteria: factual explanations describing the limits of the region or counterfactual explanations based on a principle of minimal change to reach a certain region.

            \paragraph{State Transformation.}% Peut etre à dispatché aussi
            \label{par:App_A_States_Transformation}
            
            % State representation learning (learn representations of more interpretable states)
            % SRL survey Lesort
The field of State Representation Learning (SRL) consists of learning a low-dimensional representation of the state, which evolves over time and reflects the change induced by the agent's actions. Although the main objective of SRL is the performance of the learned policy, proposing a state representation composed of a small number of comprehensible features makes it possible to provide interpretability, as stated in the survey Lesort et al.  (\citeyear{DBLP:journals/nn/LesortRGF18}).
The objective of SRL is to learn a mapping function which, for a history of observations, returns the state in a low-dimensional representation. To do this, supervised approaches use the true high-dimensional state, while unsupervised approaches (the ones described in the survey) do not. 
            % SRL Toolbox Raffin
Raffin et al. (\citeyear{DBLP:journals/corr/abs-1809-09369}) propose a toolbox for SRL methods to compare approaches in different environments, using visualisation and metrics.

            % Mix Symbolic DRL Garnelo from raw perceptual data to low dimensional symbolic representation
In Garnelo et al. (\citeyear{DBLP:journals/corr/GarneloAS16}), the agent module decomposes into two sequential parts: a neural network that learns to compress a state into a symbolic form, and a Q-learning algorithm that learns to reason based on the symbolic representation.
In the tested environment, the neural network represents an image symbolically by a set of objects, their characteristics and interactions. This approach outperforms the compared DQN, and allows the agent's behavior to be interpreted through symbolic states.
            
            % Type of sub-states chosen d'Avila Garcez
D'Avila Garcez et al. (\citeyear{DBLP:journals/corr/abs-1804-08597}) present a simplified version of the above approach as well as an extension. The symbolic representation of a state is composed of sub-states reflecting the relative position of the agent with respect to an object.
Learning and action choice are modified to take account of this representation in the proposed extension. This addition of common sense results in a high-performance policy.
            
            % Conceptual embbedings for DRL agent Dai constrain representation space to interpret DNN (hierarchical conceptual embedding method + prior knowledge) (not really state representation)
By combining a hierarchical learning architecture with conceptual embedding techniques to embed prior knowledge, the agent learns a more interpretable representation of the state (Dai, Ouyang, et al. \citeyear{DBLP:journals/apin/DaiOZLD23}).
A saliency map is generated on the hidden layers including the embedding of prior knowledge, to determine the impact of this knowledge on the agent's decision-making. 
              
            % Encoding abstract state (with some actionable features) François-Lavet
François-Lavet et al. (\citeyear{DBLP:conf/aaai/Francois-LavetB19}) propose to combine the model-based and model-free approaches of RL algorithms to learn an abstract state representation.
A specific loss is added for learning the representation of an abstract state so that features of the abstract state are impacted by the agent's actions. This makes the representation interpretable. 
The method is applied in a deterministic environment, but can be adapted for stochastic environments.
            
            % Encoding binary attributes + build graph Zhang 
The HRL method described in A. Zhang et al. (\citeyear{DBLP:conf/icml/ZhangSLSF18}) proposes to learn a representation of states in the form of a set of attributes, as well as a transition function between these attributes. In this work, the binary attributes are learned in a supervised manner and correspond to the relative positions of objects in a 3D environment.

                    % Env model (bird-view semantic mask) Chen
 In J. Chen et al. (\citeyear{DBLP:journals/tits/ChenLT22}), a latent model of the environment is learned at the same time as the agent's policy in order to reduce the sample learning complexity and generate semantic bird-eye masks.
 For an urban autonomous driving problem, the input consists of a front view image and a lidar image. 
 The generated masks (learned from ground truth) are based on the input to display the map, the route that will be taken, the surrounding objects and the car position. 
            
            % Correct critical states spurious correlation (debug policy) Gajcin (pas sûr de garder)
ReCCoVER (Gajcin and Dusparic \citeyear{DBLP:conf/atal/GajcinD22}) detects causal confusion in the agent decision-making on a set of critical states. From each critical state, the policy is tested on a set of alternative environments. These environments consist of imposing a certain value on a feature of the state.
To detect causal confusion, it is necessary to learn a set of feature-parameterised policies for each subset of features. Causal confusion occurs when the agent's policy does not perform well in an alternative environment compared with a feature-parameterised policy. 
ReCCoVER then proposes a correction for these problematic states by providing the user (in this context, a developer/expert) with the feature subset on which the agent should base itself when it encounters this state.

%------------------------------------------------
%------------------------------------------------
\subsection{Visual Analysis}
\label{sec:App_A_visual_analysis}
%------------------------------------------------
%------------------------------------------------

    %------------------------------------------------
    \subsubsection{Visual Toolkit}
    \label{sec:App_A_visual_toolkit}
    %------------------------------------------------

                % Policy Explainer Mishra
Mishra et al. (\citeyear{DBLP:journals/corr/abs-2104-02818}) propose a total of 8 tools to be used to understand the agent's policy. This interface, which is described in Figure \ref{fig_ch1:policyExplainer}, requires the features describing a state to be interpretable.
PolicyExplainer provides a global view of the frequency of actions and rewards, the policy, the value function  (cf. panels A-C  in Figure \ref{fig_ch1:policyExplainer}) and can be used to look at particular states and their associated Q-value, or a particular trajectory (cf. panels D-G in Figure \ref{fig_ch1:policyExplainer}). In addition, the \emph{`Why?'}, \emph{`Why not?'} and \emph{`When?'} questions are answered by providing the state features that have an impact on the choice of action (cf. panel H in Figure \ref{fig_ch1:policyExplainer}). 

%    \begin{figure*}[h]
%        \centering
%        \includegraphics[width=1.0\linewidth]{images/PolicyExplainer.png}
%        \caption[PolicyExplainer visual interface]{PolicyExplainer visual interface \cite{DBLP:journals/corr/abs-2104-02818}. The different tools are divided into three parts. Panels A-C provide a global view of the frequency of actions and rewards, the policy, the value function. Panels D-G provide a detailed analysis of particular states and their associated Q-value, or a particular trajectory. Panel H answers the questions ‘Why?’, ‘Why not?’ and ‘When?’ using state features.}
%        \label{fig_ch1:policyExplainer}
%    \end{figure*}

The other visualisation interfaces are designed for experts, to help them with debugging. 
                % DRLViz (expert users only) Jaunet
DRLViz (Jaunet et al. \citeyear{DBLP:journals/cgf/JaunetVW20}) analyses the latent internal memory of a DRL agent in the context of a video-based environment.
                % DQNViz (expert users only) Wang
DQNViz (J. Wang et al. \citeyear{DBLP:journals/tvcg/WangGSY19}) focuses on the interpretation of DQN training and illustrates it using the Atari Breakout game.
                % DynamicsExplorer
DynamicsExplorer (W. He et al. \citeyear{DBLP:conf/apvis/HeLBWS20}) focuses on a DRL agent that includes a Long-Short-Time-Memory (Hochreiter and Schmidhuber \citeyear{DBLP:journals/neco/HochreiterS97}) layer for encoding environment dynamics.
                % MDPvis
MDPvis (McGregor et al. \citeyear{DBLP:journals/vlc/McGregorBDHMM17}) is used to analyse and modify an MDP and optimise the policy.

    %------------------------------------------------
    \subsubsection{Latent Space Visualization}
    \label{sec:App_A_interpreting_DQNs}
    %------------------------------------------------

                % Understand DQN Zahavy
Zahavy et al. (\citeyear{DBLP:conf/icml/ZahavyBM16}) propose to visualise the activations of the DQN by applying t-Distributed Stochastic Neighbor Embedding (t-SNE) (Van der Maaten and G. Hinton \citeyear{van2008visualizing}) to reduce the dimension.
Before visualisation, a set of data is collected, including hand-crafted features such as the agent's position, so that clusters can be identified and analysed. In addition, saliency maps are also generated on the different states collected. By analysing the dynamics between clusters, the authors identified that the DQN has learned a hierarchical aggregation of the state space.

                % Visualizing dynamics Zrihem
Following on from this, Zrihem et al. (\citeyear{DBLP:journals/corr/ZrihemZM16}) present a method for modelling the t-SNE representation of the DQN by a Semi Aggregated MDP, a human-interpretable approximation of the MDP.
Also, clustering algorithms are used to identify the structure of the t-SNE maps, instead of having to construct features by hand. These clusters are states of the Semi-MDP and the probability of moving from one cluster to another defines the transition matrix of the Semi-MDP. An example of Semi-MDP obtained based on the policy of an agent that has learned to play Atari 2600 Breakout (Bellemare et al. \citeyear{DBLP:journals/jair/BellemareNVB13}) is shown in Figure \ref{fig_ch1:visual_analysis}.

%    \begin{figure*}[h]
%        \centering
%        \includegraphics[width=1.0\textwidth]{images/Visualize_DQN.png}
%        \caption[Semi-MDP for the Atari 2600 Breakout environment]{Semi-MDP built on top of a t-SNE map for the Atari 2600 Breakout environment \cite{DBLP:journals/corr/ZrihemZM16}. The t-SNE map is the set of points on which the Semi-MDP is based. It is made up of states (or clusters of MDP states), linked by transitions weighted by their probability of occurrence. Each cluster is modelled by an image of the game where the position of the ball in each state of the cluster is displayed in red.}
%        \label{fig_ch1:visual_analysis}
%    \end{figure*}

                % Annasamy
A specific architecture is presented by Annasamy et al. (\citeyear{DBLP:conf/aaai/AnnasamyS19}) with the aim of interpreting the DQN when the agent's states are images. The agent learns to focus on keys of a store composed of an action and a Q-value.
These keys can be assimilated to cluster centres. Using deconvolutions, these cluster centres $(a,q)$ are then visualised in order to understand the important elements of the reconstructed image that lead to the choice of $a$ and a Q-value $q$. 
This image represents an aggregate of states. For example, in Mrs Pacman, the visualisation of an action-return pair displays the different positions of the yellow blob in which the agent chooses the pair $(a,q)$.

    %------------------------------------------------
    \subsubsection{Inspection}
    \label{sec:App_A_inspection}
    %------------------------------------------------

                % Evolution / Relationship between metric(s) ETEMOX Parra-Ullauri
ETeMoX (Ullauri et al. \citeyear{DBLP:journals/sosym/UllauriGBZZBOY22}) is a framework for keeping track of the evolution of certain metrics as the agent is trained. This framework consists of three parts.
The Translator, which transforms the data collected from the agent's interaction with the environment. 
The Complex Event Processing (Luckham and Frasca \citeyear{luckham1998complex}), which filters the data to select only the relevant ones.
The Temporal Graph Database, which represents relationships between multiple evolving metrics with time dimension. 
In this work, three RL algorithms were compared using ETeMoX by monitoring the overall reward, the exploration/exploitation trade-off and a feature specific to the environment used.
            % Display several information through plots + LIME Dethise
Dethise et al. (\citeyear{DBLP:conf/sigcomm/DethiseCK19}) study the \emph{Pensieve} agent by looking at, for example, the probability of action choice or confidence in the choice of actions. In addition, a study of the global importance of the agent's state features is carried out using LIME  (Ribeiro et al. \citeyear{DBLP:conf/kdd/Ribeiro0G16}).

            % LIME
This method estimates the influence that each feature of input $x$ has on the model's output. In our context, the input is a state of the agent, the model the agent's policy and the output an action. 
To calculate the influence of features, a transparent surrogate model is learned locally around $x$. This surrogate model is a linear model. To train it, a set of points is generated by perturbing $x$ and weighted as a function of distance from $x$. 
After training, the weights of the linear model are used to explain the influence of each feature of $x$ on the output of the model.
In Dethise et al., LIME is used on a set of states to provide the average contribution of each feature to the agent's choice of actions.

\section{Summaries of Sequence-level Methods}
\label{sec:App_B}

% Introduction
This appendix provides a summary of each sequence-level method mentioned in the paper. These methods explain sequences of interaction between the agent and the environment. 
As a reminder, a total of three ways of explaining sequences have been identified.
\emph{Counterfactual Sequences} allow a comparison of the agent's behavior with a sequence that uses an alternative policy, and thus determine the agent's strengths and weaknesses. 
The \emph{Important Elements} of a sequence are those that have the greatest impact on the agent's ability to achieve a given objective. 
The use of \emph{Human-readable MDP} makes it possible to understand, in the context of a sequence, the information available to the agent, the actions and the dynamics of the environment. 

%------------------------------------------------
%------------------------------------------------
\subsection{Counterfactual Sequence}
\label{sec:App_B_SOTA_counterfactual_sequence}
%------------------------------------------------
%------------------------------------------------

        % Contrastive X expected consequences Van der Waa
Van der Waa et al. (\citeyear{DBLP:journals/corr/abs-1807-08706}) propose to create a counterfactual sequence based on the user's query. An alternative policy is obtained by modifying the agent's Q-value so that the policy locally follows the user's contrastive query.
The sequences are generated using the most probable transition at each time step (note that the transition function was obtained by training). 
The explanation consists of describing the sequences with the occurrence of the actions performed and states encountered, as well as the positive and negative outcomes.

        % Experiential X Alabdulkarim (state and sequence)
In a context where the environment includes different reward classes, Alabdulkarim and Riedl (\citeyear{DBLP:journals/corr/abs-2210-04723}) train a set of influence predictor models for each reward type during agent training. 
The counterfactual sequence is generated by considering the action (or sequence of actions) proposed by the user, then using the agent's policy. 
The sequences are compared using influence predictors (i.e. a calculation of the average influence of each type of reward in each sequence), the cost of the actions and the final reward obtained.
An example showing the interest of this approach is presented in Figure \ref{fig_ch1:experential}.

%    \begin{figure}[]
%        \centering
%        \includegraphics[width=0.6\linewidth]{images/Experiential_Xp.png}
%        \caption[Example of explanation using influence predictors.]{Possible explanations for the sequence of actions that the agent (represented by a robot in cell (3,1)) performs in a 2D environment to reach the green cell \cite{DBLP:journals/corr/abs-2210-04723}. The user's question is written in red. Explanation A (in black and marked by a red cross) is based solely on the agent's policy, while explanation B (in blue marked with a correct tick) uses influence predictors. According to authors, explanation B is of better quality.}
%        \label{fig_ch1:experential}
%    \end{figure}

        % Counterfactual sequences ACTER Gajcin
The evolutionary algorithm, named ACTER (Gajcin and Dusparic  \citeyear{DBLP:journals/corr/abs-2402-06503}), generates a diverse set of counterfactual sequences to propose to the user sequences that could have prevented the agent from reaching a state of failure.
The algorithm performs multi-criteria optimisation, in accordance with the 5 properties defined upstream. 
Among these, validity ensures that the counterfactual avoids failure and proximity ensures that the sequence of actions resulting from the agent and the counterfactual one are as similar as possible.

%    \begin{figure*}[]
%        \centering
%        \includegraphics[width=1.0\textwidth]{images/EDGE.png}
%        \caption[Explanation via EDGE of an episode from a game of Atari 2600 Pong.]{An explanation of an episode from a game of Atari 2600 Pong \cite{DBLP:conf/nips/GuoWKX21}. The objective is to determine the important time-steps for the defeat of the agent controlling the green pad.
%        The colored bar below each state of the episode represents its level of importance (yellow represents a low level, orange a medium level and red a high level).}
%        \label{fig_ch1:EDGE}
%    \end{figure*}
    
        % Counterfactual sequence (MDP) Tsirtsis (pas pour le RL) (use of causal model)
The objective of Tsirtsis et al. (\citeyear{DBLP:conf/nips/TsirtsisDR21}) is to provide a counterfactual sequence that leads to a better outcome under the constraint that at most k actions are modified with respect to the agent's sequence. This work is carried out in a MDP context where the transition function is modelled by the Gumbel-Max structural model (Oberst and Sontag  \citeyear{DBLP:conf/icml/OberstS19}). The algorithm proposes a polynomial-time solution based on dynamic programming.
In the case where the state space is continuous, this problem is NP-hard (Tsirtsis and Rodriguez \citeyear{DBLP:conf/nips/TsirtsisR23}). Tsirtsis and  Rodriguez then proposed a search method based on the A* algorithm.

        % Similar to Erwig work but with hierarchical types of risks ('what', 'why', 'where'?) Stefik
For a specific drone parcel delivery environment, Stefik et al (\citeyear{stefik2021explaining}) use MSX (Erwig et al.  \citeyear{erwig2018explaining}) (which is described later) to explain in natural language an action sequence in terms of risk factors.
The risk is decomposed into different factors, so as to be able to understand what types of risk the agent is trying to avoid by performing a certain action. 
In this paper, the authors answer two questions related to the agent's action sequence: \emph{`Why?'} by comparing two sequences on the different risk factors and \emph{`What?'} by identifying in a sequence the action that leads to the greatest difference in risk.

%------------------------------------------------
%------------------------------------------------
\subsection{Important Elements}
\label{sec:App_B_SOTA_important_elements}
%------------------------------------------------
%------------------------------------------------

        % EDGE Guo
The EDGE algorithm is a self-explainable model that predicts, for a given episode, the final agent’s reward (W. Guo, X. Wu, et al. \citeyear{DBLP:conf/nips/GuoWKX21}). The provided explanation takes the form of a set of important time-steps within the agent’s interaction episode with the environment.
An explanation of a game of Atari 2600 Pong (Bellemare et al.  \citeyear{DBLP:journals/jair/BellemareNVB13}) is shown in Figure \ref{fig_ch1:EDGE}.

        % video-based env / extract critical states of trajectory for predicting return Liu
In a video-based environment, Liu et al. (\citeyear{DBLP:conf/iccv/LiuZLWFGS23}) identify the critical states of a sequence sufficient to predict the final reward obtained by the agent. Two models are learned for this method: the return predictor, which predicts the agent's final reward based on a (partial) video (i.e. a sequence of frames), and the critical state detector, which identifies critical frames by learning to mask non-critical ones.
        
        % HXP + B-HXP
The History eXplanation based on Predicates (HXP) method (Saulières et al.  \citeyear{DBLP:conf/pkdd/SaulieresCS23})  and its variant, named Backward-HXP (Saulières et al. \citeyear{DBLP:conf/ecai/SaulieresCS24}) are part of this line of work.
The aim of these methods is to provide, within a sequence, the most important actions for the realisation of a predicate. 
This predicate is respected in the final state of the sequence studied, and may represent the success or failure of an agent, or other characteristics specific to the environment studied.

%------------------------------------------------
%------------------------------------------------
\subsection{Human-readable MDP}
\label{sec:App_B_SOTA_human-readable_MDP}
%------------------------------------------------
%------------------------------------------------

    %------------------------------------------------
    \subsubsection{Surrogate Model}
    %------------------------------------------------

            % Contrastive exp. learn symbolic model (based on pre-defined user vocab) describing actions using preconditions and actions cost based on a state Soni / Sreedharan (same work) ((action and sequences) (Soni_Utkshar_thesis.pdf)
Relying on a user-defined vocabulary, Sreedharan et al. (\citeyear{DBLP:conf/iclr/SreedharanSVSK22}) present the idea of learning propositional concepts to explain the agent's action choices in a contrastive manner.
With these propositional concepts, actions are represented in a symbolic way by preconditions, effects and their cost. 
The user proposes an alternative sequence of actions to the agent. 
The explanation shows that the user's sequence does not lead the agent to achieve the objective, or leads it into an invalid state, or describes that the proposed sequence is more costly than the agent's sequence.

            % Explain a set of transitions via explanatory messages. Goal is to find the user-type of the user to provide personalized explanations Soni (Soni_Utkshar_thesis.pdf)
Soni (\citeyear{soni2024towards}) proposes to explain a set of transitions (or sequence) using a sequence explanatory message.
Firstly, by collecting a set of data by interacting with various users, the method groups users into user-types and learns their associated labelling function. These functions allow to determine whether an explanation works for a user-type, and is therefore useful for providing personalised explanatory messages. 
Once this has been done, explanations can be generated for an unknown user. However, the user's type is not given. To provide personalised explanation, the problem consists in solving a POMDP where the hidden part of the state corresponds to the user's type.
    
    %------------------------------------------------
    \subsubsection{Inherently Understandable}
    %------------------------------------------------

            % Actions interpretable by design (image reconstruction context) Li
In a context of sequential reconstruction of magnetic resonance images, Li et al. (\citeyear{DBLP:conf/aaai/LiFANZ20}) use RL to successively apply interpretable pixel-wise operations.
In this environment, the action space is directly interpretable because each action corresponds to common Computer Vision filters, such as the Sobel filter or the Gaussian filter.

\section{Summaries of Action-level Methods}
\label{sec:App_C}

This appendix provides a summary of each action-level method mentioned in the paper. 
These methods explain the agent's choice of action. 
As a reminder, the various methods have been grouped into two types of explanation: \emph{Feature Importance}, which determines the features of the agent's state that led it to choose an action, and \emph{Expected Outcomes}, which justifies the agent's decision by providing its future potential impact. 

%------------------------------------------------
%------------------------------------------------
\subsection{Feature Importance}
%------------------------------------------------
%------------------------------------------------

    %------------------------------------------------
    \subsubsection{Saliency Maps}
    \label{sec:App_C_saliency_maps}
    %------------------------------------------------

                \paragraph{Attention Mechanism}
                \label{par:App_C_attention_mechanism}

                % Attention: (some works learns a specific representation of the state (could be classified differently))
                    % Mott
In Mott et al. (\citeyear{DBLP:conf/nips/MottZCWR19}), the current state of a Long Short Term Memory is used to make a set of requests, sent to an MLP. Its output is used, together with the agent state, to generate saliency maps. This system is called \emph{attention head}.
Experiments have shown that the agent has learned to focus its attention on regions/objects present in the image.

                    % Itaya
Based on A3C (Mnih, Badia, et al. \citeyear{DBLP:conf/icml/MnihBMGLHSK16}), Itaya et al. (\citeyear{DBLP:conf/ijcnn/ItayaHYFS21}) propose two attention masks: one for the policy and one for the state value. These attention masks are generated from the feature map extracted by the first part of the neural network, called the feature extractor, and then used to predict the agent's action and estimate the value of the agent's state.

                    % Nikulin
Two feature extractors are presented in Nikulin et al. (\citeyear{DBLP:conf/iccvw/NikulinIAN19}): Sparse FLS and Dense FLS (where FLS stands for Free Lunch Saliency). The main objective of this work is to provide an attention mechanism that makes both visualisations interoperable, without impacting the agent's performance.
Both approaches and baselines were tested using Atari-head (R. Zhang, Walshe, et al. \citeyear{DBLP:conf/aaai/ZhangWLGMWZHB20}), a dataset containing human gameplay and visual tracking. None of the approaches studied stood out from the others.

                    % Action region scoring mechanism (similar to attention?!) Liu
The Action Region Scoring (ARS) module proposed in Y. Liu et al.  (\citeyear{DBLP:conf/ksem/Liu0CJ22}) is used both to explain the choice of action and to improve learning.
In the CNN, at the output of each convolutional layer to which a ReLU activation has been applied, an ARS module is used to identify the important regions and then combined with the current image representation. This helps learning by indicating the regions to focus on. 
For the explanation part, the saliency map is produced by retrieving the outputs of the ARS modules and combining them.
    
                    % Ou
In an approach based on fuzzy rule learning, Ou et al.   (\citeyear{DBLP:journals/tfs/OuCWL24}) propose to use the Compact Convolutional Transformer (Hassani et al. \citeyear{DBLP:journals/corr/abs-2104-05704}) to provide attention masks. In addition to this visualisation, the choice of agent is explained by providing the influence of each fuzzy rule.

                    % Shi 
Based on the U-net architecture (Ronneberger et al. \citeyear{DBLP:conf/miccai/RonnebergerFB15}), SSINet takes the form of an encoder-decoder to obtain an attention mask of the input image (Shi et al. \citeyear{DBLP:journals/pami/ShiHSWLW22}).
SSINet is built in such a way as to respect 2 properties: maximum behavior resemblance and minimum region retaining. 
The first describes that the agent's prediction must be consistent between the image and the image overlapped with the attention mask and the second describes that the attention mask must focus on as little information as possible, so as to provide sparse explanations.

                    % Tang 
Tang et al. (\citeyear{DBLP:conf/gecco/TangNH20}) only provide the agent with a subset of patches from the original image for its decision making. Indeed, the image is first cut into a set of patches, which are then evaluated using a self-attention module.
Finally, the $k$ most important patches are extracted from this evaluation and given %according to attention are given 
to the agent. This technique makes it possible to understand on which patches of the original image the agent bases its decision.

                    % Josef 
In Josef and Degani (\citeyear{DBLP:journals/ral/JosefD20}), different types of input are tested for an unmanned ground vehicle navigation problem. Several attention mechanisms are implemented depending on the features (visual or not).
Thus, the self-attention mechanism used to explain the problem is applied to an image or a vector of features.          

                    % Bao (top-down and bottom up saliency maps (input: video frames)) / 
The DRIVE model (Bao et al. \citeyear{DBLP:conf/iccv/Bao0K21}) is proposed for a traffic accident anticipation problem using videos.
Two attention maps are generated for the same frame $f_t$: a bottom-up approach which generates a map based directly on $f_t$ and a top-down approach which applies a transformation to $f_t$ before generating the map. 
This second approach focuses the attention mechanism on a risky region of $f_t$, using a foveal vision module based on a fixation point p predicted by the agent with the previous frame $f_{t-1}$.

                    % Yang (3 types of visualization) (learns a specific representation of the state (could be classified differently))
Region-sensitive Rainbow (Z. Yang et al. \citeyear{DBLP:journals/corr/abs-1812-11276}) is an enhancement to Rainbow DQN (Hessel et al.  \citeyear{DBLP:conf/aaai/HesselMHSODHPAS18}) which uses a region-sensitive-module to determine the important regions in which the agent focuses to choose an action.
This module is similar to an attention mechanism, and returns importance scores for $k$ regions of an image. 
Of the three ways of visualising salient regions, the most user-friendly approach has been retained. It simply consists of a binary saliency map which displays just the salient region to the user (replacing the rest of the image with black pixels).

                    % Reward oriented attention (Yang)
Focusing on the immediate reward, Yang et al. (\citeyear{DBLP:journals/tsmc/YangWTSHS24}) propose to generate attention masks that reflect as much as possible the agent's reward.
The reward obtained by using the policy from state $s$ is compared with the reward obtained by using it from state $s'$ constructed by combining $s$ and the attention mask. To minimise the absolute difference of these rewards, an RL approach is proposed. An extension of this method is provided to handle multi-step rewards.

                    % Image -> attention -> verbal explanation Wang
The approach proposed by Wang et al. (\citeyear{DBLP:conf/ro-man/WangYZLS19}) is different from the rest of the work. In addition to an attention map, an explanation in the form of natural language is produced.
The architecture of the algorithm presented breaks down into three parts: an encoder, which encodes the image into a set of features, an attention mechanism, which returns the salient features and the decoder, which generates a verbal explanation from the salient features. 
A total of four attention mechanisms have been proposed and compared with the BLEU metric (Papineni et al. \citeyear{DBLP:conf/acl/PapineniRWZ02}) for the textual explanation. The best of these is called \emph{adaptive attention}, which consists of assigning a weight to the features dynamically using the last word generated by the verbal explanation.

                    % Wang (+ global insight through 2k input 'explained) 
Wang and Wang (\citeyear{DBLP:conf/isc2/WangW22}) use an attention mechanism on non-visual inputs to understand the agent's decision choices.
                    % Zhang (task relevant features (not images))
In the same idea, Zhang et al. (\citeyear{DBLP:journals/ral/ZhangMYLYLL21}) present the Temporal-Adaptive Feature Attention algorithm. This identifies the most relevant features in the agent's choice of action.
In MuJoCo (Todorov et al. \citeyear{DBLP:conf/iros/TodorovET12}) environments, the weights of attention show that the features specific to the agent's position are more important than those linked to its velocity in the choice of action throughout the duration of an episode.

                    % LRP Huber 
The approach proposed by Huber et al. (\citeyear{DBLP:conf/ki/HuberSA19}) does not require any additional module or specific architecture, but generates, as with the attention mechanisms, a saliency map. To do this, Layer-wise Relevance Propagation (Bach et al. \citeyear{bach2015pixel}) is used: this is a general concept that identifies relevant pixels during the forward-pass by looking at the activations of each neuron.
To limit the information to the most relevant, the authors use Mopuri et al.  (\citeyear{DBLP:journals/tip/MopuriGB19}): during the forward pass, neurons are monitored layer by layer, and only the neuron with the highest activation in each convolutional layer is selected. Relevance propagation is then restricted to the paths that connect these selected neurons across layers, allowing to filter out less relevant contributions and produce compact saliency maps.

                \paragraph{Perturbation-based Approach}
                \label{par:App_C_perturbation}

                % Perturbation: 
                    % Greydanus (Value function)
The perturbation method proposed by Greydanus et al. (\citeyear{DBLP:conf/icml/GreydanusKDF18}) involves adding spatial uncertainty around the pixel at coordinates $(i,j)$. This is done by interpolating the original image with a Gaussian blur applied around $(i,j)$.
In practice, instead of applying this perturbation for each pixel, it was applied per group of 5 pixels, which reduces the computational cost and produces good saliency maps. 
Using the A3C learning algorithm, saliency maps are generated and displayed jointly for the agent's policy and for its value function $V$ (i.e. this captures the regions that are important for the agent's choice of action and evaluation of the image).
                    
                    % Persiani (Greyd.)  + other XRL?
We have identified three works based on this method.
Built on top of the agent implementation, Persiani and Hellström (\citeyear{DBLP:conf/atal/PersianiH22}) propose a new architecture, called the Mirror Agent Model, to make the agent's behavior interpretable by the user. Both the agent model and the user model are represented by Bayesian networks. 
A metric is presented to measure the distance between the two models. The explanatory layer added to the agent (which adds a node to the Bayesian network) consists of using saliency maps based on the method of Greydanus et al. %described previously
                    % Guo (compare with human attention)
This method is also used in S. Guo et al. (\citeyear{DBLP:conf/nips/GuoZLZBHS21}) to compare the saliency maps explaining the agent with the visual attention of humans. To do this, this study uses the Atari-head dataset (R. Zhang, Walshe, et al. \citeyear{DBLP:conf/aaai/ZhangWLGMWZHB20}) and AGIL (R. Zhang, Z. Liu, et al.  \citeyear{DBLP:conf/eccv/ZhangLZWMHB18}), a model for predicting human visual attention.
The authors found that, as learning progressed, the agent's attention maps became closer to human visual attention, which can also be used to understand agent failures. 
                    % Douglas (+specific viz.) 
By modifying the method of Greydanus et al. to apply it for a 2D grid-world environment, Douglas et al. (\citeyear{DBLP:conf/tabletop/DouglasYKHMT19}) propose a display called Towers of Saliency, to interpret the agent's behavior over an entire episode.
                    
                    % Puri 
Puri et al. (\citeyear{DBLP:conf/iclr/PuriVGKDK020}) construct saliency maps based on two properties: specificity and relevance. This method is called Specific and Relevant Feature Attribution (SARFA).
The Specificity property defines that the salience of a feature must focus on the perturbation of the action to be explained, $a$. Thus, with $s$ the base state and $s'$ the perturbed state, the salience must be high if $Q(s,a) - Q(s',a)$ is substantially greater than $Q(s,a') - Q(s',a')$, $\forall a' \neq a$. 
The Relevance property defines that the salience of a feature should only have an impact on the $Q$ value of $a$. In other words, the salience of a feature must be low if the perturbation also affects the $Q$ values of the other actions.

                    % Yan (advantage function based)
Rather than directly using the $Q$ function (Puri et al.) or the $V$ function (Greydanus et al.) to measure the impact of a perturbation, Yan et al. (\citeyear{yan2023research}) propose using the Advantage function. Given a state $s$, an action $a$ and a policy $\pi$, the advantage of performing $a$ from $s$ and then following $\pi$ is: $A^{\pi}(s,a) = Q^{\pi}(s,a) - V^{\pi}(s)$. To locally perturb the image, a Gaussian blur is applied (as in Greydanus et al.).

                    % Huber (benchmark + identify problems)
Huber et al. (\citeyear{DBLP:journals/frai/HuberLA22}) evaluate a total of 5 perturbation-based approaches by inspecting the dependency of the methods on the NN parameters learned by the agent using the \emph{sanity checks} metric (Adebayo et al.  \citeyear{DBLP:conf/nips/AdebayoGMGHK18}) as well as their fidelity on the agent's reasoning using the \emph{insertion} metric (Petsiuk et al. \citeyear{DBLP:conf/bmvc/PetsiukDS18}).
Among these approaches, there are the two aforementioned works: SARFA (Puri et al.) and Noise Sensitivity (Greydanus et al.).
\emph{Sanity checks} consist of successively randomising the layers of the neural network and calculating a saliency map each time. These saliency maps are then compared with the original, with the expected positive result being that the maps differ significantly from the original. All the methods depend on the learned parameters, but Noise Sensitivity shows little dependency.   
The \emph{insertion} metric starts from a noisy image (using black occlusion and uniform random perturbation) and iteratively reconstructs the base image by resetting the correct pixel values, starting with the most salient pixels indicated by the saliency map. The expected positive result is that the agent's action prediction should quickly be similar to that of the agent on the noiseless image. This would mean that the saliency map has put forward the `critical' pixels in the agent's choice of action. With this metric, SARFA is one of the two best approaches. 
The authors suggest further research into the generation of perturbation-based saliency maps, by first determining which type of perturbation might be appropriate for the task in hand.

                    % Object:
The following papers use saliency maps to focus on objects rather than pixels.
                    % (template matching) (Q-value based) Iyer (perturbation-based but not called)
The work by Iyer et al. (\citeyear{DBLP:conf/aies/IyerLL0SS18,DBLP:conf/gcai/LiSI17}) uses a computer vision technique called template matching to recognise objects in the image and use them as a basis for saliency maps. To provide the agent with more information for choosing an action, object channels are added to the base image.
These object channels are extracted using the recognizer object and are used to determine the positions of the various objects detected in the image. The perturbation in the image $s$ consists of masking an object, resulting in $s_o$ and determining its impact on the choice of action by comparing the $Q$ values. 
The maps generated highlight the objects that are salient in the agent's decision-making.
A comparison between a `classic’ pixel saliency map and an object saliency map, based on a state-action pair from a game of Atari 2600 Ms PacMan (Bellemare et al.  \citeyear{DBLP:journals/jair/BellemareNVB13}), is shown in Figure \ref{fig_ch1:object_saliency_map}.
                    % Anderson (for object) (perturbation based) 
In Anderson et al. (\citeyear{DBLP:journals/tiis/AndersonDSJNICO20}), a user study was carried out comparing object saliency maps with reward decomposition (which is described later), a combination of the two and the control strategy, which simply displays the impact of the action on the state and the score obtained. In summary, participants understood the agent's behavior better when the explanation contained reward decomposition, although this may have led to cognitive overload for some participants.

%    \begin{figure*}[]
%        \centering
%        \includegraphics[width=1.0\linewidth]{images/Object_saliency_map.png}
%        \caption[Two saliency maps for a state-action pair from the Ms PacMan environment.]{Two saliency maps for a state-action pair from the Ms PacMan environment \cite{DBLP:conf/aies/IyerLL0SS18}. The action is \emph{`move right'} and the sub-figure `a' is the state from which a pixel saliency map (sub-figure `b') and an object saliency map (sub-figure `c') are generated. The saliency of an element is represented using a grey scale, where the darker the element, the more salient it is. In sub-figure `c', we can see that the agent has focused on Ms PacMan and the pills in the right path to choose the \emph{`move right’} action. Conversely, it did not focus on the part of the image that symbolises 1 life for the player, as shown by the white square in the object saliency map.}
%        \label{fig_ch1:object_saliency_map}
%    \end{figure*}

                \paragraph{Gradient-based Approach}
                \label{par:App_C_gradient}

                % Gradient: 
Gradient-weighted Class Activation Mapping (Grad-CAM) (Selvaraju et al. \citeyear{DBLP:conf/iccv/SelvarajuCDVPB17}) is a method that can be used on CNN's without having to modify the architecture or re-train the model, in order to create saliency maps.
To find out the influence of pixels on a class c for a given image, a forward pass is performed, the gradient of class c is calculated, then the signal is propagated with a backward pass. 
Importance weights of the feature maps (where a feature map is the output of a convolutional layer of a CNN) are calculated and then the saliency map is generated by applying a ReLU operation on the weighted linear combination of the feature maps. This method has been used in several works to explain RL agents (Dao, Huff et al. \citeyear{DBLP:conf/ssci/DaoHL21}; Joo and K. Kim \citeyear{DBLP:conf/cig/JooK19}; Nie et al. \citeyear{DBLP:conf/ies2/NieHO19}; Weitkamp et al. \citeyear{DBLP:journals/corr/abs-1902-00566}).

                    % Joo (Grad-CAM)
Joo and K. Kim propose an architecture combining A3C (Mnih, Badia, et al.  \citeyear{DBLP:conf/icml/MnihBMGLHSK16}) with Grad-CAM in order to visualise the salient parts of the image for the agent's choice of action.
                    % Nie ('Visualizing Deep Q-Learning to Understanding Behavior of Swarm Robotic System') (Grad-CAM + DeconvNet)
To describe the behavior of a swarm robotic system, Nie et al. combined it with a deconvolutive network to produce saliency maps.
                    % Weitkamp
Weitkamp et al. use it by replacing ReLU activation with ELU activation.
                    % Dao (Grad-CAM + summarize crit. state) ('Learning Sparse Evidence- Driven Interpretation to Understand Deep Reinforcement Learning Agents')
To reduce the number of important snapshots from the agent's learning phase, Dao, Huff, et al. use Grad-CAM. Similar states with common attentions are thus grouped together. The SmoothGrad method (Smilkov et al. \citeyear{DBLP:journals/corr/SmilkovTKVW17}), which adds noise to the image so that the Grad-CAM saliency map is less noisy, was used. Note that in this paper, the method is not used to explain, but to make another explainable approach more sparse.
                    
                    % Hilton
An in-depth analysis of an agent that has learned in the CoinRun environment (Cobbe et al. \citeyear{DBLP:conf/icml/CobbeKHKS19}) is proposed by Hilton et al. (\citeyear{hilton2020understanding}) by analysing a hidden layer of the agent that has learned to recognise objects in CoinRun.
The integrated gradient method (Sundararajan et al. \citeyear{DBLP:conf/icml/SundararajanTY17}) is used to generate saliency maps.
                    
                    % Jacobian
In order to provide additional information for the visual analysis of a DQN, Zahavy et al. (\citeyear{DBLP:conf/icml/ZahavyBM16}) calculate the Jacobian of the DQN as a function of the input image.

                    % SHAP-CAM He (with shap to vector input + global explanations based on these explainable methods)
He et al. (\citeyear{he2021explainable}) propose an approach that does not use gradients, but works in a similar way to Grad-CAM. In addition to the saliency maps, textual explanations are provided. The agent uses an image and a feature vector to choose an action $a$.
The explanation of $a$ based on the image is produced by a new method using both CAM (B.  Zhou et al. \citeyear{DBLP:conf/cvpr/ZhouKLOT16}) (on which Grad-CAM is based) and SHAP (Lundberg and S. Lee \citeyear{DBLP:conf/nips/LundbergL17}). The textual explanation of $a$ based on the vector is provided using SHAP.
Note that both approaches are also used to provide global explanations of agent behavior.

    %------------------------------------------------
    \subsubsection{Model-agnostic Approach}
    \label{sec:App_C_model-agnostic_approach}
    %------------------------------------------------

            % Model Agnostic approaches
            \paragraph{SHAP}
            \label{par:App_C_SHAP}
Most of the works focus on SHAP, using either Kernel SHAP (Carbone \citeyear{carbone2020explainable}; Løver et al. \citeyear{lover2021explainable}; Remman, Strümke, et al. \citeyear{DBLP:conf/amcc/RemmanSL22}; Rizzo et al. \citeyear{DBLP:conf/itsc/RizzoVC19}; Y. Wang et al. \citeyear{DBLP:conf/aaaiss/WangME20}) or DeepSHAP (Liessner et al. \citeyear{DBLP:conf/icaart/LiessnerDW21}; Remman and Lekkas \citeyear{DBLP:conf/eucc/RemmanL21}; Schreiber et al. \citeyear{schreiber2021towards};  Theumer et al. \citeyear{theumer2022explainable}; K. Zhang et al. \citeyear{DBLP:journals/tcss/ZhangZXGG22}).

                % SHAP 
                    % Kernel SHAP Rizzo
                    % Kernel SHAP Wang
For example, Rizzo et al. use Kernel SHAP for a traffic light control problem and Wang et al. for an automatic crane control problem.
                    % Kernel SHAP Carbone (comparison with LMT)
                    % kernel SHAP + k-means Lover (comparison with LMT + LIME)
Two works (Carbone; Løver et al.) propose to compare SHAP with linear model trees (LMT), already presented in the policy explanation section. In Carbone, the author suggests that LMT and SHAP `capture similar relationships between input features'.  Løver et al. also compare LIME and use a K-means summarizer to reduce the number of samples to be used for the calculation of SHAP values. Both studies show that LMT is the fastest at providing an explanation, and Løver et al. state that LIME is too slow to be used in real time.
                    % Causal SHAP describes direct and indirect effect of a feature (alteration of Kernel SHAP + comparison with Kernel SHAP) Remman
Kernel SHAP and Causal SHAP (Heskes et al. \citeyear{DBLP:conf/nips/HeskesSBC20}) are compared in Remman, Strümke, et al. Causal SHAP is a variant of Kernel SHAP which computes SHAP values by taking into account the dependencies between the features of a state. The results highlight the benefits of using Causal SHAP.
                    
                    % DeepSHAP Zhang
                    % DeepSHAP Theumer (context: MARL (local one agent and local all agents))
DeepSHAP is used in K. Zhang et al. for a power system emergency control problem and in Theumer et al. where the explanation in a MARL context consists of an explanation of the decision of a single agent or the decisions of all agents.
                    % Deep SHAP joint variables of important variables are not Remman
Remman and Lekkas use DeepSHAP to understand the impact of different variables in a robotic lever manipulation problem. Values are calculated and displayed for each agent's decision over an entire episode.
                    % (not specified) Liessner
                    % (not specified) Schreiber
Based on the same use, Liessner et al. propose to calculate values with DeepSHAP in a longitudinal control task and Shreiber et al. in a traffic signal control task.

                    % SHAP Sequeira interpretable features need to be set
Within the IxDRL framework (Sequeira and Gervasio \citeyear{DBLP:conf/xai/SequeiraG23}), SHAP is used both to study the impact of features on the agent's decision-making in a single state and also globally, by displaying the average impact of features on a set of states.
                    % Beechey
In the same vein, Beechey et al. (\citeyear{DBLP:conf/icml/BeecheySS23}) propose two methods for explaining locally (i.e. from a given state) and globally the agent's performance, i.e. its expected return.
                    
                    % New algo Tabular SHAP (Xiong)
In addition to proposing a benchmark specific to the XRL domain, Xiong et al. (\citeyear{DBLP:journals/corr/abs-2402-12685}) present an algorithm entitled TabularSHAP. It is limited to explaining tabular states.
To use it, the user first needs a set of interactions of the agent within the environment, which will be used to learn an ensemble tree model. Finally, TreeSHAP (Lundberg, Erion, et al. \citeyear{DBLP:journals/corr/abs-1802-03888}), which is specific to tree-based models, is used on the basis of the ensemble tree model learned.

                \paragraph{LIME}   
                \label{par:App_C_LIME}
                % LIME
                    % Kotevska
In order to analyse the behavior of an agent on a heating ventilation and air-conditioning control task, Kotevska et al. (\citeyear{DBLP:conf/bigdataconf/KotevskaMKDASZ20}) propose three modules: Model assessment, Model local view and Model global view.
The first consists of a probabilistic analysis of the agent's behavior and a statistical analysis of the impact of features on its behavior. 
The second involves using LIME for an agent decision and the third consists of using visualisation tools, e.g. a partial independence plot, to understand the impact of features on agent decisions. 
                    % Dethise
As already mentioned, Dethise et al. (\citeyear{DBLP:conf/sigcomm/DethiseCK19}) uses LIME to explain a single agent's decision, or the agent's global behavior.
                    % Li (paper: 'Explainable Intelligence-Driven Defense Mechanism Against Advanced Persistent Threats: A Joint Edge Game and AI Approach')
LIME is also used in H. Li et al. (\citeyear{DBLP:journals/tdsc/LiWXLG22}) to add an explanatory layer to an advanced persistent threats defense mechanism.

    %------------------------------------------------
    \subsubsection{Counterfactual State}
    \label{sec:App_C_counterfactual_state}
    %------------------------------------------------
A body of work aims to provide an answer to the question \emph{`Why does the agent perform action $a$ from state $s$ rather than action $a'$?'} by proposing a counterfactual state $s'$, close to $s$, in which the agent would have chosen $a'$.

% GAN
A number of studies (Huber, Demmler, et al. \citeyear{DBLP:conf/atal/HuberDMOA23}; Olson et al. \citeyear{DBLP:journals/ai/OlsonKNLW21}; Samadi et al. \citeyear{DBLP:journals/corr/abs-2404-18326}) have used GAN (Goodfellow et al. \citeyear{DBLP:journals/corr/GoodfellowPMXWOCB14}) for this purpose. As a reminder, this architecture consists of jointly learning a generative model $G$ that creates data and a discriminative model $D$ that determines whether the input data is generated or comes from the training dataset. After learning, only the generative model is used.

                % Olson
In order to provide a counterfactual explanation where the state is an image, Olson \emph{et al.} propose to learn, in addition to these two models, an encoder which encodes the state $s$ in order to omit information from $s$ related to the action choice.
$D$ is trained to determine the action distribution based on the encoded state. 
$G$ is trained to reconstruct $s$ based on the action distribution and the encoded state. For training, it is necessary to get an initial dataset of state-action pairs using the agent to be explained.
This method, tested on Atari2600 games, generates a state $s'$, close to $s$, in which the agent performs a different action.

                % GANterfactual Huber
In the same line, GANterfactual-RL (Huber, Demmler, et al.) is based on the StarGAN architecture (Choi et al. \citeyear{DBLP:conf/cvpr/ChoiCKH0C18}). This helps $G$ to construct states in which the agent chooses the desired counterfactual action.
Examples of explanations are shown in Figure \ref{fig_ch1:ganterfactual}.
With the help of a user study and metrics, GANterfactual-RL is more efficient and has better results in terms of metrics than the previous approach mentioned above (Olson et al.).
Among these metrics, proximity and sparsity measure the distance of the generated counterfactual state $s'$ from the base state $s$, and validity verifies that from $s'$, the action predicts the desired action $a'$. 

%    \begin{figure*}[]
%        \centering
%        \includegraphics[width=0.9\linewidth]{images/GAN-terfactual_comparison.png}
%        \caption[Comparison of counterfactual states in the Ms PacMan environment.]{Comparison of counterfactual states in the Atari 2600 Ms PacMan environment \cite{DBLP:journals/jair/BellemareNVB13} generated on the basis of a state-action pair (on the left) \cite{DBLP:conf/atal/HuberDMOA23}. The agent's action is \emph{‘move right’} and the counterfactual one is \emph{‘move left’}. The state in the middle, where Ms PacMan is missing, is generated using \cite{DBLP:journals/ai/OlsonKNLW21} and the state on the right, where Ms PacMan is safe in the left part of the map, is generated using \cite{DBLP:conf/atal/HuberDMOA23}. The first counterfactual state is not a convincing one as it shows that the agent chooses \emph{‘move left’} while Ms PacMan does not even appear, whereas in the second one, Ms PacMan is (partially) visible.}
%        \label{fig_ch1:ganterfactual}
%    \end{figure*}

                % Use of GAN with saliency map as input Samadi
Another competitive approach is proposed by Samadi et al. to generate counterfactual states using saliency maps.
The proposed method, called SAFE-RL, is based on attentionGAN (H. Tang et al.  \citeyear{DBLP:journals/tnn/TangLXTS23}). G learns to generate a counterfactual state by taking as input the state, the associated saliency map and a counterfactual action. D learns to differentiate true states from counterfactuals.
                
                % Counterfactual states + Metrics to analyse agent + user interface Druce
With states also corresponding to images, Druce et al. (\citeyear{DBLP:journals/corr/abs-2106-03775}) present counterfactual states in their interface to help user comprehension. In this work, counterfactual states are obtained by pre-defined interventions on state $s$, such as the player's position in the image.
For the same state, several counterfactual states are displayed to see how the agent would have acted according to certain modifications.

                % RACCER Gajcin : generate a set of actions to use to arrive in a state x' in which action a' has a high probability to occur.
The RACCER algorithm (Gajcin and Dusparic \citeyear{DBLP:conf/atal/GajcinD24}) is used to identify a sequence of actions that leads to a counterfactual state in which the agent has a high probability of performing the desired action. This method is not image-specific, unlike the previous methods, and provides a recourse for the user to perform a certain action (with a certain probability).
However, RACCER requires access to the environment dynamics.

    %------------------------------------------------
    \subsubsection{Others}
    \label{sec:App_C_others}
    %------------------------------------------------

                    % Risk actionable features Davoodi
To use the method proposed by Davoodi and Komeili (\citeyear{DBLP:conf/smc/DavoodiK21}), the features of the agent's state must be actionable. A feature is said to be actionable if the user knows which action or sequence of actions will increase/decrease the value of the feature. 
The aim of the method is to explain, for a given state $s$, the impact of features in reaching risky states. To do this, a transition graph $T$ limited by a number of actions is constructed from $s$, then a linear model is learned on all the states of $T$ to estimate the risk. 
The model weights describe the importance of the different features in reaching a risky state.

                % Extract the most relevant variable + extra domain knowledge info (Elizalde) MDP context
In Elizalde et al. (\citeyear{DBLP:conf/micai/ElizaldeSNR09}), the aim is to explain a recommended action in an MDP by returning the most relevant feature of the current state $s$, with respect to the value function $V^{\pi}$ (which represents the expected reward of following $\pi$ from $s$). To determine this feature, we need to measure the impact of the change in valuation of each feature of $s$ on $V^{\pi}$.
Each feature is tested separately, leaving the other features fixed. In addition to the relevant feature, a verbal explanation is proposed based on a knowledge-base dataset.

%------------------------------------------------
%------------------------------------------------
\subsection{Expected Outcomes}
\label{sec:App_C_expected_outcomes}
%------------------------------------------------
%------------------------------------------------

    %------------------------------------------------
    \subsubsection{State}
    \label{sec:App_C_state}
    %------------------------------------------------
                    
                \paragraph{Causal Lens}
                \label{par:App_C_causal_lens}
                % Causal Lens (Structural Causal Models Halpern)
                    % Madumal / Madumal(bayesian network?)
In order to focus on causal relationship between action and state variables, Madumal et al. (\citeyear{DBLP:conf/aaai/Madumal0SV20}) build an Action Influence Model (AIM) which is a structural causal model (Halpern and Pearl \citeyear{DBLP:conf/ijcai/HalpernP01}) with the addition of actions.
This model takes the form of a graph which is used to answer \emph{`Why?'}, \emph{`Why not?'} questions by providing a (partial) causal chain describing the impact of the action on the features of the model. 
An extension of this work (Madumal et al. \citeyear{DBLP:journals/corr/abs-2001-10284}) introduces a distal explanation model to provide causal chain explanations using a decision tree representing the agent's policy.
The shortcoming of this work is that the causal model is hand-crafted.

                    % Volodin (learned SCM?)
CauseOccam (Volodin \citeyear{volodin2021causeoccam}) is a model-based RL framework where the model learned is a sparse causal graph.

                    % Parsimonious causal graph learnt (not explicit XRL) Herlau
Herlau and Larsen (\citeyear{DBLP:conf/aaai/HerlauL22}) also propose learning a sparse causal graph, which the agent uses during training. The search for relevant causal variables is carried out by maximising the \emph{`natural indirect effect'}. These two approaches could serve as a basis for the construction of AIMs.

                    % Yu 
With the same idea, Yu et al. (\citeyear{DBLP:conf/ijcai/YuRX23}) extend the use of AIM to problems with a continuous action space, without the need of prior knowledge of the environment causal structure. Indeed, to represent the dynamics of the environment, the causal model is learned based on the agent's interactions with the environment.
An example of a learned causal model for the LunarLander environment (Brockman et al.  \citeyear{brockman2016openai}) is shown in Figure \ref{fig_ch1:AIM}.

%    \begin{figure}[]
%        \centering
%        \includegraphics[width=0.6\linewidth]{images/AIM_YU.png}
%        \caption[Causal model for the LunarLander environment.]{Causal model for the LunarLander environment \cite{DBLP:conf/ijcai/YuRX23}. The features of a state are in blue, the actions in orange and the outcomes in green. An arrow between two elements describes a causal relation.}
%        \label{fig_ch1:AIM}
%    \end{figure}

                \paragraph{Probability}
                \label{par:App_C_probability}
                % State + probability
                % Memory-based Cruz (learn add. component)
The Memory-based eXplainable Reinforcement Learning approach (Cruz, Dazeley, and Vamplew \citeyear{DBLP:conf/ausai/CruzD019}) provides the probability of success (i.e. reaching a goal state) and the number of transitions to reach the goal state. When the agent is trained, the success probabilities $P$ and the number $T$ of transitions leading to a goal state are calculated and updated for each state.
The agent can then provide answers to the questions \emph{`Why does the agent choose action a?'}, \emph{`Why does the agent choose action a and not b?'} or \emph{`What is the probability of reaching a goal state in 8 steps from s?'}. The problem with this approach is that it is necessary to memorise $P$ and $T$ in order to explain. 
                % Memory / Learning based and Introspect (without additional memory) (learn add. component) Cruz
To overcome this problem, Cruz, Dazeley, Vamplew, and Moreira  (\citeyear{DBLP:journals/nca/CruzDVM23}) propose 2 additional methods for calculating or approximating P: learning-based and introspection-based approaches.
The first approach learns a P-table of the probability of success in parallel with learning the agent's Q-table. 
The second requires no additional memory, as the probability of success is calculated directly from the Q-values.

                    % MSE Factored MDP Khan
In a Factored MDP context, Khan \emph{et al.} (\citeyear{DBLP:conf/aips/KhanPB09}) introduce Minimal Sufficient Explanations (MSE) to provide a minimal explanation in terms of the number of templates sufficient to explain the agent's action.
Templates are phrases used to describe the probability of reaching a certain state (e.g. `[action] is likely to take you to [state description] about [lambda] times').

                % natural language: action factored differential values + statistics explanation based on database Dodson (MDP context)
Dodson et al. (\citeyear{DBLP:conf/aldt/DodsonMG11}) present a system that provides an explanation of the optimal action via a natural language paragraph in a MDP context.
One of the two explanation modules proposed is called Case-based explanation (CBE). CBE explains the usefulness of an action by using a database of previous transitions, called \emph{cases}, to calculate statistics on the \emph{cases} relevant to the recommended action. In terms of probabilities, this explanation makes it possible to explain the influence of an action on future feature values.

                    % ESX aucune publis pour l'instant :(
%Our Expected States eXplanation method presented in Chapter \ref{chap:ESX-chapter}  is similar to this type of work. The aim is to provide the user with a partial and representative description of a set of states that can be reached at horizon $k$ by taking action $a$ from state $s$.

                \paragraph{Contrastive}
                \label{par:App_C_contrastive}
                    % Counterfactual explanations based on subgoal abstraction (likelihood to reach goal + action cost) Stein
Stein (\citeyear{DBLP:conf/nips/Stein21}) provides counterfactual explanations based on the abstraction of sub-objectives. These sub-goals must be completed in order to achieve the agent's goal. 
Thus, the user can ask why the agent prefers to perform one sub-goal rather than another. The answer contrasts the two sub-goals with the probability of achieving the final goal (i.e. a state) and the cost of achieving the sub-goal. 

                    % Expected transition Tsuchiya
Tsuchiya et al. (\citeyear{DBLP:conf/aaaiss/TsuchiyaME23}) present a method similar to reward decomposition (which will be described in the next section), but focused on states. Additional neural networks are learned to estimate the Q-value of each state type.
The different state types are provided in advance by the user. 
A contrastive explanation between the agent's choice of action and that of the user is performed by displaying and then comparing the different Q-values decomposed by state type.

                % Interpretable action-value Lin
Contrastive explanations are proposed using action-values based on given human-readable features (Z. Lin et al. \citeyear{DBLP:conf/iclr/LinLF21}). To this end, deep generalized value functions (Sutton, Modayil, et al. \citeyear{DBLP:conf/atal/SuttonMDDPWP11}) are learned to predict future feature accumulation. To avoid information overload, the MSX approach (Juozapaitis et al.  \citeyear{juozapaitis2019explainable}) (described in the next section) is applied to obtain a clear comparison of the features impacted by the actions. Note that features can represent the state of the agent, but also components of the reward function.

                % From Q-values to moods concerning agent's performance Barros
The Moody framework (Barros et al. \citeyear{DBLP:conf/icdl-epirob/BarrosT0S20}) allows the agent to self-assess these decisions. Based on the agent's Q-values, its confidence of reaching the final state is calculated (Cruz, Dazeley, Vamplew, and Moreira), then converted into a pleasure/arousal scale (Costa Jr and McCrae  \citeyear{costa1996mood}) consisting of two dimensions: pleasing/unpleasing and excited/calm. Finally, a Growing-When-Required Network (Marsland et al.  \citeyear{DBLP:journals/nn/MarslandSN02}) is used to define the agent's current mood. Note that this method can also be used by an agent to evaluate the mood of other agents.
                
                \paragraph{Others}
                \label{par:App_C_others_state}
                    % (Context: Interactive Reinforcement Learning Thomaz)  Re-use instructions as behavior explanations (change in environment state) Fukuchi / Fukuchi
Interactive RL (Thomaz et al. \citeyear{thomaz2005real}) is a framework in which the agent's learning is accelerated by instructions given by an expert. In this context, the works by Fukuchi et al. (\citeyear{DBLP:conf/iconip/FukuchiOYI17,DBLP:conf/hai/FukuchiOYI17}) develop the Instruction-based Behavior Explanation (IBE) framework, which reuses expert instructions to explain the agent's choice of action $a_t$ from state $s_t$. To do this, IBE estimates $\delta(s_t)$ which represents the change in value of the features between state $s_t$ and the state reached $n$ steps later $s_{t+n}$. Then, an expert expression, which is a user interpretable signal, is associated to $\delta(s_t)$, either by a clustering method (Fukuchi et al.  \citeyear{DBLP:conf/iconip/FukuchiOYI17}), or by classification using a NN (Fukuchi et al. \citeyear{DBLP:conf/hai/FukuchiOYI17}).

                    % Semantic meaning of future state
                    % Semantic segmentation of future state (image) Pan
With a state represented by an image, Pan et al. (\citeyear{DBLP:conf/icra/PanCCCY19}) propose to predict the semantics of pixels in future states. Image semantic segmentation consists of labelling each pixel of an image.
It is used here in a vehicle driving context, where each pixel can be labelled as \emph{road, grass, sky, ...} The agent learns to predict this visual representation of the future state, which acts as an explanation.

    %------------------------------------------------
    \subsubsection{Reward}
    \label{sec:App_C_reward}
    %------------------------------------------------

The vast majority of works uses the principle of reward decomposition: Amitai, Septon et al. \citeyear{DBLP:conf/aaai/AmitaiSA24}; Anderson et al. \citeyear{DBLP:journals/tiis/AndersonDSJNICO20}; Erwig et al. \citeyear{erwig2018explaining}; Feit et al. \citeyear{DBLP:conf/acsos/FeitMP22}; Iucci et al. \citeyear{DBLP:conf/icar/IucciHTIL21}; Juozapaitis et al. \citeyear{juozapaitis2019explainable}; Rietz et al. \citeyear{DBLP:journals/nca/RietzMHSWS23}; Saldiran et al. \citeyear{saldiran2024towards}.

As a reminder, reward decomposition is used to represent the reward function in an interpretable way: it is decomposed into a set of comprehensible reward types. Thus, to understand the agent's choice of action, it is sufficient to look at the type of reward it seeks to maximise.
                % Expected reward
                    % Erwig 
                    % Juozapaitis 
Based on this method, (Erwig et al.; Juozapaitis et al.) propose Reward Difference eXplanations (RDX) and Minimal Sufficient eXplanation (MSX) to explain deep adaptive programs learned via RL (Erwig et al.) and agents from the CliffWorld and LunarLander domains (Juozapaitis et al.).
RDX simply compares 2 actions from a state by comparing their Q-values for each type of reward. An example of RDX, from Iucci et al., is shown in Figure \ref{fig_ch1:RDX}.
In accordance with RDX, MSX provides a more compact explanation of an agent's preference for action $a$ over another action $a'$: only the reward types that have the greatest impact (positively and negatively) on the preference of $a$ over $a'$ are displayed to the user.  

%    \begin{figure}[]
%        \centering
%        \includegraphics[width=0.4\textwidth]{images/RDX_Iucci.png}
%        \caption[RDX for a Human-Robot collaboration task.]{RDX for a Human-Robot collaboration task \cite{DBLP:conf/icar/IucciHTIL21}. Two actions (related to robot speed) are compared across four different reward types (on the x-axis of the chart). The y-axis shows the loss and gain for each type of reward when performing the agent's action rather than the other one.}
%        \label{fig_ch1:RDX}
%    \end{figure}
                    
Several works combine reward decomposition (by simply displaying the different Q-values or by using RDX, MSX) with other methods. 
Of these, most combine it with approaches that explain the agent's policy. 
                    % Saldiran
Saldiran et al. use a global reward decomposition and a heatmap visualisation of Q-values.
                    % Iucci
                    % Use of reward decomp (+interestingness elements) Feit
                    % Rietz (mix with other XRL)
Iucci et al. use autonomous policy explanation (Hayes and Shah \citeyear{DBLP:conf/hri/HayesS17}), Feit et al. the Interestingness Elements framework (Sequeira and Gervasio \citeyear{DBLP:journals/ai/SequeiraG20}) and Rietz et al. the HRL setting.
                    % Anderson (mix with other XRL)
                    % Amitai
Anderson et al. employ object perturbation-based saliency maps, a complementary explanation of the agent's choice of action. Amitai, Septon, et al. present a counterfactual explanation using reward decomposition.

                % Temporal RD
A new estimator proposed in Towers et al. (\citeyear{DBLP:conf/ecai/Towers0FN24}) allows the agent to predict the $n$ future expected rewards from a given state. This new estimator makes it possible to explain the agent's choice of action from a given state through three questions: \emph{`What rewards to expect and When?'}, \emph{`What observation features are important?'} and \emph{`What is the impact of an action choice?'}. The answers to these questions from a state $s$ are respectively a plot of the $n$ future expected rewards, a use of Grad-CAM (Selvaraju et al.  \citeyear{DBLP:conf/iccv/SelvarajuCDVPB17}) to understand the important features for the different future expected rewards and a contrastive explanation by displaying the $n$ future expected rewards for two different actions performed from $s$.

                % Explaining sympathetic actions (human-robot interaction) (not RL but can be used for it) Kampik
In a human-machine cooperation context, Kampik et al. (\citeyear{DBLP:conf/atal/KampikNL19}) describe different types of goals that the agent could communicate to explain a sympathetic action, i.e. an action that helps the user at the agent's expense.
Although not for RL, this type of explanation could be useful in a cooperative MARL context, where agents would explain their sympathetic action by describing the expected outcome in terms of reward.

    %------------------------------------------------
    \subsubsection{Sequence}
    \label{sec:App_C_sequence}
    %------------------------------------------------

            % Trajectory
                % Belief map Yau what chain of events the agent intended to happen as a result of a particular action choice. (learn add. component)
A belief map is learned together with the agent's Q-values, to give the user the states that the agent is trying to reach by doing $a$ from $s$ (Yau et al.   \citeyear{DBLP:conf/nips/Yau0H20}).
This belief map provides a local representation of the agent's intentions. In the Taxi problem, these expected states form a path, as shown in Figure \ref{fig_ch1:belief_map}.

%    \begin{figure*}[]
%        \centering
%        \includegraphics[width=1.0\textwidth]{images/Agent_intention.png}
%        \caption[Explanation of each action in an episode in the Taxi environment.]{Explanation of each action in an episode in the Taxi environment \cite{DBLP:conf/nips/Yau0H20}. The episode is described on two lines, starting on the left of the first line. In each state, the agent's future positions according to the belief map are displayed by colored cells. The color intensity reflects the agent's confidence in accessing the position (the more intense the colour, the more confident the agent). In this episode, the Taxi picks up the passenger at the position marked with a `G' and drops him off at the position marked with a `B'.}
%        \label{fig_ch1:belief_map}
%    \end{figure*}

                % Counterfactual action outcomes Amitai (+reward decomposition)
The CoViz algorithm (Amitai, Septon, et al. \citeyear{DBLP:conf/aaai/AmitaiSA24}), already presented in this survey with its policy summary variant, proposes to display to the user two sequences from a state $s$, one in which the action $a$ chosen by the agent is performed, and one in which a different action $a'$ is performed.
This enables the user to compare the sequences. In addition, reward decomposition is used.

                % Learn clauses Optimal trajectories (MDP context) Rouveirol (same idea as SXp)
In Rouveirol et al. (\citeyear{DBLP:conf/rulemlrr/RouveirolASV23}), the set of possible trajectories from a pair (s,a) is divided into different groups that can be explained in the same way, by training a rule-based classifier.
Subsequently, a minimal common explanation is built for each group, which takes the form of an existentially quantified conjunction of literals. 
In a MDP context, these explanations describe a partial plan made up of actions and/or state features to reach a winning state. 

                % SXp
The Scenario eXplanation (SXp) method (Saulières et al. \citeyear{DBLP:conf/icaart/SaulieresCS23}) falls into this category of work. Its objective is to summarise all possible sequences starting with action $a$ from state $s$, by simply providing $3$ representative sequences: the best-case, worst-case and most probable scenario.

                % Local clustering  states Luss 
Luss et al. (\citeyear{DBLP:conf/aaai/LussDL23}) present a method for clustering the state space locally and extracting a set of critical states. Starting from a state s, the states to cluster are obtained using the stochastic policy for $x$ steps.
Reachable states are grouped into meta-states and strategic states are identified. These states are identified as sub-goals. With the help of visualisation, these sub-goals and clusters are used to understand the agent's future behavior.

    %------------------------------------------------
    \subsubsection{Action}
    \label{sec:App_C_action}
    %------------------------------------------------

Only one work explains the choice of an action by indicating future actions that could be taken. The work (Dodson et al. \citeyear{DBLP:conf/aldt/DodsonMG11}) is closely related to the problem studied and so future work could draw on it to explain RL agents (provided that the MDP is known or at least approximated).

                % natural language: action factored differential values + statistics explanation based on database Dodson (MDP context)
Dodson et al. provide a natural language explanation for advising students on their choice of courses. This problem, represented by a MDP, is explained using two modules, one of which is called Model-Based Explanation.
This module extracts information from the MDP about the usefulness of performing the recommended action in terms of future actions. 
As an example, if the student takes course $x$, this will enable him to take courses $y,z$ in the future.

\vspace{0.5em}
\label{par:App_C_mix_explanation}
The following four works use both types of explanation of action (i.e. \emph{Feature Importance} and \emph{Expected Outcomes}).
    % Mix Feature importance / Expected outcomes
    
        % Cognitive (belief-goal-emotion) / Perceptual (confidence / counterfactual (based on LIME) PeCoX framework (not necessarily RL) Neerincx
The PeCoX framework (Neerincx et al. \citeyear{DBLP:conf/hci/NeerincxWKD18}) proposes two types of explanation based on the perceptual and cognitive aspects of the agent. PeCoX models the explanation problem in three parts: explanation generation, communication to the user and reception.
The perceptual explanation comprises the calculation of a confidence index allowing the agent to express its confidence in its decision and a contrastive method that identifies an alternative output for comparison. 
The cognitive explanation consists of providing a restricted set of goals and beliefs for the agent. The authors also suggest modelling emotions to enrich the explanation. 
Note that the way in which the agent has learned and the explanation methods are not specified in this work, the idea being to provide a general framework for explaining agent behavior.

        % Interpretable POMDP through natural language templates Wang (+ state explanations)
        % Simply translate POMDP components into natural language sentences
An approach is presented in N. Wang et al. (\citeyear{DBLP:conf/atal/WangPH16}) for directly extracting explanations based on the POMDP. The idea is to retrieve the various types of information from the POMDP model and translate them into natural language sentences using predefined templates. For example, the agent can use its expected reward outcome to justify its choice of action, the transition function to describe the probability of achieving different outcomes, or its observation to clarify it to the user.

                % Human based natural language X Ehsan / Ehsan
The method used in the work by Ehsan et al. (\citeyear{DBLP:conf/aies/EhsanHCR18,DBLP:conf/iui/EhsanTCHR19}) consists of collecting a training corpus with the help of users so that the agent can explain its choice of action `as a human would have done'.
So the first step is to collect explanations associated with state-action pairs. Users have to `think aloud' while performing the agent's task. 
The second step is to train an encoder-decoder to return, based on a state-action pair, an explanation of the choice of action from the state, in a natural language form.

\paragraph{RL as an Explanation Framework}
\label{par:App_C_RL_explanation_framework}

    % Use of RL for explanation
A body of work uses RL to generate counterfactual inputs to a model that are close to the original input, to answer the question \emph{`Why predict a rather than b from instance x?'}
        % Use of DRL to craft counterfactuals Chen / Samoilescu 
In Z. Chen, et al. (\citeyear{DBLP:conf/cikm/ChenSW0AT22}), the problem of generating a counterfactual instance of a classification or regression model is described as a decision-making problem, which is solved using a DRL algorithm. Based on the same idea, Samoilescu et al. (\citeyear{DBLP:journals/corr/abs-2106-02597}) generates batches of counterfactuals from the instance to be explained.
        % Use of MARL to craft counterfactual (two inputs) Nguyen
For the Drug Target prediction problem, a multi-agent framework is proposed in Nguyen et al. (\citeyear{DBLP:journals/corr/abs-2103-12983}) to generate counterfactuals by taking into account two distinct inputs, the drug and the target, which are assimilated to two agents with distinct actions.
        % Use of DRL to find close decision boundary point (context human in the loop) Lash
To explain the classification of a point, Lash (\citeyear{DBLP:journals/corr/abs-2206-01343}) describes an approach using DRL to find the closest point to the classified point which is located in the decision boundary, based on the user's set of preferred features. 
        % Use of RL for Explainable recommendation Wang
The use of two RL agents allows the question \emph{`Why an item is recommended?'} to be answered by a personalised explanation (X. Wang, Y. Chen, et al. \citeyear{DBLP:conf/icdm/WangCYWW018}).
To explain the choice of the recommendation system $f$, agent $1$ learns to provide an explanation consisting of a subset of items and agent $2$ learns to use this to predict the output ratings: the explanation is a subset of items sufficient to provide the same output ratings as $f$.
        % Use of DQN to build an interpretable DT (context: classification tasks) Kohler
Kohler et al. (\citeyear{DBLP:journals/corr/abs-2304-05839}) use Iterative Bounding MDP (IBMDP) (Topin, Milani, et al. \citeyear{DBLP:conf/aaai/TopinMFV21}) to build compact and efficient DT's for classification tasks. In addition to a classical MDP, IBMDP includes feature bounds, information gathering actions and a reward function that defines the interpretability-performance trade-off.
        % SXp
The SXp method (Saulières et al. \citeyear{DBLP:conf/icaart/SaulieresCS23}) learns favorable and hostile agents representing the response of the environment with RL, in order to explain an RL agent.

\end{document}